\documentclass{egpubl}
\usepackage{eurovis2025}
\SpecialIssuePaper  
\CGFccby

\usepackage[T1]{fontenc}
\usepackage{dfadobe}  
\usepackage{cite}  

\BibtexOrBiblatex

\electronicVersion
\PrintedOrElectronic
\ifpdf \usepackage[pdftex]{graphicx} \pdfcompresslevel=9
\else \usepackage[dvips]{graphicx} \fi
\usepackage{egweblnk}

\usepackage{tabu}      
\usepackage{booktabs}
\usepackage{amsfonts}
\usepackage{amsmath}
\usepackage{soul}
\usepackage{caption}
\usepackage{multirow}
\usepackage{wrapfig}
\usepackage[table,svgnames]{xcolor}
\usepackage{pdfpages}
\usepackage{enumitem}
\definecolor{mylightgrey}{RGB}{232, 232, 232}  
\definecolor{myblue}{RGB}{134, 165, 244}   
\definecolor{myred}{RGB}{236, 91, 87}   
\definecolor{DRColor}{RGB}{73, 130, 80}   
\definecolor{CRColor}{RGB}{210, 135, 95}   
\definecolor{IRColor}{RGB}{170, 97, 163}  

\graphicspath{{./figs/}}

\newcommand{\IoU}{\textrm{IoU}}

\newcommand{\R}[1] {{\bfseries R#1}}
\newcommand{\C}[1] {{\bfseries C#1}}
\newcommand{\para}[1] {\noindent{\textbf{#1}}}

\newcommand\ChangeRT[1]{\noalign{\hrule height #1}}
\usepackage{array}

\newcommand{\systemname}{{\emph{VISLIX}}} 
\newcommand{\ConceptSlicer}{{\emph{ConceptSlicer}}} 
\newcommand{\AttributionScanner}{{\emph{AttributionScanner}}}

\usepackage{cleveref}
\crefname{figure}{Fig.}{Figs.}
\crefname{section}{Sec.}{Secs.}
\crefname{subsection}{Sec.}{Secs.}
\crefname{table}{Tab.}{Tabs.}
\Crefname{figure}{Fig.}{Figs.}
\Crefname{section}{Sec.}{Secs.}
\Crefname{subsection}{Sec.}{Secs.}
\Crefname{table}{Tab.}{Tabs.}
\title[{\systemname}]{{\systemname}: An XAI Framework for Validating Vision Models with Slice Discovery and Analysis}

\author[X. Yan, X. Xuan, J. Ono, J. Guo, V. Mohanty, S. Kumar, L. Gou, B. Wang, L. Ren]
{\parbox{\textwidth}{\centering 
Xinyuan Yan$^{1}$\orcid{0000-0003-3396-1310},
Xiwei Xuan$^{2}$\orcid{0000-0002-0828-8761},
Jorge Piazentin Ono$^{3}$\orcid{0000-0002-2424-0186},
Jiajing Guo$^{3}$\orcid{0000-0003-0511-136X},
Vikram Mohanty$^{3}$\orcid{0000-0001-6296-3134},\\
Shekar Arvind Kumar$^{4}$\orcid{0000-0001-7756-0901},
Liang Gou$^{5}$\orcid{0009-0006-9138-3351},
Bei Wang$^{1}$\orcid{0000-0002-9240-0700},
Liu Ren$^{3}$\orcid{0009-0002-1813-8844}
}\\
{\parbox{\textwidth}{\centering 
$^1$Scientific Computing and Imaging Institute, University of Utah, USA\\
$^2$University of California, Davis, USA\\ 
$^3$Bosch Research North America and Bosch Center for Artificial Intelligence (BCAI), USA\\
$^4$Robert Bosch GmbH, Germany\\
$^5$Splunk Technology, USA
}
}
}

\begin{document}

\maketitle
%---------------------------------
\begin{abstract}
Real-world machine learning models require rigorous evaluation before deployment, especially in safety-critical domains like autonomous driving and surveillance.  
The evaluation of machine learning models often focuses on data slices, which are subsets of the data that share a set of characteristics. 
Data slice finding automatically identifies conditions or data subgroups where models underperform, aiding developers in mitigating performance issues. 
Despite its popularity and effectiveness, data slicing for vision model validation faces several challenges.  
First, data slicing often needs additional image metadata or visual concepts, and falls short in certain computer vision tasks, such as object detection. 
Second, understanding data slices is a labor-intensive and mentally demanding process that heavily relies on the expert's domain knowledge. 
Third, data slicing lacks a human-in-the-loop solution that allows experts to form hypothesis and test them interactively.
To overcome these limitations and better support the machine learning operations lifecycle, we introduce {\systemname}, a novel visual analytics framework that employs state-of-the-art foundation models to help domain experts analyze slices in computer vision models. Our approach does not require image metadata or visual concepts, automatically generates natural language insights, and allows users to test data slice hypothesis interactively.
We evaluate {\systemname} with an expert study and three use cases, that demonstrate the effectiveness of our tool in providing comprehensive insights for validating object detection models. 

\end{abstract}  

%---------------------------------
\section{Introduction}
\label{sec:introduction}

Computer vision models are widely employed across various domains, including autonomous driving, where they detect nearby objects~\cite{caesar2020nuscenes, gou2020vatld}, and surveillance, where they identify suspicious activities~\cite{csengonul2023analysis}. 
Although computer vision models often achieve high overall performance, they may underperform on semantically coherent subsets of data, known as \emph{data slices} or \emph{edge cases}~\cite{zhang2022sliceteller, sagadeeva2021sliceline}. For example, adverse weather can hinder car detection~\cite{zhang2023perception}, and skin tones may affect pedestrian detection across demographic groups~\cite{wilson2019predictive}. To ensure safety, robustness, and fairness, AI developers must efficiently identify, understand, and address such slices before deployment~\cite{rahwan2019machine}.

Data slicing, a popular model validation technique, automatically identifies visually consistent yet underperforming data subgroups. It often relies on image metadata---textual labels for image attributes (e.g., weather = ``rainy'', lighting = ``dark'')---to partition subgroups~\cite{bordes2024pug, zhang2022sliceteller}, which, however, is labor-intensive to acquire.
To reduce this burden, {\ConceptSlicer}~\cite{zhang2024slicing} automatically checks the presence of predefined visual concepts (e.g., ``bus = 1'', ``car = 0'') using semantic segmentation models.
However, both image metadata and segment tags can miss important image details.
More advanced methods~\cite{eyuboglu2022domino, d2022spotlight} use clustering-based techniques on image embeddings to identify error-consistent groups. However, these approaches are designed for image classifiers that consider global context and are not well-suited to object detectors, which prioritize local context~\cite{boreiko2023identifying}.

After identifying slices, experts must analyze them to form hypothesis about the model's failure modes. This interpretation step is essential for downstream tasks like model optimization~\cite{eyuboglu2022domino} and stakeholder communication~\cite{balayn2023faulty}. Yet, the process is demanding, requiring experts to examine slice images, infer failure root causes, and summarize common patterns. 
A recent study~\cite{johnson2023does} further highlights the complexity of this task, as experts may hold inconsistent or biased explanations. Besides analyzing the data slices, experts often come with their own hypothesis about model failures. However, existing slice discovery methods are meant to run only once and cannot interact with users~\cite{johnson2023does}.

To address these challenges, we present {\systemname} (VIsual SLIce eXplanations), an explainable artificial intelligence (XAI) framework that leverages Large Language Models (LLMs) and Vision-Language Models (VLMs) to support human-in-the-loop slice discovery and analysis for vision model validation, with a focus on object detection.
{\systemname} identifies slices using context-aware image embeddings to uncover systematic errors. It then generates natural language data slice explanations, using a conversational vision-language framework.
Our visual analytics system enables experts to efficiently explore slices, inspect and refine slice details, and test hypothetical scenarios through direct visual manipulation and natural language queries. In summary, our framework makes the following contributions:
\begin{itemize}[noitemsep,leftmargin=*]
\item We propose a new slice discovery method tailored for object detector validation, which does not rely on either image metadata or visual concepts. Our approach is able to automatically describe and explain problematic data slices in natural language.
\item We design a visual analytics system that enables users to efficiently explore data slices and interactively test new hypotheses by creating slices via both visual and textual queries. 
\item We validate {\systemname} with an expert study and three use cases, which demonstrate that {\systemname} efficiently finds and explains a broad spectrum of data slices. We also show how insights gained from {\systemname} can be leveraged to enhance model robustness through fine-tuning.  
\end{itemize}

\section{Related Work}
\label{sec:related-work}

\textbf{Data slice finding.}~Slice-finding techniques uncover data subgroups where ML models underperform, exposing systematic errors~\cite{chung2019slice, pastor2023hierarchical}. Tools like \emph{DivExplorer}~\cite{pastor2021looking-divexp}, \emph{SliceLine}~\cite{sagadeeva2021sliceline}, and \emph{Macrobase}~\cite{bailis2017macrobase} use frequent itemset mining (FIM) algorithms for slice discovery, such as Apriori~\cite{agrawal1994fast} and FP-growth~\cite{han2000mining}. These approaches have been adapted for image datasets by using metadata attributes to define subgroups~\cite{zhang2022sliceteller, unieval}. However, obtaining structured metadata for images remains challenging~\cite{xuan2025attributionscanner, xuan2024slim}.

{\ConceptSlicer}~\cite{zhang2024slicing} employs an image segmentation model to identify predefined concepts (e.g., \emph{person}, \emph{bus}, \emph{car}) and treat them as metadata, analyzing performance of concept combinations (e.g., $bus=1$ {\&} $car=0$). Yet, its reliance on predefined concepts limits its descriptiveness and coverage. To address this issue, many methods encode images into latent spaces and cluster them to find problematic slices~\cite{wilesdiscovering, eyuboglu2022domino, d2022spotlight}, e.g., \emph{GEORGE}~\cite{sohoni2020no} and \emph{UDIS}~\cite{krishnakumar2021udis} use over-clustering and hierarchical clustering, whereas  {\AttributionScanner}~\cite{xuan2025attributionscanner} applies K-means on attribution-weighted embeddings to find spurious correlation slices.
However, these methods target image classifiers that utilize the entire image and are not well-suited for object detection, where the object and its surroundings are crucial~\cite{liu2020deep}.

To capture local context, prior work has expanded the detection window and fused their embeddings during model training to enhance performance~\cite{li2016attentive, zhu2017couplenet}.
We adapt this approach to a different setting: post hoc slice discovery, enabling more effective validation for object detection tasks.

\noindent\textbf{Data slice explainability.}~Prior research on interpreting data slices follows two main approaches. 
The first utilizes saliency maps~\cite{lee2022viscuit, xuan2024suny} to highlight image regions influencing model predictions. 
While effective for individual images, this method requires domain expertise~\cite{balayn2023faulty}, risks confirmation bias~\cite{adebayo2018sanity}, and lacks scalability. In contrast, natural language explanations have gained traction for their clarity and utility in tasks such as stakeholder communication~\cite{balayn2023faulty}, failure reporting~\cite{cabrera2021discovering}, and model fine-tuning~\cite{zhang2024slicing}.
The rise of foundation models, including LLMs like ChatGPT~\cite{chatgpt} and LLaMA 2~\cite{touvron2023llama}, and VLMs like CLIP~\cite{Radford2021}, BLIP-2~\cite{li2023blip}, LLaVA~\cite{liu2024visual, liu2024improved}, and GPT-4~\cite{openai2023gpt4}, has enabled more sophisticated textual slice explanations. For example, \emph{Domino}~\cite{eyuboglu2022domino} uses CLIP to describe slices via predefined templates, while Jain et al.\cite{jain2022distilling} and Wiles et al.\cite{wilesdiscovering} leverage captioning models to summarize slice content. {\ConceptSlicer}~\cite{zhang2022sliceteller} applies LLMs to describe general scenes based on visual concepts, primarily for data augmentation.
However, these methods often provide high-level descriptions, overlooking nuanced errors. Our approach leverages generative VLMs and LLMs to produce template-free summaries of slice errors, capturing subtle details and significantly expanding the scope of explanations.
% We are able to not only describe the shared characteristics of a slice, but also explain why the model fails.

\noindent\textbf{Interactive systems for data slice analysis.}~Various interactive systems support slice exploration for vision model validation~\cite{zhang2022sliceteller, krishnakumar2021udis, cabrera2021discovering, cabrera2023zeno, lee2022viscuit, xuan2025attributionscanner, pastor2021looking-divexp}. 
Some systems first generate slices and present them in an overview+detail format. For instance, \emph{SliceTeller}~\cite{zhang2022sliceteller} uses \emph{DivExplorer}~\cite{pastor2021looking-divexp} to create slices and visualizes them with matrix encodings, with details in a separate panel. \emph{VISCUIT}~\cite{lee2022viscuit} lists slices generated by \emph{UDIS}~\cite{krishnakumar2021udis} and enables neuron activation inspection. {\ConceptSlicer}~\cite{zhang2024slicing} supports slice viewing, concept inspection, and training data augmentation, while \emph{Uni-Evaluator}~\cite{unieval} visualizes global performance and slices using matrix, table, and grid-based encodings. Interactive slice discovery tools include \emph{AdaVision}~\cite{gao2023adaptive}, which retrieves images via natural language for iterative testing, and \emph{ESCAPE}~\cite{ahn2023escape}, which identifies spurious associations in image classifiers with UMAP plots. 
Our system combines efficient slice inspection with interactive discovery, leveraging visual exploration and natural language queries.
\section{Background on Object Detection}
\label{sec:background}

\begin{figure}[!ht]
  \centering
  \vspace{-1pt}
  \includegraphics[width=0.95\columnwidth]{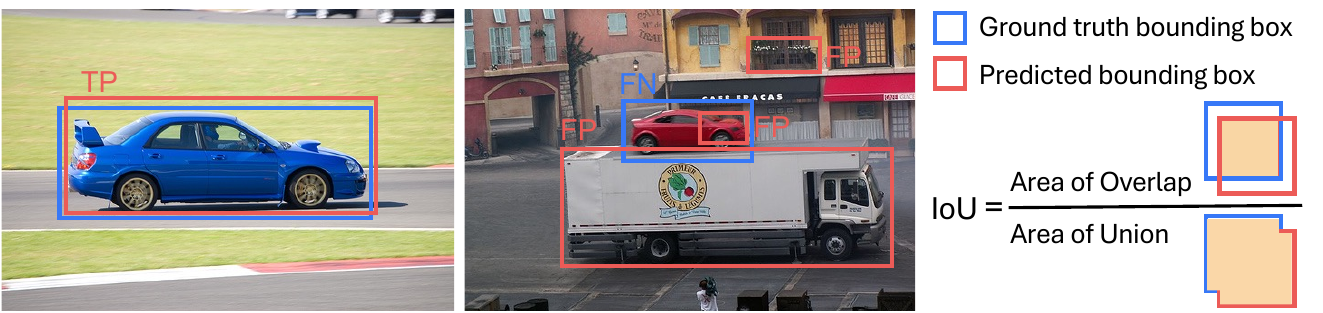}
  \vspace{-5pt}
  \caption{Illustrations of True Positive (TP), False Positive (FP), and False Negative (FN) for a car detector and IoU computation. Left: $\IoU > 0.5$ (correct). Middle: $0 \leq \IoU < 0.5$ (incorrect).}
  \vspace{-5pt}
  \label{fig:TP-FP-FN}
\end{figure}

Object detection identifies object classes (e.g., car, pedestrian) and their spatial locations (e.g., bounding boxes) in an image.
Given an object detector like a car detector, the goodness of the predicted car location is accessed via the \textit{Intersection over Union} (IoU), a metric computed as the ratio of the intersection area to the union area of predicted and ground truth bounding boxes.
A detection is considered correct if IoU exceeds a threshold, typically set to $0.5$.
With this threshold, a bounding box is labeled as one of three types, as shown in~\cref{fig:TP-FP-FN}: \textit{True Positive} (TP), where the detector correctly identifies a car; \textit{False Positive} (FP), where the detector erroneously identifies other objects as a car; \textit{False Negative} (FN), where the ground truth car is not correctly detected. 
The detection outcome is influenced by both object characteristics like color and pose, and its surrounding context that offers valuable cues for model perception~\cite{liu2020deep, zou2023object}.
For example, the unidentified red car in~\cref{fig:TP-FP-FN} (middle) could be attributed to its unusual position on top of a truck.
To quantify the overall performance of a detector, the most widely used metric is \textit{mean Average Precision} (mAP) derived from precision (i.e., $\frac{TPs}{TPs+FPs}$) and recall (i.e., $\frac{TPs}{TPs+FNs}$).   
% The faulty detections can be attributed to various factors, such as misclassification, partial detection, occlusion, and inaccurate ground truth annotations. 
See~\cite{zou2023object} for a complete survey.

\section{Domain Challenges and Design Requirements}
\label{sec:bg-requirement}

Throughout this project, we closely collaborated with two industry computer vision experts to understand their needs and gather insights. 
Both experts hold PhDs in machine learning and work on various computer vision tasks for autonomous driving and driver assistance systems. 
Based on their feedback and an extensive literature review, this section outlines the key challenges (\cref{sec:challenges}) that domain experts face in identifying and understanding data slices, particularly for object detector validation. These challenges motivate the formulation of our design requirements (\cref{sec:requirements}).

\subsection{Domain Challenges}
\label{sec:challenges}

\para{C1.~Existing slice discovery methods fall short in object detection.}  
Prior methods often rely on expert-defined image metadata~\cite{idrissi2022imagenet} or visual concepts~\cite{zhang2024slicing}. 
However, anticipating all critical slices is impractical~\cite{chung2019slice}, and such data misses complex scenarios (e.g.,~``car occluded by a tree'').  
Recent methods~\cite{eyuboglu2022domino, d2022spotlight} use image embeddings and clustering to generate slices. 
While effective for image classifiers, they struggle with object detectors, as global embeddings overlook detection locality and cannot distinguish multiple detections in the same image.  

% In model evaluation, (data) slices are subgroups of data (e.g., ``lighting = dark'') with degraded performances.  
% Prior methods for identifying slices in vision models often rely on image  metadata~\cite{idrissi2022imagenet}, which are hard to acquire and limited in expressing complex scenarios (e.g.,~``car occluded by a tree'')~\cite{zhang2024slicing}.

\para{C2.~Interpreting data slices is a nontrivial task.} 
Experts interpret slices to uncover model failure patterns, guiding refinement and deployment decisions~\cite{cabrera2021discovering}. 
This process—examining slice images, inferring failure causes, and summarizing patterns—is time-consuming and prone to bias or errors~\cite{johnson2023does}.
%This involves analyzing slice images to infer failure causes and summarize slice patterns, a process that is labor-intensive and prone to bias or errors~\cite{johnson2023does}. 
To simplify this task, prior methods have used predefined templates~\cite{eyuboglu2022domino} or sampled image captions~\cite{wilesdiscovering, jain2022distilling}, but these approaches lack flexibility and fail to adequately explain detection errors.

\para{C3.~A human-in-the-loop solution is needed for slice exploration and hypothesis testing.} 
In a typical data slicing workflow, experts run a slice discovery algorithm once and then inspect only the output slices. 
Prior studies~\cite{johnson2023does, balayn2023faulty} emphasize the need for visual analytics systems that effectively present slices to users, while accounting for cognitive load and enabling efficient slice navigation. 
Moreover, since automated methods may not cover all scenarios of interest~\cite{gao2023adaptive},
interactive workflows could be designed to leverage the stakeholder’s domain knowledge in order to define coherent subsets of data~\cite{johnson2023does}.

\subsection{Design Requirements}
\label{sec:requirements}

% Based on the identified challenges, we list four corresponding design requirements that our framework needs to fulfill.

\para{R1. Automate slice discovery for object detectors beyond metadata and visual concepts.} 
Given the limitations of prior methods (\C{1}), our framework should extract more discriminative features in order to slice the data.
Each slice should exhibit coherent error patterns, and metrics should be provided to assess its significance.

\para{R2. Provide natural language explanations for slices.} 
Since slice interpretation is labor-intensive and prone to errors (\C{2}), our framework should generate descriptive sentences that explain slice scenarios and root causes, aiding experts in their reasoning process. 
We choose free-text explanations as they can effectively describe nuanced and complex image features and are inherently interpretable by humans~\cite{marasovic2020natural, marasovic2022few}. 
% Further, textual explanations are directly actionable in various tasks like client communication and data retrieval for model enhancement.

\para{R3. Support efficient slice inspection and refinement.} 
Considering the substantial efforts involved in reviewing data slices (\C{3}), our framework should effectively represent them to users. 
The system should (\textbf{R3-1}) offer concise overviews of data slices, (\textbf{R3-2}) highlight key statistics to help users identify slices of interest, (\textbf{R3-3}) provide detailed slice information on demand, and (\textbf{R3-4}) allow users to edit slices and explanations in case of inaccuracy.

\para{R4. Enable interactive validation of user-defined slices.} 
Beyond existing slices, our framework should enable users to test hypothesis (\C{3}) by (\textbf{R4-1}) discovering new slices via the visual interface and (\textbf{R4-2}) defining slices with natural language descriptions.
%\begin{itemize}[noitemsep]
%    \item \para{R4-1.} Unveil slices that are overlooked by the slice discovery methods by directly manipulating the visualization interface.
%    \item \para{R4-2.} Define slices based on domain experts' experiences using natural language descriptions.
%\end{itemize}
Once a hypothetical slice is created, our framework should compute slice metrics and explanations to facilitate hypothesis validation.
\section{The {\systemname} Framework}

\begin{figure*}[!ht]
  \centering
  \vspace{-2pt}
  \includegraphics[width=0.8\linewidth]{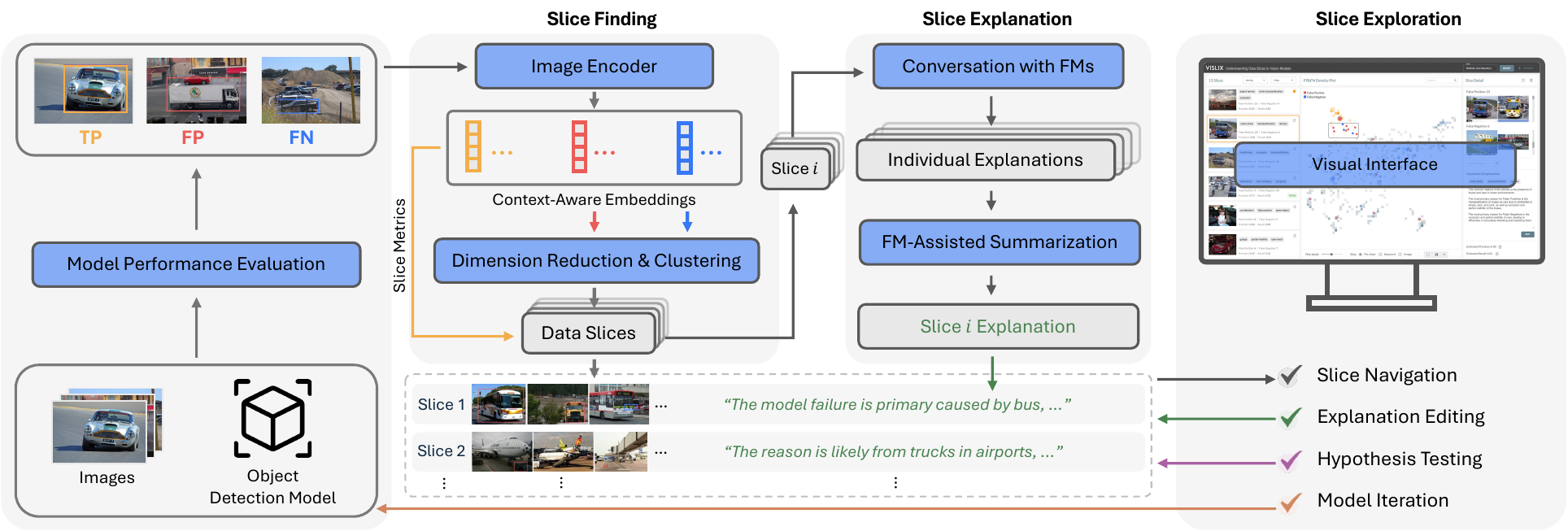}
  \vspace{-5pt}
  \caption{{\systemname} workflow. 
  \textbf{Inputs}: True Positives (TPs), False Positives (FPs), and False Negatives (FNs) from an object (car) detector and validation images. 
  \textbf{Slice finding}: Identifying data slice via image embeddings of FPs and FNs and estimating slice metrics via TPs. \textbf{Slice explanation}: Producing free-text explanations for each slice based on individual explanations of FPs and FNs, leveraging foundational models. \textbf{Slice exploration}: A visualization system that integrates all slices and explanations, enabling slice examination and validation.}
  \label{fig:framework}
  \vspace{-12pt}
\end{figure*}

Guided by the design requirements, we develop {\systemname}, a framework that automatically identifies data slices in object detectors, explains them in natural language, and enables expert exploration and hypothesis testing via a visual analytic system.  
Our framework uses large foundation models, such as VLMs for bridging images and text, and LLMs for reasoning about slice errors.  
The {\systemname} workflow (\cref{fig:framework}) processes TPs, FPs, and FNs from a trained object detector (e.g., a car detector) on validation images in three phases:  

\para{Slice Finding.} Each detection is converted into a context-aware embedding, followed by dimensionality reduction and clustering of FP and FN embeddings to identify slices. 
Slice importance is then estimated using nearby TPs (\cref{sec:slice-finding}, addressing \R{1}).

\para{Slice Explanation.} Assisted by VLMs and LLMs, we generate a free-text explanation for each slice by first producing textual explanations for each FP and FN within the slice, and then summarizing them into a slice explanation (\cref{sec:slice-explanation}, addressing \R{2}). 

\para{Slice Exploration.} 
We build a visual system that integrates slices and their explanations, enabling users to explore slices, rectify errors, test hypotheses, and enhance model performance (\cref{sec:slice-exploration}, addressing \R{3} and \R{4}). 
 At the project's inception, we utilized VLMs like BLIP-2\cite{li2023blip} and LLaVA~\cite{li2024llavamed} and LLMs like GPT 3.5~\cite{openai2023gpt4}) for their competitive performances, but {\systemname}  can easily integrate new models to keep up with model advancements.

\begin{figure}[!ht]
  \centering
  \includegraphics[width=0.9\columnwidth]{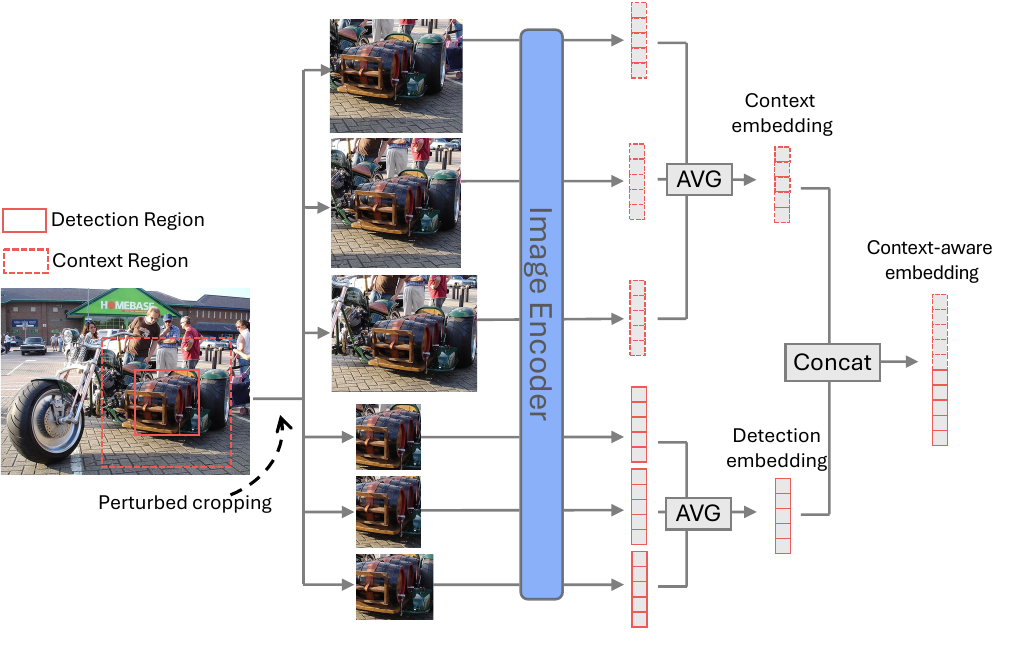}
   \vspace{-10pt}
  \caption{Context-aware embedding generation.
  }
  \vspace{-8pt}
  \label{fig:context-aware-embed}
\end{figure}

\subsection{Slice Finding}
\label{sec:slice-finding}

To identify data slices, we combine the cropping strategy~\cite{liu2020deep} that is widely used in computer vision to extract objects' local context, and embedding-based slice-finding methods~\cite{eyuboglu2022domino, d2022spotlight} that excel in image classifiers.

\para{Context-aware embeddings.}
For each FP, FN, and TP, we construct a \textit{context-aware} embedding by fusing features of the detection and its surroundings. This is achieved by enlarging the detection window and concatenating embeddings, a common practice in object detection~\cite{gidaris2015object, li2016attentive, zhu2017couplenet}. As illustrated in~\cref{fig:context-aware-embed}, for an FP detection featuring three barrels, we create a context region twice the size of the detection window, as per Zhu et al.~\cite{zhu2017couplenet}. The enlarged context reveals that the barrels are in a motorcycle sidecar, likely causing the error. To enrich feature diversity, we crop three patches for each region with random perturbations of up to $10\%$ expansion per side. The patch embeddings, obtained via BLIP-2's image encoder, are averaged to create the \textit{detection} or \textit{context} embeddings. These are concatenated to generate the final \textit{context-aware} embedding, highlighting the barrels within the broader sidecar context to aid error reasoning.

\para{Clustering-based slice discovery.}
\label{para:slice-clsuters}
We identify data slices as dense regions in the \textit{context-aware} embedding space of FPs and FNs, which reveal shared features indicative of systematic errors. Specifically, we reduce embeddings to 10 dimensions using UMAP~\cite{mcinnes2018umap}, following prior work~\cite{mcconville2021n2d, sohoni2020no} that demonstrates UMAP's effectiveness in cluster detection. Next, we apply HDBSCAN~\cite{campello2013density} to identify high-density clusters (i.e., non-noise instances) while excluding isolated failures (i.e., noise instances). 
Unlike DBSCAN, HDBSCAN can find regions with varying densities and requires fewer parameters.
We tune the clustering parameters using  Silhouette ~\cite{rousseeuw1987silhouettes}, informed by prior studies~\cite{mcconville2021n2d, sohoni2020no}. Details are offered in the supplement. \Cref{fig:slice-examples} illustrates two slices from a car detector: one in an airport setting and another featuring motorcycles.

\para{Slice metrics.} 
For every data slice, we compute two performance metrics: precision and recall. To identify the TP samples belonging to the slice, we measure the Euclidean distance between the slice and its nearby TPs in the \textit{context-aware} embedding space. 
For each slice, we compute the average nearest neighbor distance of each instance within that slice. 
A TP is assigned to the slice if its distance to the nearest instance in the slice is smaller than this average nearest neighbor distance. 
We then calculate the precision and recall to assess the significance of each slice, giving more attention to those with lower values. 

\subsection{Slice Explanation}
\label{sec:slice-explanation} 

The slice explanation is generated in two steps: first, individual explanations for FPs and FNs are computed using a VLM (LLaVA) and an LLM (GPT 3.5, hereafter GPT). Then, the individual explanations are summarized into a cohesive slice explanation.

\begin{wrapfigure}{r}{0.23\textwidth} 
    \centering
    \vspace{-4mm}
    \includegraphics[width=0.23\textwidth]{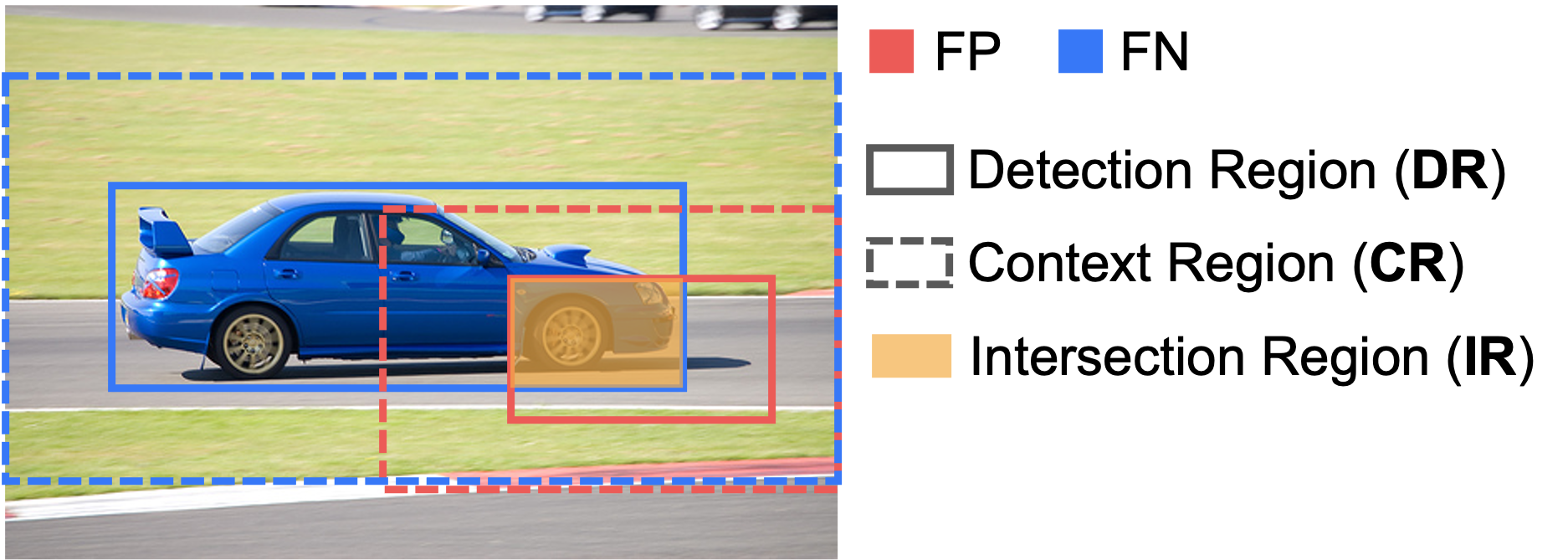}
    \vspace{-6mm}
    \caption{Region annotations.}
    \label{fig:regions-annotation}
    \vspace{-4mm}
\end{wrapfigure}

\begin{figure*}[!ht]
  \centering
  \vspace{-2pt}
  \includegraphics[width=0.8\linewidth]{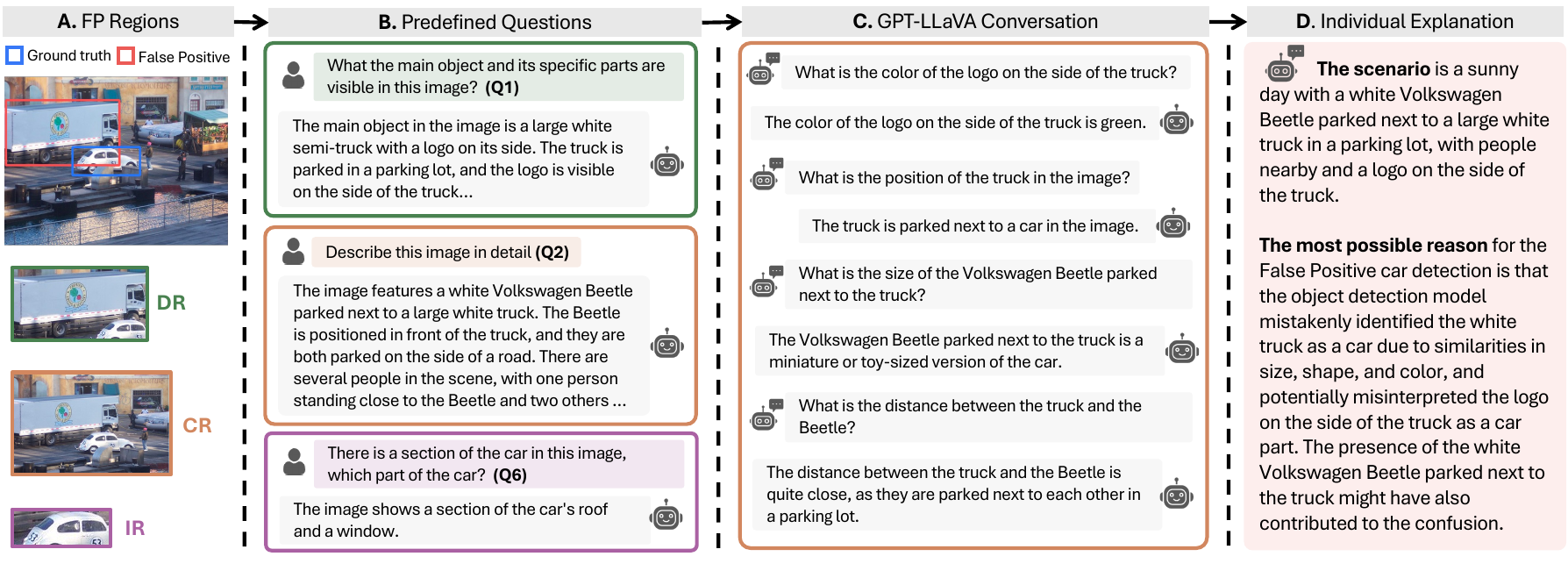}
  \vspace{-15pt}
  \caption{Explanation generation for an FP in a car detector. \textbf{A}: Three regions derived from the FP: detection region (\textcolor{DRColor}{DR}), context region (\textcolor{CRColor}{CR}), and intersection region (\textcolor{IRColor}{IR}). \textbf{B}: LLaVA answers predefined questions regarding different regions. \textbf{C}: GPT uses answers from \textbf{B} to chat with LLaVA about the \textcolor{CRColor}{CR}. \textbf{D}: GPT explains the FP based on all the acquired information.}
  \label{fig:FP-explanation-example}
  \vspace{-10pt}
\end{figure*}

\begin{figure}[!ht]
  \centering
  % \vspace{-1mm}
  \includegraphics[width=0.9\columnwidth]{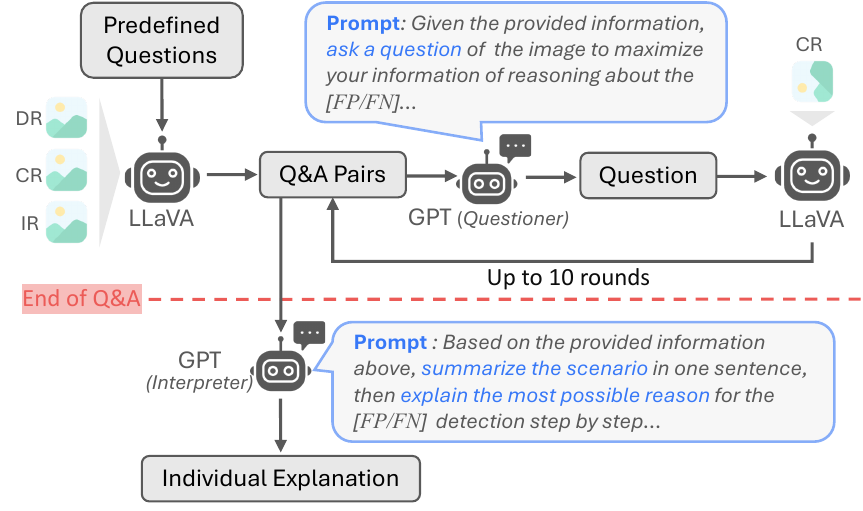}
  \vspace{-5pt}
  \caption{The process of generating an individual explanation.}
  \vspace{-12pt}
  \label{fig:LLAVA-QA}
\end{figure}

\para{Individual explanations.} 
To generate a natural-language explanation for an FP or FN, we first gather textual information about the image that may have caused the error. As shown in~\cref{fig:regions-annotation}, we crop multiple patches to improve analysis accuracy: the detection region (DR), representing the predicted region for an FP or the ground truth region for an FN; the context region (CR), which expands the detection region by 2$\times$ following~\cite{zhu2017couplenet} (see~\cref{sec:slice-finding}); and the intersection region (IR), the overlap between the detection and ground truth when {\IoU} exceeds $0.2$ based on our preliminary experiments, indicating partial detection. An example is shown in~\cref{fig:FP-explanation-example}-A.

\Cref{fig:LLAVA-QA} shows the procedure for generating an individual explanation. 
We first acquire answers from LLaVA to several predefined questions (\cref{tab:predefine-questions}) tailored for different regions that are generally critical for error reasoning. 
For instance, we prompt LLaVA to identify objects in the detection region of an FP (Q1) that may have caused the misclassification, describe the context region (Q2) for scene understanding, or describe the detected portion of the target object in the intersection region (Q6) to analyze partial detection.  ~\Cref{fig:FP-explanation-example}-B illustrates LLaVA's responses to these queries. 
\begin{table}[!ht]
\fontsize{7}{9}\selectfont
\begin{center}
\begin{tabular}{ c|p{7cm} } 
\ChangeRT{0.8pt}
\textbf{Regions} & \textbf{Predefined questions}  \\ 
\ChangeRT{0.8pt}
\textbf{DR} & \textbf{Q1.} FP: What the main object and its specific parts are visible in this image?  FN: Describe the [\textit{obj}] shown in the image and evaluate whether the entire [\textit{obj}] is clearly visible.\\

\ChangeRT{0.5pt}

\multirow{4}{*}{\textbf{CR}} & \textbf{Q2.} Describe this image in detail.  \\ 
& \textbf{Q3.} What is the weather in this image?  \\ 
& \textbf{Q4.} How is the lighting condition in this image?  \\ 
& \textbf{{Q5}\textsuperscript{*}.} Is the view of the [\textit{obj}] in the image obstructed? If so, what is obstructing it?  \\ 
\ChangeRT{0.5pt}
\multirow{1}{*}{\textbf{IR}} & \textbf{{Q6}\textsuperscript{*}.} There is a section of the [\textit{obj}] in this image, which part of the [\textit{obj}]?\\
\ChangeRT{0.8pt}
\end{tabular}
\vspace{-5pt}
\caption{Predefined questions for detection region (DR), context region (CR), and intersection region (IR). Q5 and Q6 are used exclusively for instances with $\IoU > 0.2$.}
\label{tab:predefine-questions}
\end{center}
\vspace{-20pt}
\end{table}

Since various factors can influence detection accuracy, a fixed set of universal questions is impractical. Building upon prior work~\cite{zhuchatgpt} that demonstrates GPT's ability to generate insightful questions with appropriate prompts, we prompt GPT (\textit{Questioner}) with question-answer pairs to formulate questions aimed at uncovering the error's root cause. This question is then posed to LLaVA, and its response is fed back to GPT for further question generation. This iterative dialogue continues until GPT outputs ``STOP'' or completes 10 iterations, following~\cite{zhuchatgpt} (\cref{fig:FP-explanation-example}-C); Finally, GPT (\textit{Interpreter}) summarizes the interactions between LLaVA, predefined questions, and GPT into a concise explanation of the scene and its primary cause. \Cref{fig:FP-explanation-example}-D illustrates an FP explanation. All explanations of FPs and FNs are precomputed.

\begin{figure}[!ht]
  \centering
  \vspace{-5pt}
  \includegraphics[width=0.9\columnwidth]{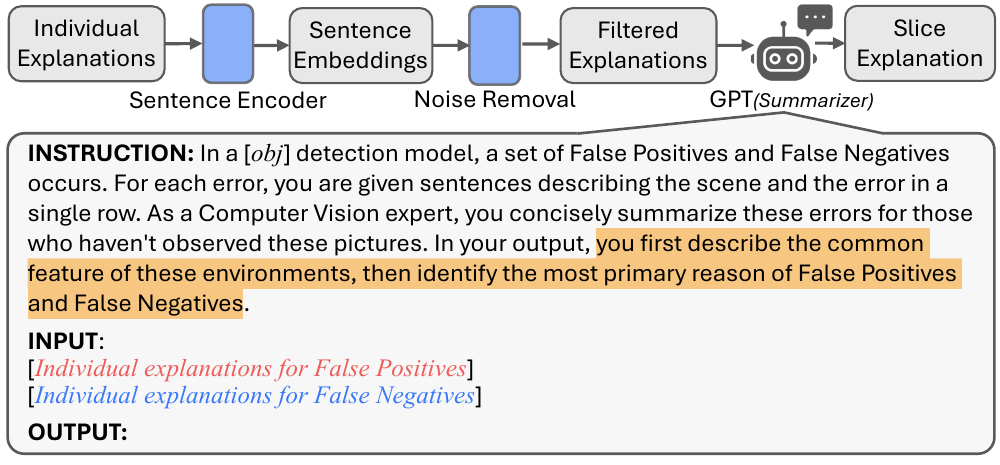}
  \vspace{-5pt}
  \caption{Pipeline for generating a slice explanation from individual explanations and the prompt template used for GPT.}   
  \label{fig:slice-explanation-process}
  \vspace{-8pt}
\end{figure}

\begin{figure*}[!ht]
  \centering
  \vspace{-2pt}
  \includegraphics[width=0.82\linewidth]{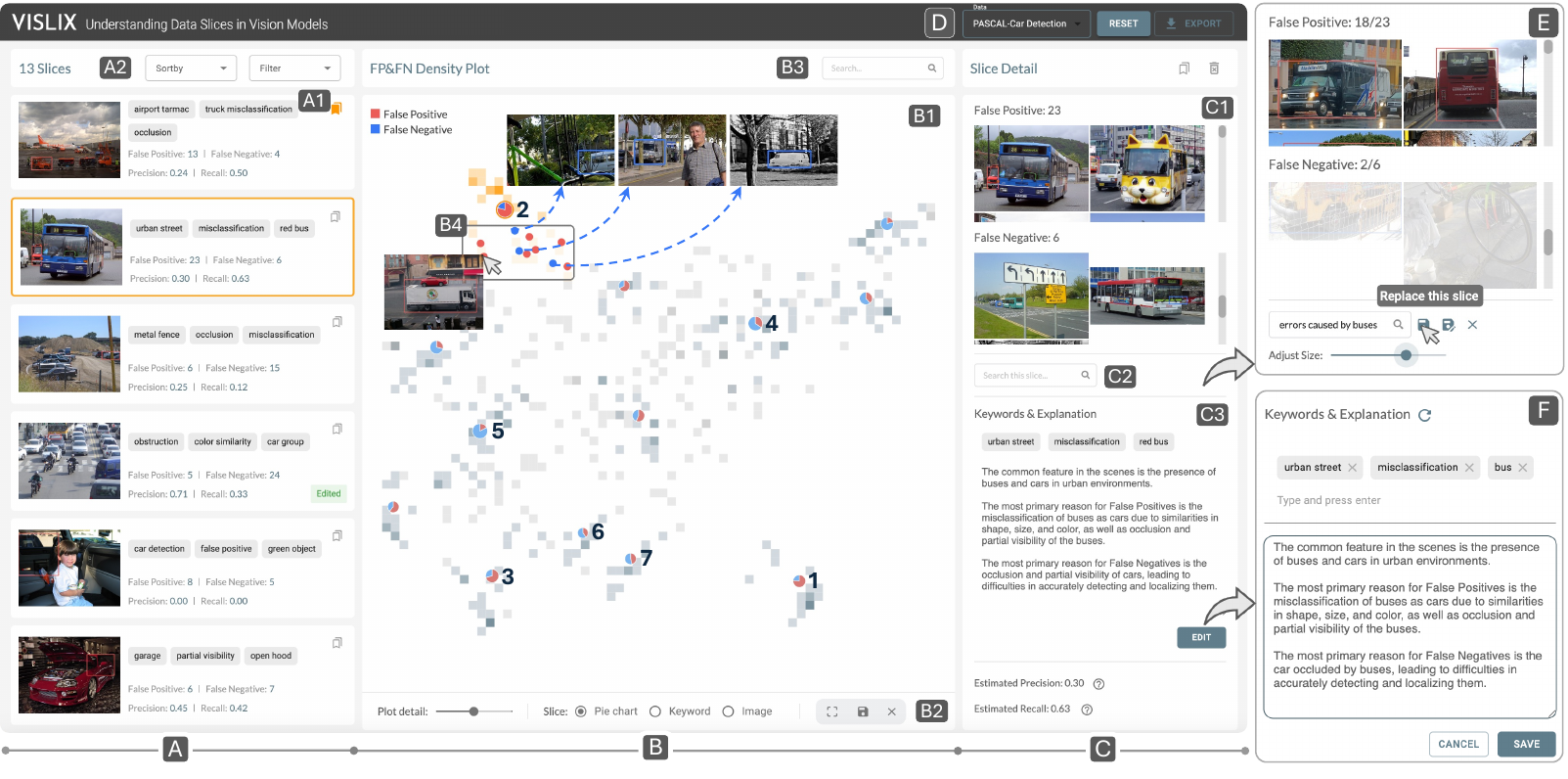}
  \vspace{-12pt}
  \caption{The \systemname{} interface visualizes slices and explanations from a car detector. \textbf{Slice list view (A)} shows all slices as summary cards (A1), with sorting and filtering options (A2). \textbf{Slice plot view (B)} includes a density plot (B1) showing slice locations in the embedding space (with slices 1–7 annotated for reference), an editing bar (B2) for operations like brushing to create hypothetical slices (B4), and a global search bar (B3) for hypothetical slice creation via queries.
 \textbf{Slice detail view (C)} presents details of a selected slice (slice 2), supporting query-based refinement (E) and explanation editing (F). \textbf{Toolbar (D)} enables data switching, operation resets, and slice export.}
  \vspace{-12pt}
  \label{fig:interface}
\end{figure*}

\para{Summarizing individual explanations.} 
We derive slice explanations by aggregating individual FP and FN explanations using GPT (\textit{Summarizer}). However, these explanations can be inconsistent or erroneous due to hallucinations or outliers, affecting quality.  
To address this, as shown in~\cref{fig:slice-explanation-process}, we convert explanations into text embeddings with a sentence transformer, then select those closest to the centroid, ensuring up to $80\%$ of the slice is represented or the total token count stays within $2000$ (input length limit). Using these filtered explanations, GPT (\textit{Summarizer}) generates a description of the scene and identifies root causes for FPs and FNs.
\begin{figure}[!ht]
  \centering
  \includegraphics[width=0.95\columnwidth]{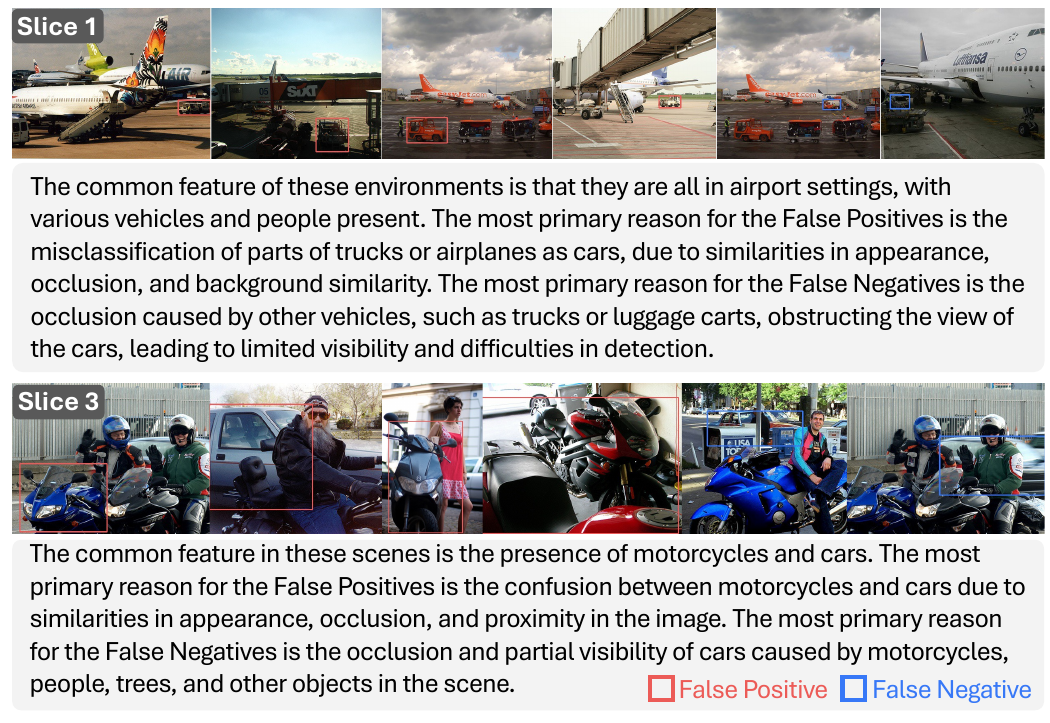}
  \vspace{-10pt}
  \caption{A random sample of FPs and FNs in slices 1 (top) and 3 (bottom) and their explanations in the car detector.}
  \vspace{-12pt}
  \label{fig:slice-examples}
\end{figure}
\Cref{fig:slice-examples} shows two slices from a car detector with their explanations.  
Full prompts and additional examples are provided in the supplement.

\subsection{Slice Exploration Through an Interactive System} 
\label{sec:slice-exploration}

To enable users to analyze the generated slices and explanations (\R{3}) and test slice hypotheses (\R{4}), we present a visual analytics system (\cref{fig:interface}) with four linked views: slice list view (A), slice plot view (B), slice detail view (C), and a toolbar (D).
View A lists all produced slices (\R{3-1}), summarizing key information in cards  (\cref{fig:interface}-A1). 
To prevent overwhelming users, each card shows three keywords that capture the slice's scenarios and error causes, generated by GPT using the same input as the slice explanation (\cref{fig:slice-explanation-process}) but with modified task descriptions.
Further details are provided in the supplement. 
In the following sections, we detail views B and C and discuss our key design decisions in between.

\subsubsection{Slice Plot View} 
\label{sec:viewB}

The \textbf{2D density plot} (\cref{fig:interface}-B1) provides a slice overview in the embedding space using the 2D dimensionality reduction result of UMAP applied to the \textit{context-aware} embeddings of FPs and FNs. 
To better reflect the data distribution used in slice discovery (\cref{sec:slice-finding}), UMAP parameters are largely preserved, with the component number reduced from 10 to 2 and the minimum distance increased to 0.15 to mitigate visual clutter. 

Next, we divide the scatterplot into a $64\times64$ grid by default and calculate the densities of noise and non-noise instances, as determined by HDBSCAN (\cref{sec:slice-finding}), in each cell. 
These densities are then mapped to two distinct colormaps---gray for noise and blue (or orange, its complementary color, when the slice is selected) for non-noise---and blended together as the grid rendering.
Thus, this plot characterizes the data distribution while emphasizing slices. We employ the density plot as the primary visualization due to its scalability for dense scatterplots and, more importantly, its intuitive representation of HDBSCAN’s mechanism, where clusters emerge from high-density areas, thus enhancing user trust in the data slices. 
For instance, variations in cluster area and color indicate slice size and compactness, whereas spatial locations reflect slice similarities, enabling users to quickly pinpoint slices of interest and facilitate a straightforward comparison among slices (\R{3-1} and \R{3-2}). We discuss alternative design in the supplement.

\textbf{Editing bar} (\cref{fig:interface}-B2) offers three operations on the density plot. 
On the left, users can adjust the granularity (grid resolution) of B1 to adapt to varied data scales.
In the middle, three types of slice encodings are offered based on experts’ suggestion: pie charts (depicting slice size and the percentage of FPs and FNs), the first slice keyword, and one representative image. 
Each encoding is centered within its respective slice.
Further, the slice list view (\cref{fig:interface}-A) and plot view (\cref{fig:interface}-B) are cross-filtered: 
selecting a slice in view B highlights both its density area and the corresponding slice card in view A in orange, and vice versa (\R{3-2}).

With the brush tool (\cref{fig:interface}-B4), users can select an area of interest by clicking and dragging, which creates a semitransparent overlay on the view.
FP and FN instances within the selected region are displayed as red and blue circles, respectively.
Hovering over a point reveals its corresponding image.
If users identify consistent patterns in this area, they can click the save icon to generate a hypothetical slice.
The backend computes the slice's statistics and explanations, which are then displayed in the interface.
Users can choose to keep or delete it once they test their hypotheses (\R{4-1}).

\textbf{Global search bar} (\cref{fig:interface}-B3) allows experts to create hypothetical slices via textual queries (\R{4-2}).
As shown in \cref{fig:snow-slice}, after users input a scenario description, the system retrieves instances matching the query based on cosine similarity between sentence embeddings.
Embeddings of individual explanations are precomputed and stored in the backend.
By default, instances with a similarity score above 0.5 are returned.
A slider lets experts adjust the similarity threshold to refine results.
Filtered FPs and FNs appear as circles on the density view, with images revealed on mouseover.
If the results align with expectations, users can save the slice.

\subsubsection{Slice Detail View}
\label{sec:viewC}
The slice detail view (\cref{fig:interface}-C) provides details about the selected slice  (\R{3-3}). C1 displays FPs and FNs with bounding boxes rendered on full images, and allows users to toggle between the full image and a magnified detection area. C2 includes a search box for refining slices by querying specific scenarios; filtered instances are determined by query similarity, and highlighted in the density plot and view E. Users can save these results as a new slice or replace the current one. C3 shows keywords, explanations, and slice metrics. To address potential hallucinations, users can switch to editing mode (\cref{fig:interface}-F) by clicking “Edit,” where they can refresh keywords and explanations via GPT or manually edit them (\R{3-4}).

\section{Expert Study}
\label{sec:expertstudy}

To evaluate {\systemname}, we conducted a series of structured interviews with six ML experts (E1 - E6), who were recruited by e-mail. 
They are not coauthors of this paper and have not previously seen {\systemname}. 
The panel consisted of industry ML practitioners (researchers and engineers), with an average age of $33.167 \pm 5.707$ years, and varying STEM education backgrounds: four holding doctoral degrees, one master's degree, and one bachelor's degree in Computer Science. All experts have worked in the field of Advanced Driver Assistance and Autonomous Driving Systems (ADAS/AD), and have had prior experience in developing object detection models. 
Collectively, these six experts possess an average of $6.667 \pm 3.077$ years of  experience in the field. 

\para{Procedure.}
Experts were asked to analyze edge cases for four object detection models: ``car,'' ``person,'' ``chair,'' and ``dog''.  
We divided the experts into two groups to evaluate both {\systemname}'s effectiveness and, exclusively, the explanation quality.   
The first group used {\systemname} to analyze ``car'' and ``person'' detectors and rated textual explanations for ``chair'' and ``dog'' on a 5-point Likert scale, while the second group did the reverse.  
During the practical part of the study, experts received a ten-minute tutorial, explored the system for thirty minutes using the ``think-aloud'' protocol, and completed a post-study questionnaire.  

This section is organized as follows: we first describe the model and data used in the interviews. Then, we describe three use cases derived from the expert study, where our system was used to inspect the detectors. Finally, we summarize the experts' feedback.

\subsection{Model and Data Description}

We use the PASCAL image dataset~\cite{pascal-voc-2007} and select four objects from differing categories with varying sizes: person (Person), dog (Animal), car (Vehicle), and chair (Indoor). 
For each object, we first extract images containing it, and then split them into training and validation sets with a ratio of $60\%$ to $40\%$, respectively. 
We train a detector for each object using a popular object detection model, Fast R-CNN~\cite{girshick2015fast}, and end the training process when the loss falls below $0.05$ or after $40$ epochs. 
Then, we obtain detections (TPs, FPs, and FNs) of the validation set for each detector, and execute {\systemname} once to generate data slices, slice explanations, and integrate them into the visual analytics system. 
We report the statistics for each detector in~\cref{tab:pascal-data}.

\begin{table}[!ht]
\fontsize{7}{9}\selectfont
\begin{center}
\vspace{-3pt}
\resizebox{\columnwidth}{!}{
    \begin{tabular}{ c|c|c|c|c|c|c } 
    \ChangeRT{0.8pt}
    \textbf{Object} & \textbf{Ground truth} & \textbf{mAP} & \textbf{TPs} & \textbf{FPs} & \textbf{FNs} & \textbf{$\#$ Slices}   \\ 
    \ChangeRT{0.8pt}
    
    \textbf{Dog} & 1227 & 86.71\% & 541 & 165 & 64 & 7  \\ 
    \ChangeRT{0.5pt}
    \textbf{Car} & 2364 & 75.56\% & 752 & 186 & 209 & 13   \\ 
    \ChangeRT{0.5pt}
    \textbf{Chair} & 2906 & 47.83\% & 668 & 481 & 530 & 15 \\ 
    \ChangeRT{0.5pt}
    \textbf{Person} & 10129 & 75.42\% & 3217 & 2066 & 808 & 24  \\ 
    \ChangeRT{0.8pt}
    \end{tabular}
}
\vspace{-7pt}
\caption{Detector statistics: ground truth size, mAP, sizes of TPs, FPs, FNs on the validation set, and the number of output slices.}
\label{tab:pascal-data}
\end{center}
\vspace{-8pt}
\end{table}

\subsection{Use Case 1: Slice Analysis of a Car Detector} 
\label{sec:usecase1}
E1, E3, and E5 would like to identify systematic errors in a car detector in order to write a validation report. 
{\systemname} identified $13$ data slices (\cref{tab:pascal-data}), some of which are shown in~\cref{fig:interface}.

\para{Slice understanding}. 
When examining slice 1, all experts recognize the FPs and FNs mainly occur at the airport. E3 notices that this slice is distant from others in the density plot view. 
The textual explanation (\cref{fig:slice-examples}) further states that FPs are caused by misclassification of ``trucks'' or ``airplanes'' in airports as cars, whereas FNs are caused by trucks or luggage carts obstructing cars. The experts agree with the explanation but provide additional causes. 
For example, E3 identified some labeling issues: \emph{``The model is identifying a car, but the labeler did not.''}

\para{Slice refinement.} 
E5 investigates slice 2 highlighted in~\cref{fig:interface}, and understands the primary cause is misclassification or occlusion caused by buses.
E5 observes several FNs incorrectly labeling trucks as cars (\cref{fig:interface}-B4). 
E5 believes the mislabeling issue in the data needs to be fixed and thus saves several truck images. 
To refine this slice, the user searches for \emph{``errors caused by buses''} (\cref{fig:interface}-E), adjusts the range bar to exclude trucks and clicks the ``Replace'' button to replace the original slice.  
After clicking the ``Edit'' button, the user may change the keyword ``red bus'' to ``bus'' and modify explanation slightly for a more accurate description (\cref{fig:interface}-F).

\para{Slice navigation.} 
E1 sorts slices (\cref{fig:interface}-A1) by precision and notices that slice 3 has a low precision ($0.12$), indicating FPs are likely to occur.
The slice examples and explanations are shown in~\cref{fig:slice-examples}, suggests that the detector often misclassified motorcycles as cars, or that motorcycles obstructed the car. 

\begin{figure}[!ht]
  \centering
  \vspace{-5pt}
  \includegraphics[width=0.8\columnwidth]{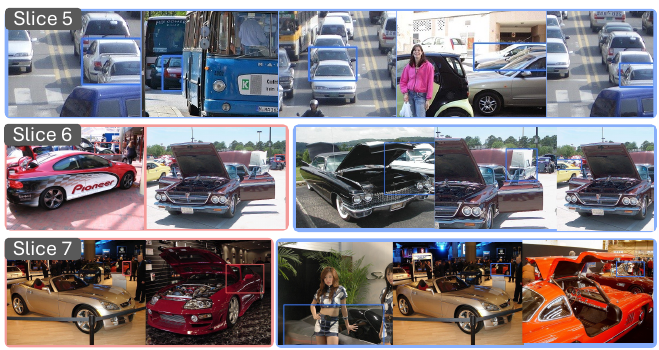}
  \vspace{-5pt}
  \caption{Car detector: \textcolor{myred}{\textbf{FPs}} and \textcolor{myblue}{\textbf{FNs}} in slices 5, 6, and 7.}
  \vspace{-10pt}
  \label{fig:slice3-5}
\end{figure}

Through iterative analysis, E1 saves more slices of interest (\cref{fig:slice3-5}). 
For example, slice 5, with the largest number of FNs, mainly occurs on streets with overlapping cars.
The error is likely caused by low labeling quality, as described by the explanation, \textit{``the ground truth car is partially obstructed by other objects such as buses, cars, buildings, fences, mirrors, windows.''} 
Slices 6 and 7, spatially close in the plot view, mostly appear in indoor and outdoor commercial scenes such as exhibitions, typically with open car hoods. 
These slices present unusual car states, and contain out-of-distribution examples where the detector fails.

\begin{figure}[!ht]
  \centering
  \vspace{-5pt}
  \includegraphics[width=0.9\columnwidth]{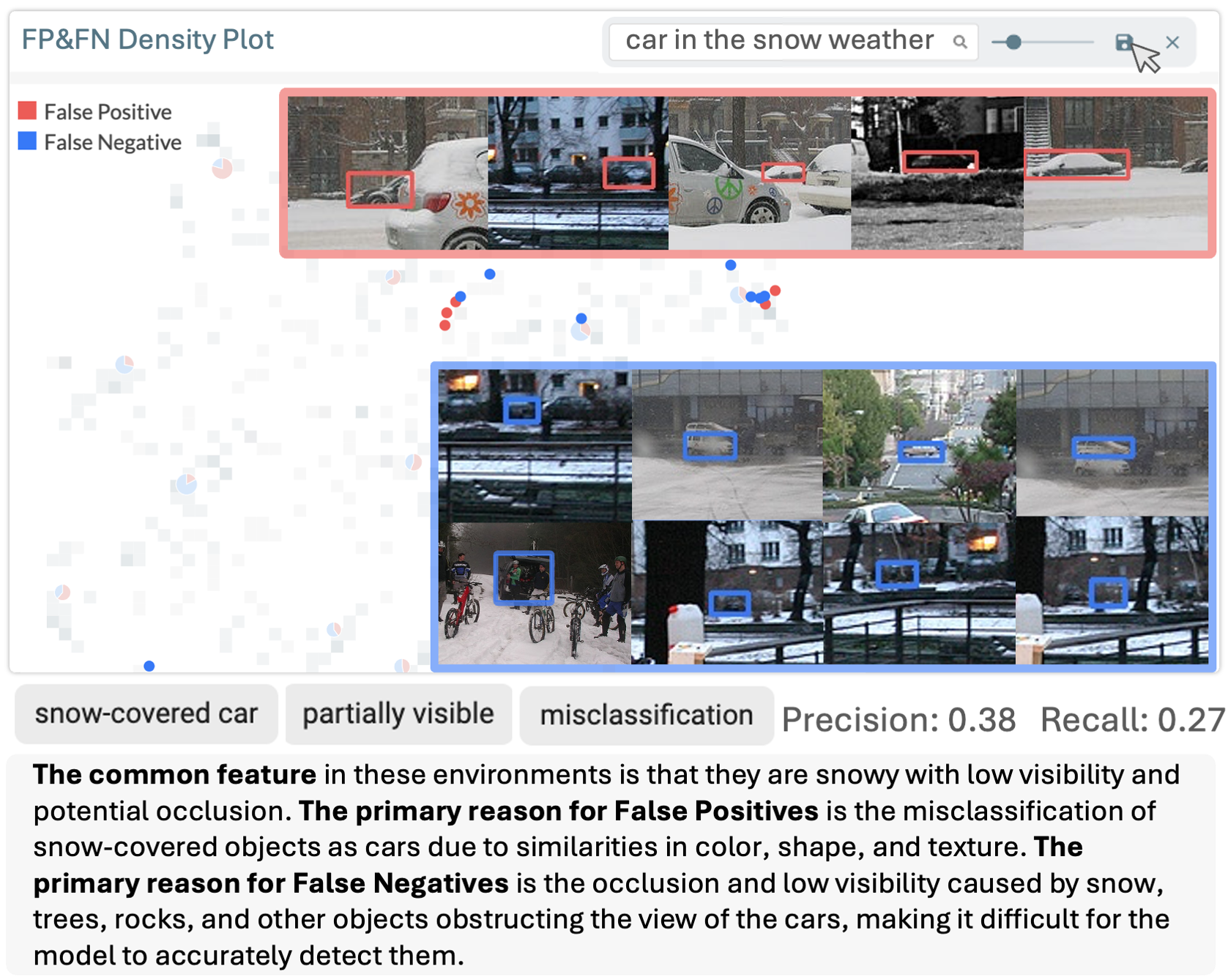}
  \vspace{-5pt}
  \caption{Hypothetical slice in snowy weather created by queries.} 
  \vspace{-15pt}
  \label{fig:snow-slice}
\end{figure}

\para{Hypothesis testing.} 
While investigating the data slices, E5 finds some images containing snow, and wants to investigate whether snow negatively impacted the car detection.
To test their hypothesis, as shown in~\cref{fig:snow-slice}, E5 globally queries \textit{``Car in the snow weather''} and then adjusts the slider to include $5$ FPs and $8$ FNs to form a hypothetical slice that has a precision of $0.38$ and a recall of $0.27$. 
From the instances and explanations, E5 deduces that cars covered by snow or tree branches, in low visibility, can impact the model performance. 

\begin{figure*}[!ht]
  \centering
  \vspace{-3pt}
  \includegraphics[width=0.81\linewidth]{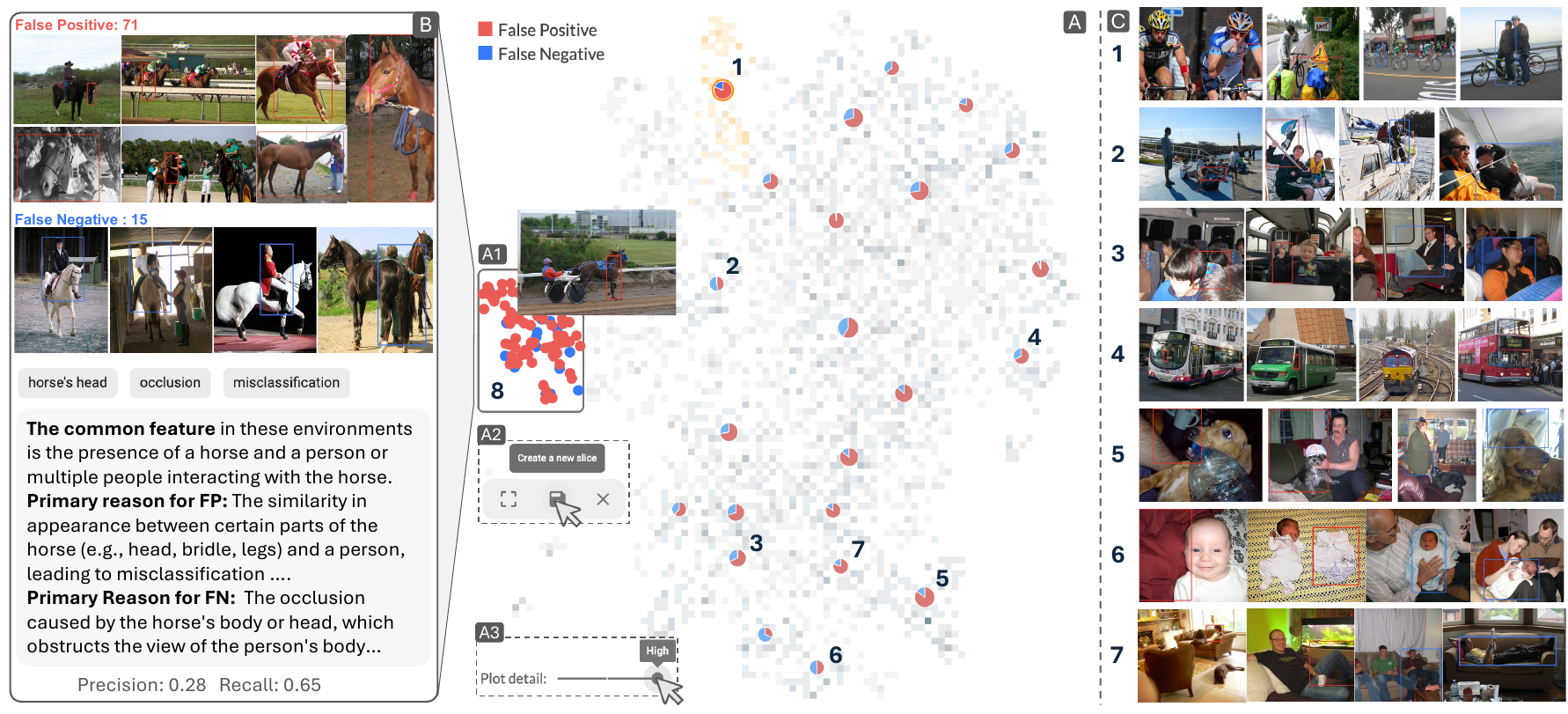}
  \vspace{-7pt}
  \caption{Illustration of the slices in the person detector used in use case 2. \textbf{(A)} The density plot view with a high plot detail (A3). The area A1 is brushed to form a new slice (A2). \textbf{(B)} The new slice information, such as a sample of False Positives (FPs) and False Negatives (FNs), and explanations. \textbf{(C)} Shows two FPs and two FNs for each of slices 1-7. The corresponding slice numbers (1-8) are annotated in (A).}
  \vspace{-13pt}
  \label{fig:usecase-2}
\end{figure*}  

\subsection{Use Case 2: Improving a Person Detection Model} 
\label{sec:usecase2} 

The same experts, E1, E3 and E5, also use {\systemname} to analyze a person detection model, as shown in ~\cref{fig:usecase-2}.  
After adjusting the plot level of detail for a more granular inspection of the embedding (\cref{fig:usecase-2}-A3), E3 notices a dense cluster on the left side of the view that the algorithm does not identify as a slice. 
Using the brushing tool, E3 observes numerous images featuring horses and persons (\cref{fig:usecase-2}-A1) and posits it as a missing slice. 
Thus, he creates a new slice for these predictions to validate his hypothesis (\cref{fig:usecase-2}-A2). 
This new slice, shown in~\cref{fig:usecase-2}-B, has a low precision of $0.32$, and the explanation suggests the failure reason to be misclassification or occlusion caused by horse features such as heads and legs.  This slice is then saved for further analysis.  
E1, E3, and E5 also identify other critical slices based on their experience, and all saved slices (1-8) are shown in~\cref{fig:usecase-2}. 
These slices typically have coherent explanations and are caused by occlusion or inaccurate detections in diverse scenarios such as bicycle (1) and boat (2) scenes, people inside vehicles (3, 4), infants (6), and indoor scenarios (5, 7).
 
Based on the feedback from the users, we fine-tune the model with a focus on these eight slices.
The strategy is to retrieve images from training data that resemble the images in these slices, and then use them to train the model with one more epoch.
Specifically, we employ two approaches to extract images.
The first is embedding-based, where we export the slice images from {\systemname} and convert them into embeddings through the encoder of BLIP-2. 
Then, we compute the embedding center and select the training images whose embeddings are closest to the center.
The second approach uses CLIP scores~\cite{Radford2021}, which measure the alignment between images and text descriptions and have been widely used in image retrieval~\cite{johnson2023does}. 
Following Johnson et al.~\cite{johnson2023does}, we use the template \textit{``A photo of [...]''}, together with the slice description generated by {\systemname}.  
Then, we extract training images that have higher CLIP scores with the slice description.
For each approach, we curate a new training set.

The model is then trained for one more epoch using the re-weighting technique~\cite{idrissi2022simple} for worst-group optimization: we select three times the number of images for each slice, combine the original eight slices, and replicate  them to match the size of the original training set. The retraining is done on three datasets: the original dataset, the embedding-based augmented dataset, and CLIP score-based training dataset. We refer to three conditions as \textit{Original}, \textit{Embedding}, and \textit{CLIP Score}.
We then validate each model and recalculate the slice metrics (precision and recall) and model metrics (precision, recall, and mAP).
Slice metrics are approximated by considering all objects in these images. 
Overall, across all three conditions, the \textit{Embedding} and \textit{CLIP Score} conditions demonstrate the best trade-off between slice precision and recall, achieving comparable mAP values of $0.7878$ and $0.7845$, respectively---approximately $3\%$ higher than the baseline mAP of $0.7542$ before fine-tuning. In contrast, the \textit{Original} condition yields a mAP of $0.7265$, falling below the baseline.
Notably, the \textit{CLIP Score} condition exhibits the best slice-level performance, with $5$ slices showing the biggest improvements in precision and another $5$ slices showing the biggest improvements in the recall. This might be due to the fact that \textit{CLIP Score} retrieves more diverse images than \textit{Embedding}. We offer the detailed results in the supplement.

\subsection{Use Case 3: Analysis of a Dog and Chair Detector} 
E2, E4, and E6 analyze the detectors for the classes ``dog'' and ``chair''. Here, we summarize their insights regarding the model's edge cases. 
\textbf{Dog detector}: All experts quickly observe that the model frequently confuses dogs with other animals. E2 notes, \emph{``The explanations are really good: sheep and cows recognized as a dog.''} Additionally, E4 and E6 identify a recurring issue when a person is playing with a dog, where occlusion by the person often causes bounding box detection errors. 
\textbf{Chair detector}: All experts agree that occlusion by people sitting on chairs is the primary cause of most model failures. As E4 explains, \emph{``An obvious thing would be people sitting on chairs.''} However, other factors also affect detection performance. A noteworthy observation is  that chairs with unique designs are often missed. E2 remarks, \emph{``The false negative cases are chairs that look different and have some different kind of design.''} Furthermore, E6 identifies couches as a frequent source of mistakes. After querying for \emph{``couches,''} he suggests  that many false positives could be attributed to inconsistent labeling.

\subsection{Expert Feedback} 
Experts have praised {\systemname}, noting its user-friendliness and functionality. 
E1 highlights the system's ability to provide insightful findings from edge cases, enhancing hypothesis testing: \emph{``The initial findings from various edge cases provide valuable insights into failure scenarios. Delving into these groups has inspired me to uncover further reasons, enabling easy testing of my hypotheses.''}
Furthermore, E2 and E5 appreciate the quick access to detailed and organized data slices, which facilitates a deeper understanding of the model. E2 mentions that \emph{``the slices are clustered nicely and are visually and semantically consistent''}.
The system's intuitive user interface is praised by E3: \emph{``It is simple to use and pretty easy to adapt to.''} 
E4 and E6 remark on the flexibility and capabilities of the tool for exploring the model's mistakes. E6, in particular, highlights the system is a \emph{``very intuitive tool to understand the edge cases and hone in on them based on the explanations provided''}.
Overall, {\systemname} has been recognized for its effective design and functionality, aiding users in efficiently navigating and analyzing complex data scenarios.

{\systemname} has generally been well-received, but it has also drawn some constructive criticism from experts. 
E1 and E5 observe that the explanations provided by the system can be superficial and may fail to address the specific reasons behind the edge cases. 
E2 and E4 raise concerns about limitations due to dataset quality, such as inconsistent labeling. 
However, they also note that the {\systemname} could be used to identify these inconsistencies and improve data label quality. 
E5 and E6 suggest enhancements to the user interface, particularly for image navigation, such as zooming and panning, and displaying ground truth and predictions separately to better support detailed examination of the data. 
These insights shed light on areas for potential refinement to enhance the system's capabilities and user experience. 
Experts have also evaluated the {\systemname} tool and explanations using a 5-point Likert scale. The results reflect a positive overall impression of {\systemname}; see the supplement for details. 

\para{Explanation evaluation.}   
\begin{figure}
    \centering
    \vspace{-1pt}
    \includegraphics[width=1\linewidth]{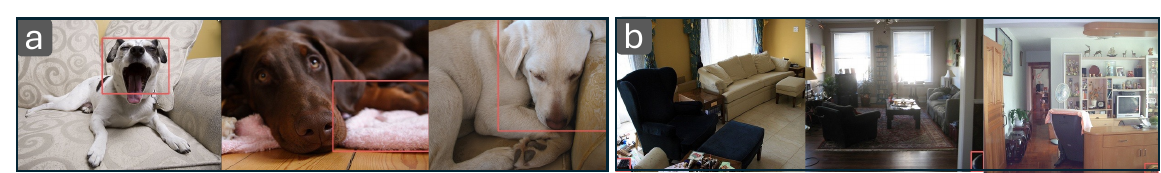}
    \vspace{-15pt}
    \caption{Examples of two slices with low-quality FP explanations. (a) a dog detector. (b) a chair detector.}
    \vspace{-15pt}
    \label{fig:badExplantions}
\end{figure} 
We further analyze explanations with strong disagreement or average scores below neutral. 
One notable issue is model hallucinations: 
as shown in~\cref{fig:badExplantions}, the explanation of slice (a - dog detector) states, \textit{``model mistaking objects or elements in the scene, such as towels, pillows, patterns, or furniture, as part of the dog''}, whereas partial detection and occlusions of towels or pillows are more likely the reason.
However, we find most individual explanations are sound, frequently mentioning \textit{``only part of the dog is detected''} and surroundings of \textit{``towel, couch, and pillow''}.  
We speculate that LLM incorrectly connects these elements. 
Additionally, we notice that {\systemname} struggles to interpret small detection: ~\cref{fig:badExplantions} slice (b - the chair detector) is wrongly explained as \textit{``the model mistaking objects or textures in the scene, such as fabric, spheres, reflective surfaces, or wigs, as chairs''}.
We attempted to refine the prompt by including the ratio of the detection region to the image, expecting the explanation would mention size-related challenges. However, we found that the model is often insensitive to such a ratio, misinterpreting a tiny detection as a large portion of the image. 
An alternative is to explicitly show the detection size in the system to inform experts, which we leave for future work.

\subsection{Comparison with Other Slice Discovery Systems}

\begin{figure}[!ht]
    \centering
    \includegraphics[width=1\linewidth]{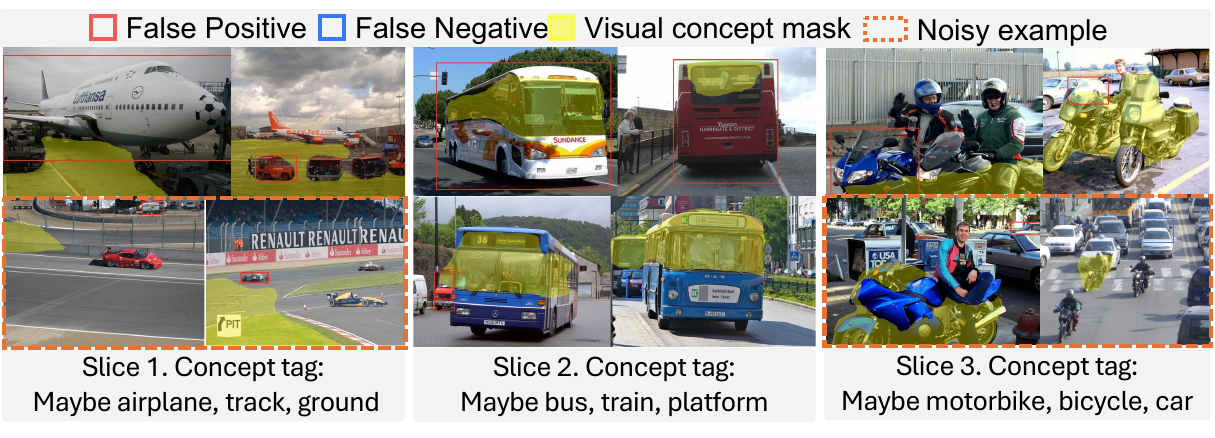}
    \vspace{-20pt}
    \caption{Samples (\textcolor{myred}{FPs} and \textcolor{myblue}{FNs}) of three slices in the car detector identified by {\ConceptSlicer}. The relevant visual segments are masked in yellow. Spurious segments that are irrelevant to the detection are highlighted with dotted boxes.}
    \label{fig:comparion-slices}
\end{figure}

The most relevant work to ours is {\ConceptSlicer}~\cite{zhang2024slicing}, which identifies slices in object detectors using visual concepts.  
We run {\ConceptSlicer} on the car detector (Sec. 6.2) and identify 9 slices with accuracy at least 5\% below average. We then remove visually inconsistent slices and merge highly overlapping ones caused by concept quality issues (e.g., a single label for different object types or multiple labels for the same object). \Cref{fig:comparion-slices} shows samples from the resulting three slices, which align with the three slices identified by {\systemname}: airplanes (\cref{fig:slice-examples}-1), buses (\cref{fig:interface}-2), and motorcycles (\cref{fig:slice-examples}-3), respectively.  

We observe that {\systemname} often produces finer-grained slices.  
For example, {\ConceptSlicer} groups all motorbike-present images regardless of detection relevance (\cref{fig:comparion-slices}-3), while {\systemname} isolates motorbike misclassification or occlusion cases, e.g., motorbike/person interactions (\cref{fig:slice-examples}-3). Similarly, {\ConceptSlicer} merges distinct but visually similar scenarios like ``airport runways'' and ``race tracks'' (\cref{fig:comparion-slices}-1), whereas {\systemname} separates them (\cref{fig:slice-examples}-1).  Moreover, {\systemname} identifies slices missed by {\ConceptSlicer} due to its reliance on visual concepts. For instance, slice 5 (\cref{fig:slice3-5}, ``car overlap'') and slices 6 and 7 (missing concepts like ``exhibition'' and ``car hood'') are absent in {\ConceptSlicer}. Abstract factors like image quality and weather are also overlooked by {\ConceptSlicer}, missing slices for small detections and snowy conditions (\cref{fig:snow-slice}).  
These examples show that visual concepts alone struggle to capture key features like image context (e.g., location, environment) and object relationships (e.g., proximity, positions). {\systemname} bridges this gap by leveraging foundation models to explain detection mistakes.  

We also compare {\systemname} with {\AttributionScanner}~\cite{xuan2025attributionscanner}, a metadata-free method using attribution-weighted embeddings to detect slices in image classifiers. Adapting {\systemname} to the landbird/waterbird classification task, we found 7 problematic slices for {\AttributionScanner} and 5 for {\systemname}. Our results suggest {\AttributionScanner} focuses on spurious background correlations, while {\systemname} captures and explains nuanced patterns, revealing model vulnerabilities to bird species, scenes, spurious correlations, and labeling issues. Further details are provided in the supplement.  

\section{Discussion and Future Work}
\label{sec:discussion}

\para{Improving the slice discovery method.} 
In {\systemname}, the context region is obtained by enlarging the detection window twice.
Although this strategy aligns with the previous work~\cite{zhu2017couplenet}, it may still capture insufficient context, particularly for small detections, resulting in inconsistent slices. A cropping strategy that prioritizes detection while maintaining sufficient context is needed.
Furthermore, enabling user control of UMAP and HDBSCAN parameters could be a future enhancement~\cite{johnson2023does}, but it must carefully balance added complexity and computational cost.

\para{Enhancing the quality of slice explanations.}  
We break down the slice explanation task into three steps (\cref{sec:slice-explanation}): individual explanations, noise reduction, and explanation aggregation.  
We leverage prompt engineering~\cite{chen2023unleashing} (e.g.,~chain-of-thought reasoning, role-playing) and integrate knowledge of common detection errors. While these improve explanation quality, model hallucinations may still cause ungrounded explanations, such as confusing occlusion with misclassification or misinterpreting tiny detections. 
To address this, future work includes user-modifiable prompts, golden explanations as few-shot examples~\cite{dong2022survey}, and TPs as contrastive examples~\cite{ross2021explaining}.  
Additionally, predefined questions (\cref{tab:predefine-questions}) may not generalize; e.g., the ``weather'' question (Q3) is unsuitable for indoor settings. Allowing experts to tailor core questions to applications is another way to improve explanations.

\para{Improving the visual analytics system.} 
The {\systemname} interface helps users extract insights that are otherwise difficult to obtain. For example, as shown in~\cref{sec:usecase1}, the density plot (\cref{fig:interface}) allows experts to identify isolated slices (slice 1), detect noise (2), and recognize visually similar slices (6, 7) by analyzing cluster density and distribution, enhancing slice inspection. Hypothesis testing via query (\cref{fig:snow-slice}) and visual selection (\cref{fig:usecase-2}) further aids model validation by uncovering previously unnoticed slices. 
Inspired by expert feedback and recent research, we plan to enhance our system by (1) displaying ground truth for incorrect detections, enabling the identification of mislabeled data, and (2) expanding the search space beyond the validation set (e.g., via image retrieval) to improve hypothesis testing and enable model refinements with additional training data~\cite{boreiko2023identifying, wilesdiscovering, zhang2024slicing}. 

\para{Extending to other vision tasks.} 
We plan to generalize {\systemname} to tasks like semantic segmentation by using segment contours for the future.  
However, defining context and intersection regions for irregularly shaped segments requires careful consideration.

\section{Conclusion}
\label{sec:conclusion}
We present {\systemname}, an XAI framework that leverages VLMs and LLMs to validate computer vision models in a human-in-the-loop manner.
{\systemname} automatically discovers and explains data slices, particularly in object detectors, and empowers experts to efficiently explore slices and interactively test hypothetical slices through a visual analytics system. 
Additionally, this framework is model-agnostic and can continually benefit from the ongoing advancement of foundation models.
%Through three use cases, we demonstrate its capability to interpret diverse slices and enhance model robustness.
%Feedback from six computer vision experts validates its efficacy and sheds light on challenges and opportunities in using foundation models for slice explainability.

\noindent\textbf{Acknowledgments.} 
This work started when X. Yan, X. Xuan, and L. Gou worked
with Bosch Research North America.  
It was partially supported by NSF grants IIS-2205418 and DMS-2134223, and a seed grant from the Utah Board of Higher Education's Deep Technology Initiative.

%---------------------------------

\bibliographystyle{eg-alpha-doi} 
\bibliography{main}


\section*{Supplementary material}

In this supplement, we first discuss visualization design alternative (Appendix A). We then provide the complete prompt templates used for GPT and LLaVA in the process of explanation generation (Appendix B). We also provide all slices and their explanations for the four detectors (car, dog, chair, and person) evaluated in our work, as well as a sample of individual detection explanations from which the slice explanations are derived (Appendix C). We include details on clustering in slice discovery (Appendix D), filtering techniques in slice explanations (Appendix E), experimental and implementation details (Appendix F), and Use Case 2 (Appendix G). We also include expert ratings on *VISLIX* (Appendix H) and a detailed comparison with *AttributionScanner* (Appendix I).

**Appendix A:** Visualization Design Alternative

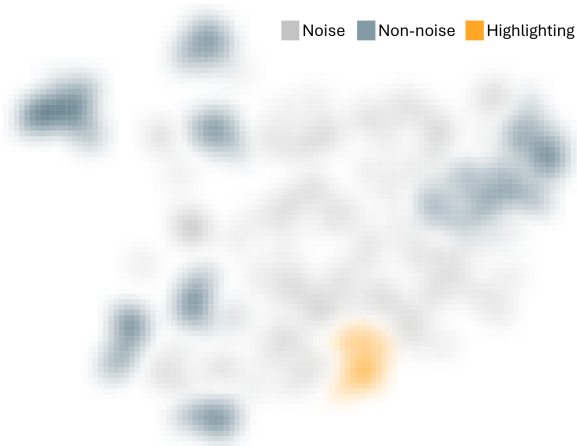

**Figure 1:** *A 2D density plot based on kernel density estimate.*

Fig. 1 presents a design alternative to the 2D histogram-based density plot (View B1) that we have explored during VISLIX's development. This version, based on a Gaussian 2D kernel density estimate, provides a continuous and smoothed representation of the data distribution. We initially chose this approach for its visual smoothness, compared to the discrete blocks of a 2D histogram. However, we found that this design could be misleading, as it may still render regions without actual instances, which is particularly problematic when using brushing operations. The iteration indicates that, when addressing domain-specific questions, rule-of-thumb design principles might be misaligned with expert needs, underscoring the importance of continuous feedback throughout the design process to ensure accurate and effective visual communication.

**Appendix B:** Prompt Templates

**Individual explanations.** Fig. 2 shows the prompt template for asking a question to LLaVA about an image, which is used in answering both predefined questions and questions generated by GPT. Fig. 6 shows the template used to prompt GPT to ask a question about the context region of a False Positive, bridging the GPT-LLaVA conversation. We start by describing the detection background using answers to predefined questions, then task GPT with generating a new question. Specifically, we organize several predefined questions, such as those about weather and lighting conditions, and their answers as a dialog before the task description, serving as in-context learning examples. Once GPT generates a new question, we send it along with the context region to LLaVA to collect the answer. We then append the question and answer to the end of the dialog and prompt GPT to generate another question. This iterative process continues for 10 rounds or until GPT outputs "STOP". This template applies to False Negatives after slight modifications, as shown in Fig. 7.

After collecting answers to predefined and GPT-generated questions, we use the content from Fig. 6 to prompt GPT to generate the individual explanation (Fig. 3).

**Slice explanations.** Fig. 4 displays the prompt template for GPT to generate a slice explanation, based on individual explanations of FPs and FNs within this slice. Similarly, Fig. 5 exhibits the prompt of generating slice keywords using the same input but different tasks.

**Appendix C:** Slice and Individual Detection Explanations

**Slice explanations.** We provide all slices and their explanations generated by our framework corresponding to four object detectors analyzed in the expert study. These include the car detector, dog detector, chair detector, and person detector. Specifically, for each detector, each slice showcases 10 random samples of False Positives (FP) and False Negatives (FN), accompanied by its textual explanations, as with the format used in the expert study.

**Individual detection explanations.** We also present sample explanations for individual FPs and FNs from the four object detectors. Explanations that are clearly incorrect are highlighted in red.

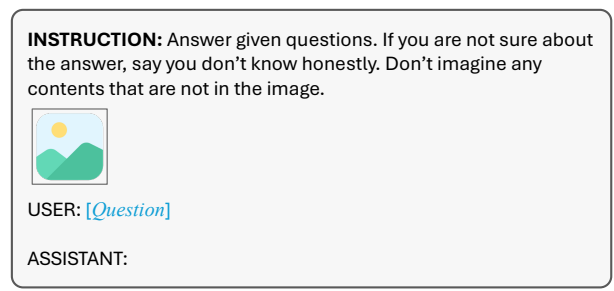

**Figure 2:** *Prompt template used for LLaVA to answer a question about an image.*

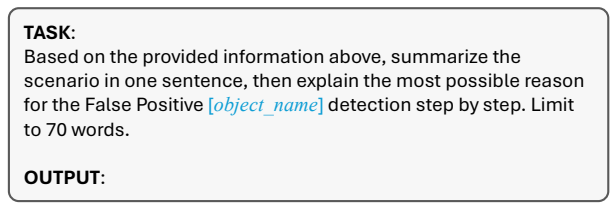

**Figure 3:** *Task description used for GPT (Interpreter) to explain a False Positive detection.*

> **INSTRUCTION:** In a [*object_name*] detection model, a set of False Positives and False Negatives occurs. For each False Positive and False Negative, you are given sentences describing the scene and the error in a single row. Your task is to provide three keywords from these descriptions, which summarize these environments and the most primary reason of these errors.
>
> - Be concrete and explicit.
> - You think step by step.
> - Each keyword has at most 3 words.
> - You only output three keywords separated by ','.
>
> **INPUT**:
> [*Individual explanations for False Positives*]
> [*Individual explanations for False Negatives*]
>
> **OUTPUT**:
> <Keyword1>,<Keyword2>,<Keyword3>

**Figure 4:** *Prompt templates of GPT (Summarizer) for creating slice explanations.*

> **INSTRUCTION:** In a [*object_name*] detection model, a set of False Positives and False Negatives occurs. For each False Positive and False Negative, you are given sentences describing the scene and the error in a single row. As a Computer Vision expert, you concisely summarize these errors for those who haven't observed these pictures. In your output, you first describe the common feature of these environments, then identify the most primary reason of False Positives and False Negatives.
>
> - You think step by step.
> - Be concrete and explicit.
> - You ONLY output the most likely reason and ignore other reasons.
> - Be succinct and limit your output within 70 words.
>
> **INPUT**:
> [*Individual explanations for False Positives*]
> [*Individual explanations for False Negatives*]
>
> **OUTPUT**:

**Figure 5:** *Prompt templates of GPT for creating slice keywords.*

**Appendix D:** Details on Clustering in Slice Discovery

Our slice discovery method identifies data slices by first applying dimensionality reduction using UMAP, followed by clustering with HDBSCAN. The UMAP parameter setup largely follows prior work [MSRPC21, SDA*20], including setting the minimum distance between points to 0 for a more precise representation of the data. However, McConville et al. [MSRPC21] sets the number of dimensions, $k$, to the number of distinct labels in the dataset, which is not applicable to our data. Instead, we have experimented with $k$ values ranging from 2 to 10 and have observed no significant differences in the resulting slices when applying clustering methods. To preserve as much information as possible, we empirically set $k$ to 10. Additionally, we fine-tune two parameters via grid search: the number of neighbors considered in UMAP and the minimum number of points required to form a core point in HDBSCAN. The optimal parameter combination is selected using the Silhouette metric [Rou87] that is widely used to measure the clustering quality.

**Appendix E:** Centrality-Based Filtering for Slice Explanations

To improve the quality of slice explanations, we adopted a centrality-based strategy to filter out noisy individual explanations within a slice. These explanations often arise from VLM/LLM hallucinations or suboptimal parameter configurations during slice discovery. Our approach involves converting each individual explanation into a sentence embedding and selecting explanations closest to the embedding center until 80% are retained or the token limit of 2000 is reached. Our preliminary experiments show that this strategy produces higher-quality and more consistent slice explanations compared to random selection.

For instance, in Fig. 8, we present a slice of false positives (FPs) from a car detector along with explanations generated using centrality- and random-based strategies. This slice primarily involves buses, with some focusing on trucks. Images are sorted by descending centrality based on the sentence embeddings of their explanations. Additionally, we highlight the individual explanations of the 5 most central FPs (1–5) and the 5 least central FPs (19–23), which are filtered out by the centrality-based method. The most central explanations typically include bus misclassifications and are accurately described, while the least central ones often contain noise or errors—such as truck misclassification, excessively long explanations, or hallucinations, highlighted in grey text. The centrality-based method thus provides a more focused and concise slice explanation by describing primarily buses.

**Appendix F:** Experimental and Implementation Details

We conducted experiments related to the VLMs, including BLIP-2 and LLaVA, on a Linux system equipped with an Intel Core i7-5930K CPU (12 cores, 3.50 GHz), 128 GB of RAM, and one Nvidia Titan X GPUs with 12 GB of memory. The GPT3.5 API was accessed through the LangChain library (version 0.2). For the visual analytics system, we utilized JavaScript, D3 (version 7.8.5), Material UI (version 6.1.9), and the React framework(version 18.3.1) for the frontend, while the backend was developed using Python (version 3.9) and Flask (version 3.0.3).

As shown in Tab. 1, for each evaluated object, we report the total time spent on generating context-aware embeddings, performing slice discovery, and generating all individual explanations. All of these data are precomputed before being integrated into the visual analytics system. Only the slice explanation is generated interactively when users test hypothetical slices. On average, the time to generate a slice explanation is 1.70 seconds (SD = 0.25).

| Object | Embeddings | Slice discovery | Individual explanations |
|---|---|---|---|
| Dog | 1 min, 39.12 s | 4 min, 15.12 s | 1 h, 4 min, 28.60 s |
| Car | 2 min, 59.49 s | 4 min, 23.24 s | 1 h, 58 min, 45.59 s |
| Chair | 6 min, 49.77 s | 5 min, 52.80 s | 4 h, 55 min, 36.25 s |
| Person | 17 min, 13.64 s | 12 min, 45.40 s | 14 h, 35 min, 52.04 s |

**Table 1:** *The total runtime for context-aware embedding generation, slice discovery, and individual explanation generation for each dataset.*

**INSTRUCTION:**
You are an expert in Computer Vision, specializing in analyzing the reasons behind False Positive (FP) detections in object detection models. You are familiar with key concepts in this domain, including bounding box, ground truth, among others.

Your expertise covers various factors contributing to False Positives, including but not limited to, occlusion, size, visibility, viewpoint, localization error, background, spurious correlation, missing or inaccurate ground truth, and so on.

In the context of a *[object_name]* detection model, a False Positive detection of a *[object_name]* has occurred, which is depicted by a bounding box. The size of the detection region is *[width]* x *[height]* pixels.
The description of the False Positive detection region is: < *[Q1-detection_caption]* >.
Only parts of the ground truth *[object_name]* are detected, the detected parts are: < *[Q6-intersection_caption]* >. *(Apply to partial detection)*
The detection region fully encloses the ground truth *[object_name]*, a description of the ground truth is:< *[groundtruth_caption]* >. *(Apply to over detection)*

Furthermore, there's a cropped image that encompasses the False Positive detection and its surroundings, which is described as: < *[Q2-context_caption]* >.
Given the information provided above, ask me questions about the cropped image to maximize your information of reasoning about the False Positive. Don't mention "False Positive" in your question. Carefully asking me specific and informative questions. Each time ask only one question without giving an answer. I'll put my answer beginning with "Answer:".

**Q&A about the image:**
Question: What is the weather in this image?
Answer: *[Q3-weather]*
Question: How is the lighting condition in this image?
Answer: *[Q4-lighting]*
Question: Is there any obstruction blocking the *[object_name]*? If so, what is obstructing the *[object_name]*? *(Apply to IoU>0.2)*
Answer: *[Q5-obstruction]*

**TASK:**
 Next Question(You can still ask *[count]* questions. Please output "STOP" when no questions is needed). Question:

**Figure 6:** *Prompt template used for promoting GPT (Questioner) to raise a question for a False Positive detection.*

|  | Baseline | | Original-Train set | | Embedding-Train set | | CLIP Score-Train set | |
|---|---|---|---|---|---|---|---|---|
|  | Precision | Recall | Precision | Recall | Precision | Recall | Precision | Recall |
| Slice 1 | 0.5471 | 0.7881 | 0.6483 | 0.7966 | 0.6452 | 0.8475 | 0.6821 | 0.8729 |
| Slice 2 | 0.6071 | 0.7083 | 0.6282 | 0.6806 | 0.6105 | 0.8056 | 0.6129 | 0.7917 |
| Slice 3 | 0.5196 | 0.6883 | 0.5543 | 0.6623 | 0.5729 | 0.7143 | 0.5895 | 0.7273 |
| Slice 4 | 0.3879 | 0.6 | 0.4565 | 0.56 | 0.3712 | 0.6533 | 0.3952 | 0.6533 |
| Slice 5 | 0.3907 | 0.7763 | 0.381 | 0.7368 | 0.5214 | 0.8026 | 0.537 | 0.7632 |
| Slice 6 | 0.5775 | 0.7069 | 0.6338 | 0.7759 | 0.6184 | 0.8103 | 0.6912 | 0.8103 |
| Slice 7 | 0.4146 | 0.7556 | 0.4684 | 0.8222 | 0.5362 | 0.8222 | 0.5968 | 0.8222 |
| Slice 8 | 0.4188 | 0.7273 | 0.3507 | 0.6727 | 0.4343 | 0.7818 | 0.4286 | 0.7636 |
| All | 0.6089 | 0.799 | 0.6071 | 0.7777 | 0.6304 | 0.8325 | 0.6551 | 0.8271 |
| mAP | 0.7542 | | 0.7265 | | 0.7878 | | 0.7845 | |

**Table 2:** *Metrics on slices (Precision and Recall) and model (Precision, Recall, and mAP) across baseline and three training sets (Original, Embedding, CLIP Score). Values below baseline are gray. The highest values for each row are blue.*

**Appendix G:** Details on Use Case 2

In the second use case, we fine-tune the person detector for an additional epoch using the original training data (*Original*), training data retrieved from image embeddings of the selected eight slices (*Embedding*), and training data retrieved from the textual descriptions of those slices (*CLIP Score*). The results for the three conditions are shown in Tab. 2, where the baseline represents the detector before fine-tuning; metric values lower than the baseline are colored gray, and the highest values for each row are colored blue. For the eight slices' metric values, the *Original* condition shows the most declines, with 8 of 16 values decreasing. In contrast, the *Embedding* and *CLIP Score* conditions each have only 1 decrease. Besides, each condition possesses 3 (*Original*), 6 (*Embedding*), and 10 (*CLIP Score*) highest metric values. Although higher precision or recall values are generally desirable, they must be balanced to ensure the model's overall performance, reflected by the mAP value. The *Original* condition, despite having the 3 highest metric values, has an mAP of 0.7265, lower than the baseline of 0.7542. Yet, the *Embedding* and *CLIP Score* conditions achieve similar mAP values of 0.7878 and 0.7845, approximately 3% higher than baseline.

**Appendix H:** Expert Ratings of *VISLIX*

The experts have also evaluated *VISLIX* using a 5-point Likert scale, shown in Fig. 9. The results reflect a positive overall impression of *VISLIX*.

> **INSTRUCTION:**
> You are an expert in Computer Vision, specializing in analyzing the reasons behind False Negative (FN) detections in object detection models. You are familiar with key concepts in this domain, including bounding box, ground truth, among others.
>
> Your expertise covers various factors contributing to False Positives, including but not limited to, occlusion, size, visibility of parts (truncation), viewpoint, localization error, missing or inaccurate ground truth, and so on.
>
> In the context of a [*object_name*] detection model, a False Negative has occurred in an image, meaning that the ground truth [*object_name*] isn't correctly detected by the model. The ground truth is depicted by a bounding box. The size of the ground truth is [*width*] x [*height*] pixels. The description of the ground truth [*object_name*] is: < [*Q1-groundtruth_caption*] >.
> Only parts of the ground truth [*object_name*] are detected, the detected parts are: < [*Q6-intersection_caption*] >. *(Apply to partial detection)*
> There is a False Positive bounding box fully enclosing the ground truth [*object_name*], a description of the False Positive region is: :< [*FP_caption*] >. *(Apply to over detection)*
>
> Furthermore, there's a cropped image that encompasses the ground truth [*object_name*] and its surroundings, which is described as: < [*Q2-context_caption*] >.
> Given the information provided above, ask me questions about the cropped image to maximize your information of reasoning about the False Negative. Don't mention "False Negative" in your question. Carefully asking me specific and informative questions. Each time ask only one question without giving an answer. I'll put my answer beginning with "Answer:".
>
> **Q&A about the image:**
> Question: What is the weather in this image?
> Answer: [*Q3-weather*]
> Question: How is the lighting condition in this image?
> Answer: [*Q4-lighting*]
> Question: Is there any obstruction blocking the [*object_name*] ? If so, what is obstructing the [*object_name*]? *(Apply to IoU>0.2)*
> Answer: [*Q5-obstruction*]
>
> **TASK:**
>  Next Question(You can still ask [*count*] questions. Please output "STOP" when no questions is needed). Question:

**Figure 7:** *Prompt template used for promoting GPT (Questioner) to raise a question for a False Negative detection.*

Experts evaluated the quality of slice explanations using a Likert scale based on their agreement with the statement: "The edge case explanation is correct for this group." Overall, 65% of explanations received agreement ("Agree" or "Str. Agree"), 21% were rated as "Neutral", and 14% were disagreed with ("Disagree" or "Str. Disagree").

**Appendix I:** Detailed Comparison with *AttributionScanner*

*AttributionScanner* [XOG*25] similarly clusters image embeddings to identify slices but is specifically designed to uncover spurious correlations [XDLM24] in classification models. To ensure a fair comparison, we adapt *VISLIX* to *AttributionScanner*'s second use case that analyzes a bird classification model (landbird/waterbird) using the Waterbirds dataset. Specifically, we extract context-aware embeddings based on bird bounding boxes and make minor adjustments to explanation prompts to better align with the classification task. We apply both approaches to identify slices in the official validation split of 1,199 images, with an overall accuracy of 85.74%. *VISLIX* provides 28 data slices with 84.93% average model accuracy per slice, while *AttributionScanner* provides 46 slices with an average model accuracy per slice being 86.90%. We further examine slices with an accuracy below 70%, significantly below the overall accuracy, resulting in 7 slices for *AttributionScanner* and 5 for *VISLIX*.

As shown in Fig. 10, for *AttributionScanner*, six slices are spuriously associated with water backgrounds (e.g., beach, ocean), leading to the misclassification of landbirds as waterbirds, while one slice is linked to a land background. However, since *AttributionScanner* uses attribution-weighted embeddings that emphasize model-attended regions (in red), they likely capture background elements and overlook the foreground for problematic instances, leading to dispersed bird species within each slice. In contrast, *VISLIX*'s context-aware embeddings focus on birds and their surroundings, combined with HDBSCAN, producing more consistent slices that misclassify specific bird species or birds in particular environments. For example, in Fig. 10, we identify slices that misclassify gadwalls and mallards as landbirds (slice 1) and misclassify kingfishers near water bodies as waterbirds (slice 5). The slice explanations also reveal possible spurious correlations, such as *"association of birds with water due to contextual cues like water bodies, reflections, or behaviors near water"* for slice 5. In particular, we observe that the slice of "towhees in forests" (slice 16) is mislabeled as waterbirds in the data set, which in fact are landbirds.

These findings suggest that the model may be vulnerable to certain bird species, affected by spurious correlations, and reveal potential labeling issues. Moreover, we do not observe significant overlap between the two sets of slices, suggesting that the two approaches offer distinct insights that complement each other, leading to a more comprehensive validation.

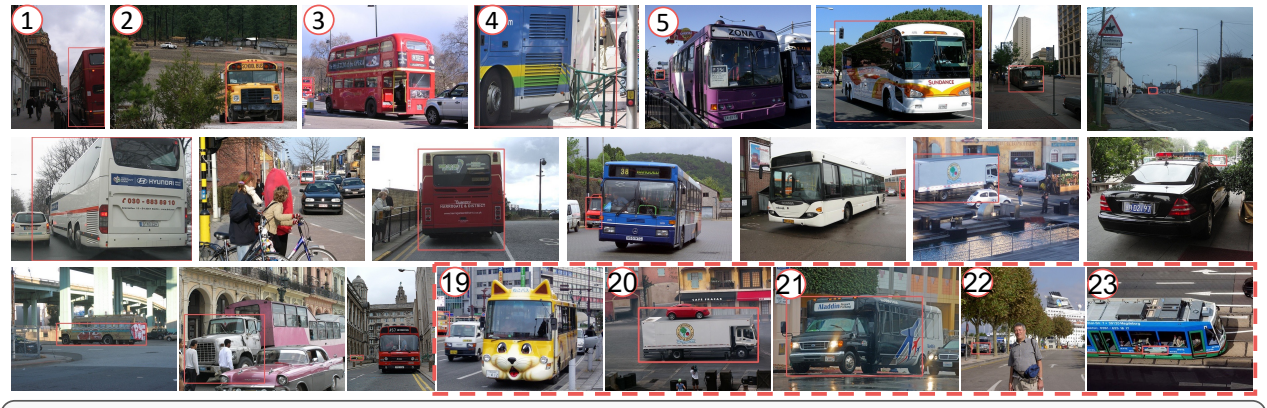

**Slice Explanation using centroid individual explanations**
The common feature in the scenes is the presence of buses and cars in urban environments. The most primary reason for False Positives is the misclassification of buses as cars due to similarities in shape, size, and color, as well as occlusion and partial visibility of the buses.

**Slice Explanation using random individual explanations**
Common feature of these environments: The scenes generally involve urban or city street settings with various objects such as trucks, buses, cars, buildings, trees, and people present. The most common reason for False Positive car detection is the presence of large vehicles, such as trucks or buses, which have similar visual features to cars, leading to confusion in the object detection model.

**Individual Explanations**
1. The scenario is that a false positive car detection occurred in an image of a red double-decker bus driving down a city street. The most possible reason for the false positive detection is that the object detection model mistakenly identified the bus as a car due to similarities in shape, size, and appearance, and potentially due to occlusion of certain parts of the bus or errors in localization.
2. The scenario depicts a yellow school bus parked on a dirt road with people, trucks, and a small building in the background. The most possible reason for the False Positive car detection is that the model may have mistakenly identified the bus as a car due to similarities in shape, size, and color, as well as the presence of headlights on the bus. Additionally, occlusion and the partial visibility of the bus may have contributed to the incorrect detection.
3. The scenario is an urban street scene with a red bus parked next to a tree, a person standing nearby, two small red cars, and a traffic light in the background. The most possible reason for the False Positive CAR detection is that the object detection model mistakenly identified the red bus as a car due to its similar shape and size, as well as the presence of a partially visible car-like structure on the back end of the bus, leading to a misclassification.
4. The scenario is a sunny day with a blue and yellow bus parked on the side of a street, a man standing next to it, and cars parked nearby. The most possible reason for the False Positive car detection could be the similarity in color and shape between the bus and a car, leading to the model mistakenly detecting the bus as a car.
5. The scenario depicts a busy city street with a bus driving down the road. The most possible reason for the False Positive car detection is that the bus in the image shares some visual similarities with a car, such as its shape and color, leading the object detection model to mistakenly identify the bus as a car. Additionally, occlusion, size, and visibility of parts may have contributed to the false positive.

...

19. The scenario is a rainy urban environment with a bus parked next to a building, a car parked behind the bus, and two people nearby. The most possible reason for the False Positive CAR detection could be the presence of similar features between the bus and a car, such as their shapes and colors, leading to a confusion in the object detection model. Additionally, the occlusion and size of the bus may have contributed to the false positive detection.
20. The scenario depicts a sunny day with a dark nighttime lighting condition, featuring a red car driving on top of a white truck carrying a load of fruits and vegetables. The most probable reason for the False Positive car detection is the spurious correlation between the red car on the truck and the actual ground truth car, leading to a misclassification by the object detection model.
21. The scenario is of a false positive car detection where a black bus is parked on the side of the road, partially visible, with a wet street and a building nearby. The most possible reason for the false positive car detection could be the misclassification of the bus as a car due to the following factors: 1. Occlusion: The bus is partially visible, making it difficult for the model to accurately detect its shape and distinguish it from a car. 2. Spurious ... (**Omitted due to the long length**)
22. The image depicts a city street with parked trucks, a partially visible car, a person walking, and a tree, and the most probable reason for the False Positive car detection is the occlusion caused by the large trucks, which blocked the view of the car's front end, resulting in only the back end of the car being detected.
23. The scenario involves a False Positive detection of a car in an image, which is actually a computer screen displaying a blue background with a white keyboard. The most possible reason for the False Positive detection could be the spurious correlation between the blue background on the computer screen and the color of the van in the image, leading the model to mistakenly identify the keyboard as a car.

**Figure 8:** *A slice of false positives (FPs) in the car detector and their corresponding explanations. The images display all 23 FPs in red bounding boxes, sorted by the descending centrality of sentence embeddings derived from their individual explanations. Below, two slice-level explanations are shown: one generated using the top 80% most central individual explanations and another using a random 80% selection. We also show individual explanations of the 5 most central explanations (1–5) and the 5 least central ones (19–23) that are excluded by the centroid method. We highlighted problematic or irrelevant explanations in gray.*

Computer Graphics Forum published by Eurographics and John Wiley & Sons Ltd.

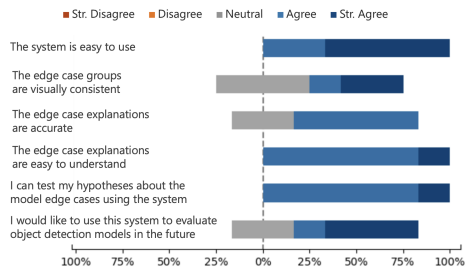

**Figure 9:** *Expert ratings of the system on a five-point Likert scale.*

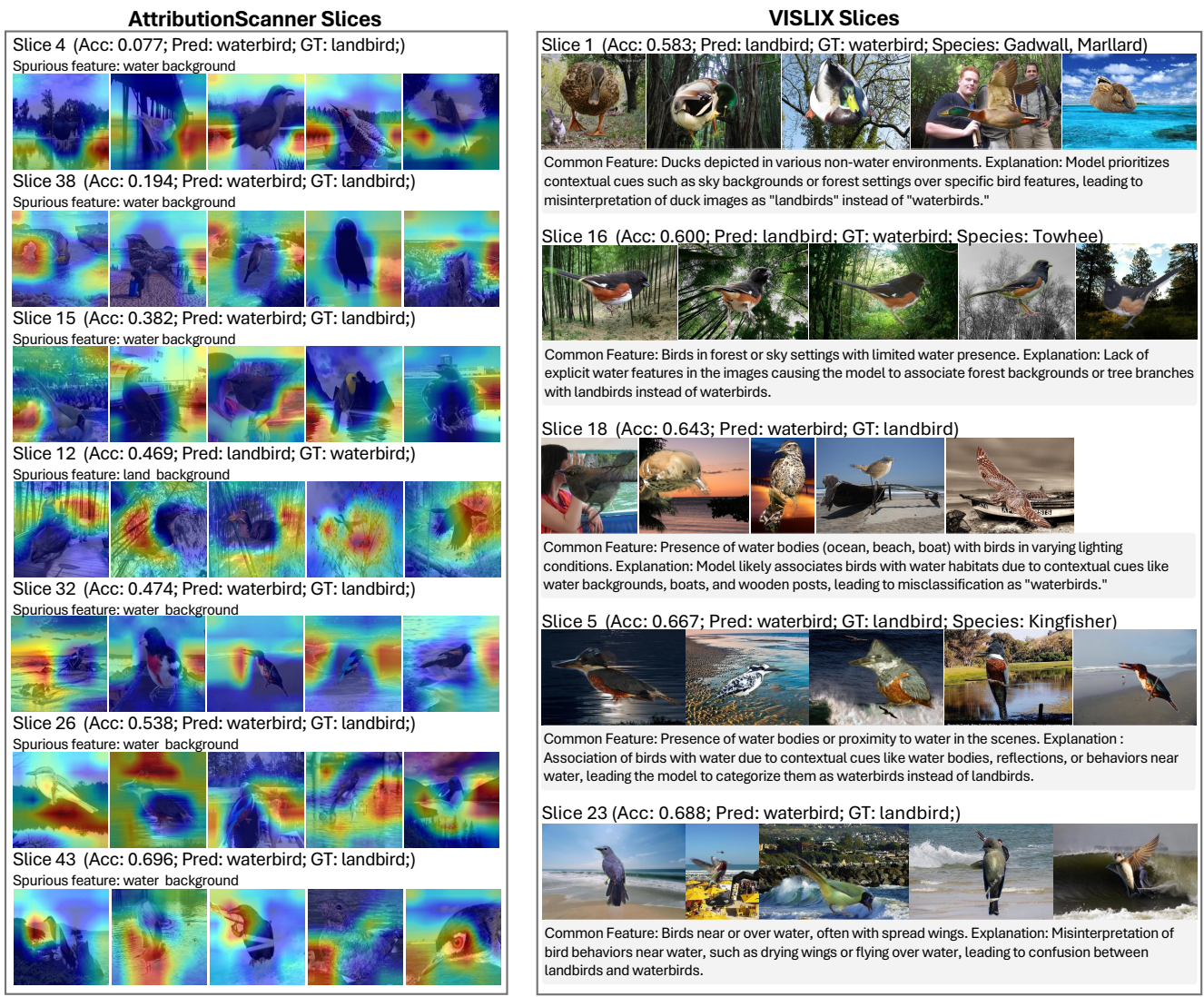

**Figure 10:** *Slices with accuracies below 70% identified by* AttributionScanner *(left, 7 slices) and* VISLIX *(right, 5 slices), for the waterbirds/landbirds classification model. Each slice displays 5 random examples and their predictions (Pred) and ground-truth labels (GT). For* AttributionScanner, *we show spurious features and attribution maps. For* VISLIX, *we provide generated slice explanations, and indicate species based on dataset metadata when they are consistent within the slice.*

# Edge Cases From a Car Detection Model

We present 13 edge cases derived from a **car** detection model. Each case showcases 10 random samples of False Positives (FPs) and False Negatives (FNs), accompanied by a textual explanation summarizing the errors of this edge case.

FPs (indicated by red bounding boxes) denote instances where the model incorrectly identifies objects.

FNs (indicated by blue bounding boxes) denote instances where the model fails to detect the ground truth.

**Edge Case 1** | False Positives: Red bounding box ; False Negatives: Blue bounding box

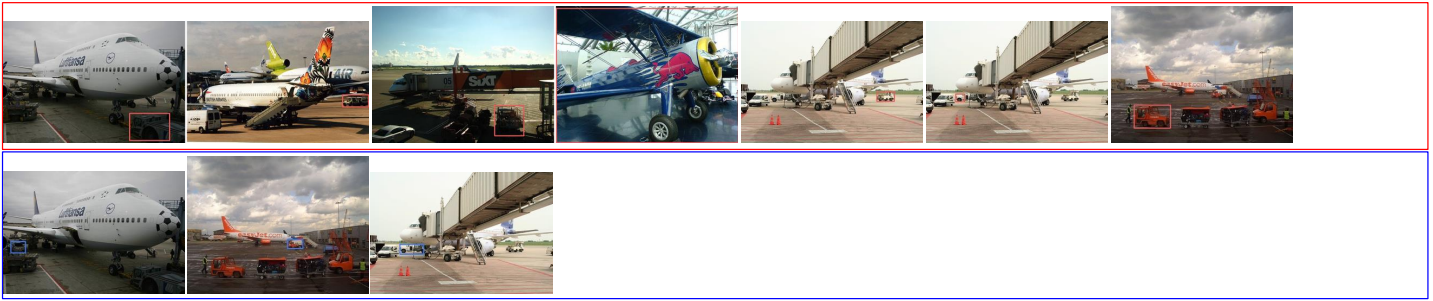

**Explanation:** The common feature of these environments is that they are all in airport settings, with various vehicles and people present. The most primary reason for the False Positives is the misclassification of parts of trucks or airplanes as cars, due to similarities in appearance, occlusion, and background similarity. The most primary reason for the False Negatives is the occlusion caused by other vehicles, such as trucks or luggage carts, obstructing the view of the cars, leading to limited visibility and difficulties in detection.

**Edge Case 2** | False Positives: Red bounding box ; False Negatives: Blue bounding box

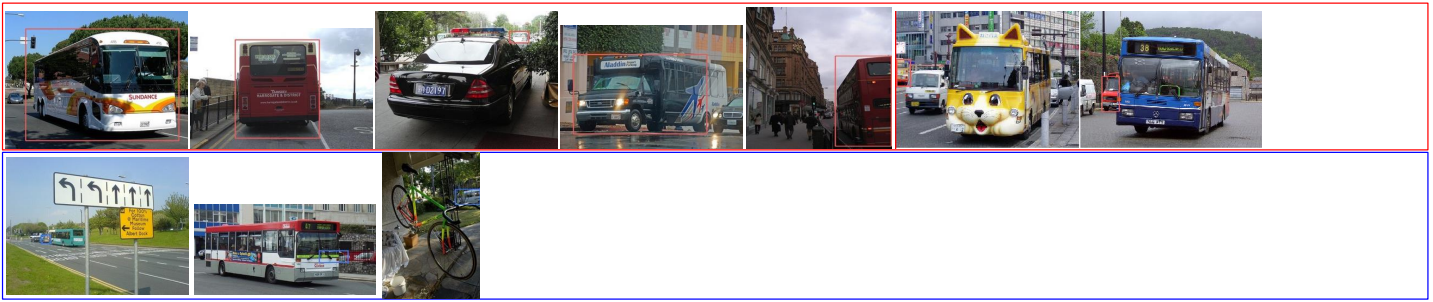

**Explanation:** The common feature in the scenes is the presence of buses and cars in urban environments. The most primary reason for False Positives is the misclassification of buses as cars due to similarities in shape, size, and color, as well as occlusion and partial visibility of the buses. The most primary reason for False Negatives is the occlusion and partial visibility of cars, leading to difficulties in accurately detecting and localizing them.

Edge Case 3 | False Positives: Red bounding box ; False Negatives: Blue bounding box

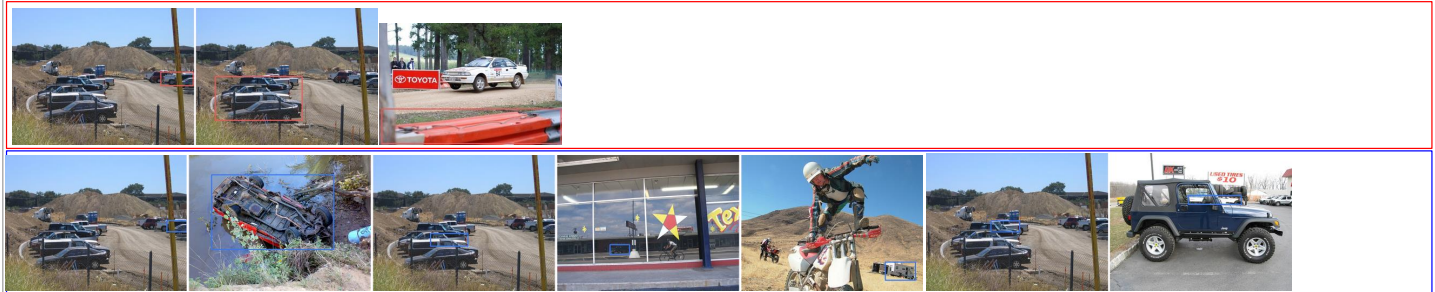

**Explanation:** Common feature: All scenarios involve a car detection model in a parking lot scene with various obstructions and occlusions. Primary reason for False Positives: Similarity in color, shape, and size between objects like electrical boxes and a car, along with occlusion caused by fences or other vehicles. Primary reason for False Negatives: Occlusion caused by other vehicles, fences, or objects obstructing the view of the car, limited visibility of car parts, and challenging lighting conditions.

Edge Case 4 | False Positives: Red bounding box ; False Negatives: Blue bounding box

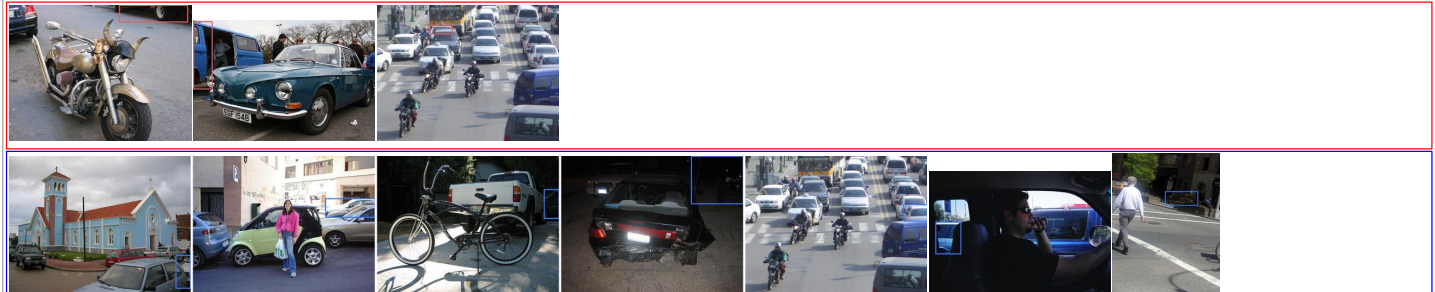

**Explanation**: The common feature of these environments is the presence of occlusion and partial visibility of the cars. The most primary reason for False Positives is the model mistaking other objects, such as car doors, hoods, tires, or curtains, as cars due to their similarity in color, shape, or prominent appearance. The most primary reason for False Negatives is the occlusion caused by other cars, buildings, fences, mirrors, bushes, or people, obstructing the view of the ground truth cars and making it difficult for the model to accurately detect and localize them.

Edge Case 5 | False Positives: Red bounding box ; False Negatives: Blue bounding box

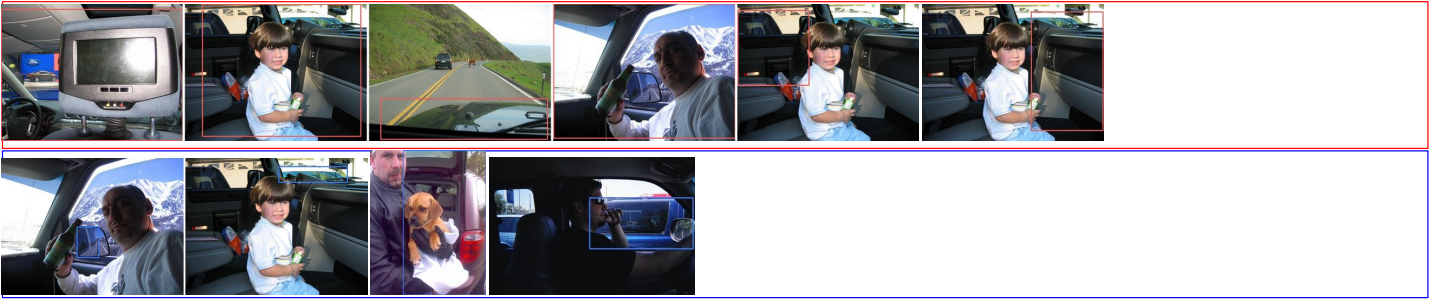

**Explanation:** The common feature in these environments is that they contain elements that can be visually similar to cars or cause occlusion of the car itself. The primary reason for the False Positives is the misinterpretation of objects such as green objects, reflections in mirrors, and dashboard screens as cars. The primary reason for the False Negatives is the occlusion of the car by objects like people, dogs, and mirrors, as well as the limited viewpoint and inaccurate ground truth.

Edge Case 6 | False Positives: Red bounding box ; False Negatives: Blue bounding box

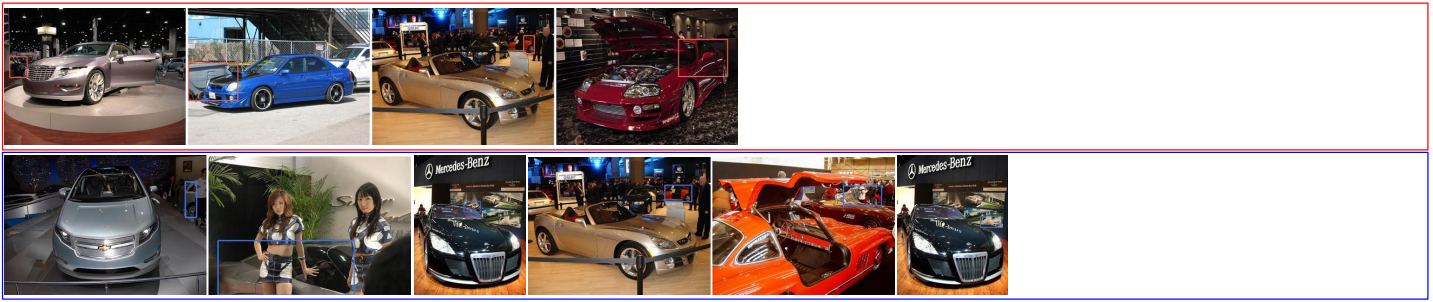

**Explanation**: Common Feature: The scenarios involve partially visible cars in various environments, including garages, showrooms, and parking lots, with occlusion caused by people and other objects. Most likely reason for False Positives: Partial visibility of cars, particularly the open hood and windshield, combined with the presence of people or other objects that may confuse the model. Most likely reason for False Negatives: Occlusion caused by people and other cars, dim lighting conditions, and the presence of other objects/people obstructing the view of the cars.

Edge Case 7 | False Positives: Red bounding box ; False Negatives: Blue bounding box

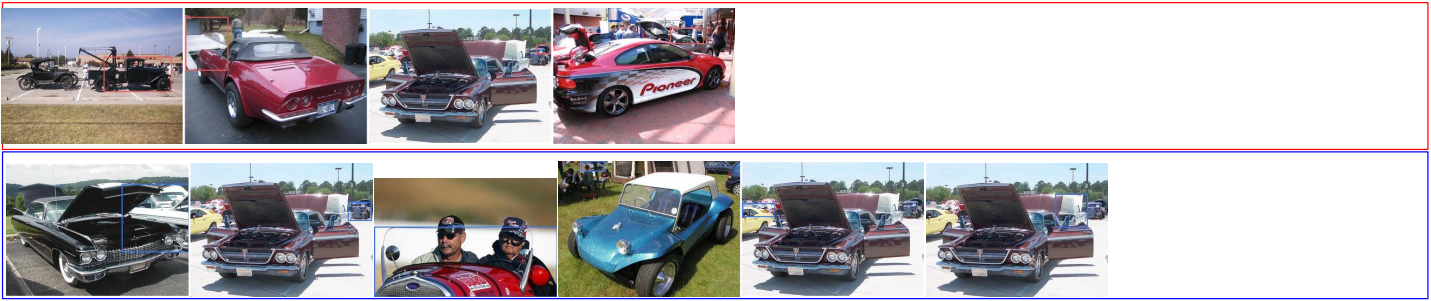

**Explanation:** Common feature of the environments: The scenarios involve parking lots or commercial venues with multiple cars, people, and occlusion present. Most primary reason for False Positives: Similar shapes, colors, or characteristics in the surrounding objects confuse the object detection algorithm. Most primary reason for False Negatives: Occlusion caused by people, other cars, or trailers obstructing the view of the cars, combined with potential localization errors.

Edge Case 8 | False Positives: Red bounding box ; False Negatives: Blue bounding box

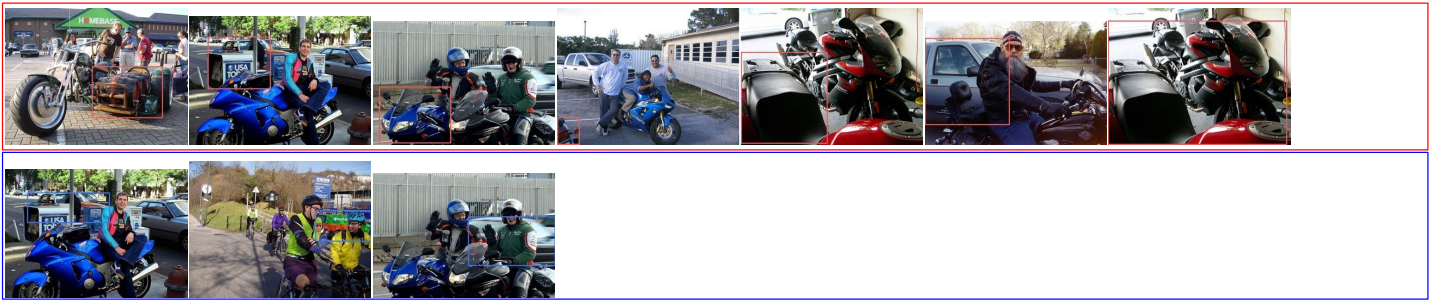

**Explanation**: The common feature in these scenes is the presence of motorcycles and cars. The most primary reason for the False Positives is the confusion between motorcycles and cars due to similarities in appearance, occlusion, and proximity in the image. The most primary reason for the False Negatives is the occlusion and partial visibility of cars caused by motorcycles, people, trees, and other objects in the scene.

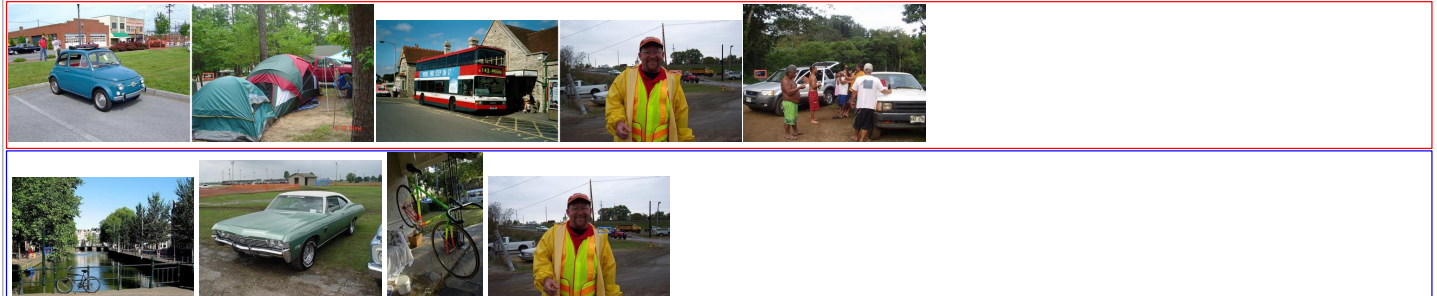

**Edge Case 9** | False Positives: Red bounding box ; False Negatives: Blue bounding box

**Explanation:** Common Feature: All scenarios involve partial or complete occlusion of the cars by different objects such as trucks, bushes, branches, trees, and obstructions created by the environment. Primary Reason for False Positives: The misinterpretation of partial visibility and occlusion by other objects as complete car detection. Primary Reason for False Negatives: The difficulty in accurately detecting cars due to partial or complete occlusion by various objects, such as trucks, bushes, branches, trees, and obstructions created by the environment.

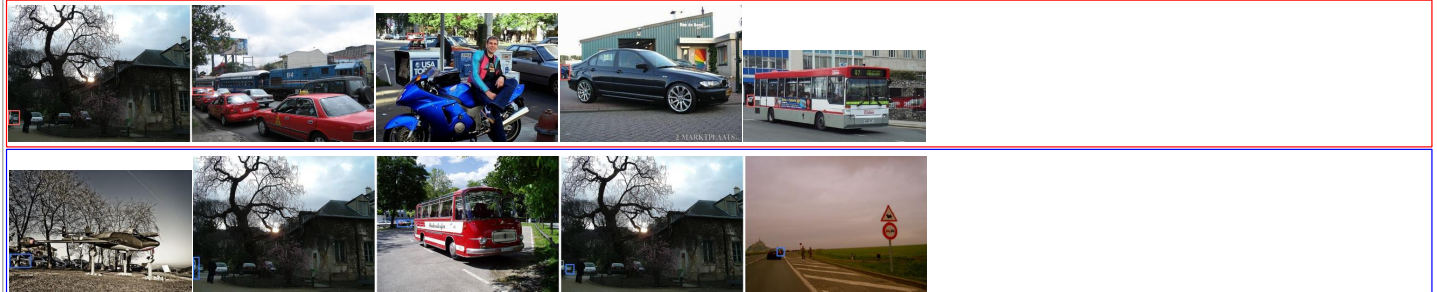

**Edge Case 10** | False Positives: Red bounding box ; False Negatives: Blue bounding box

**Explanation**: Common Feature: The scenarios involve parking lots and busy streets with various cars and occlusion. False Positives: The most primary reason for false positives is occlusion and limited visibility of the car due to other parked vehicles, taxi cabs, or trees. The detection model mistakenly classifies the visible parts as cars. False Negatives: The most primary reason for false negatives is occlusion caused by other cars, trees, or people, leading to limited visibility of the car. The model struggles to detect the complete vehicle accurately in these scenarios.

Edge Case 11 | False Positives: Red bounding box ; False Negatives: Blue bounding box

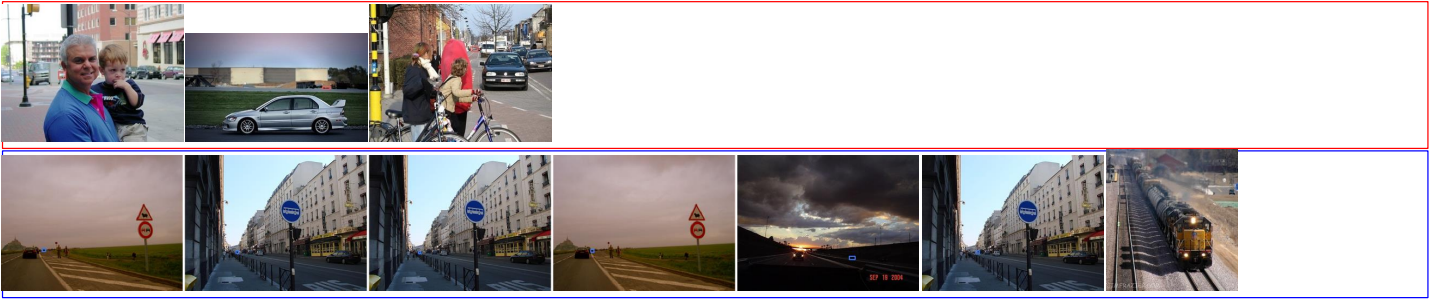

**Explanation:** The common feature of these environments is that they involve various factors that make it challenging for the car detection model to accurately detect and localize cars. The primary reason for False Positives is the presence of similar colors and shapes in the background or other objects. The primary reason for False Negatives is a combination of factors including occlusion, low visibility, dark lighting conditions, small car size, and blurry backgrounds.

Edge Case 12 | False Positives: Red bounding box ; False Negatives: Blue bounding box

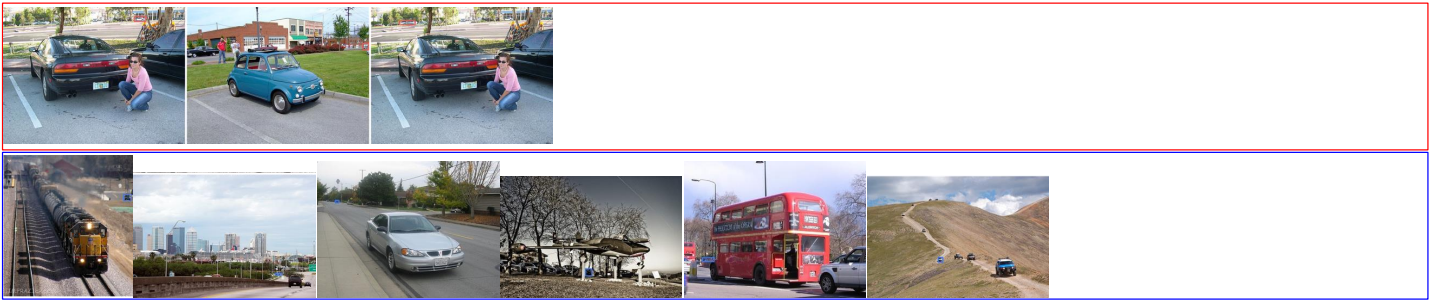

**Explanation**: The common feature of these environments is that they all involve parking lots or street scenes with partially visible cars and various occlusions. The primary reason for False Positives is the partial visibility of the car, occlusion by other objects, and misinterpretation of the visible parts as complete cars. The primary reason for False Negatives is occlusion caused by other cars, fences, trees, or buildings, along with the low angle, dim lighting conditions, and similar colors blending with the surroundings.

Edge Case 13 | False Positives: Red bounding box ; False Negatives: Blue bounding box

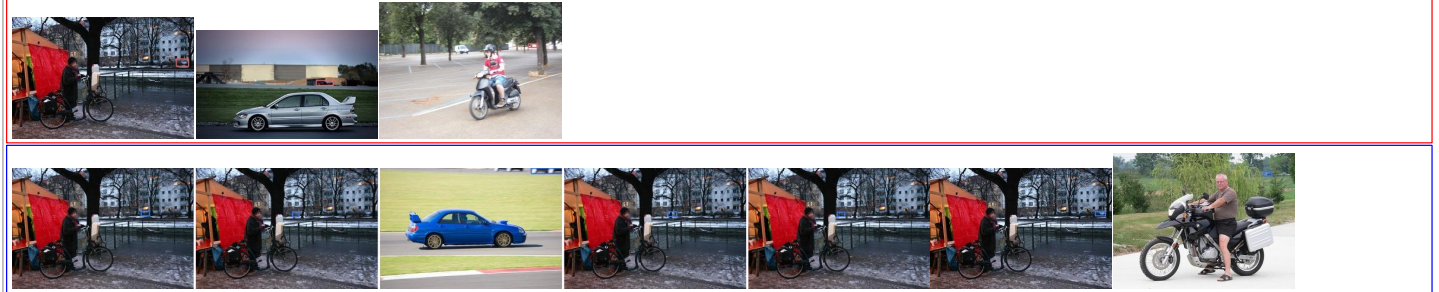

**Explanation:** Common feature: The scenarios involve various obstacles obstructing the view of the cars, such as tree branches, rocks, snow banks, and other cars. The lighting conditions are often dark, and the images are sometimes blurry or black and white. Primary reason for False Positives: The presence of other objects or structures, such as a clock tower, a black object, or a person, that resemble or confuse the model into incorrectly detecting them as cars. Primary reason for False Negatives: The occlusion caused by tree branches and other objects, combined with low visibility due to darkness, blurriness, or small size of the cars. The model also struggles to accurately detect the shape and features of the cars in these challenging environments.

## Edge Cases From a Dog Detection Model

We present 7 edge cases derived from a **dog** detection model. Each case showcases 10 random samples of False Positives (FPs) and False Negatives (FNs), accompanied by a textual explanation summarizing the errors of this edge case.

FPs (indicated by red bounding boxes) denote instances where the model incorrectly identifies objects.

FNs (indicated by blue bounding boxes) denote instances where the model fails to detect the ground truth.

Edge Case 1 | False Positives: Red bounding box ; False Negatives: Blue bounding box

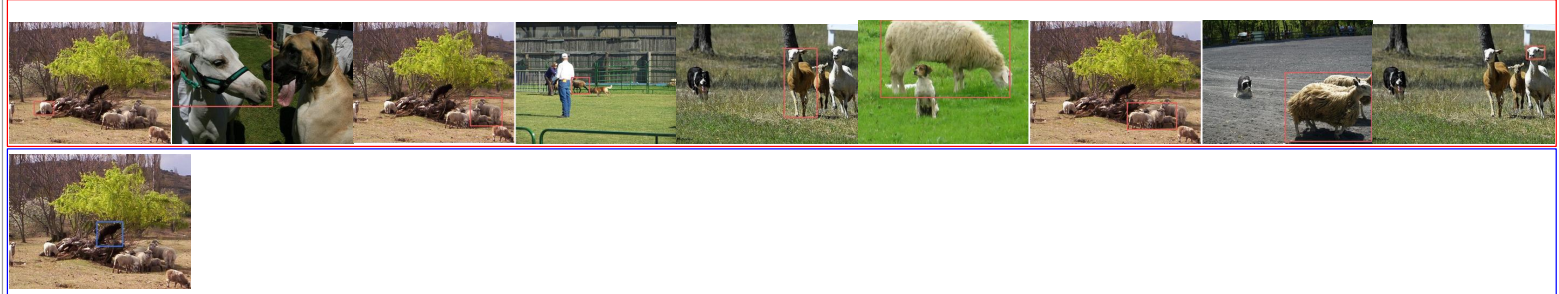

**Explanation:** Common Feature: Grassy fields with sheep and other animals present. Most Primary Reason for False Positives: Misclassification due to similarities in appearance, size, shape, color, and occlusion between sheep and dogs. The dog detection model may also confuse other animals like deer, horses, and goats as dogs. Most Primary Reason for False Negatives: Partial occlusion and visibility of the dog's body, particularly the head, by surrounding objects such as trees and branches, leading to difficulty in accurate detection by the model.

Edge Case 2 | False Positives: Red bounding box ; False Negatives: Blue bounding box

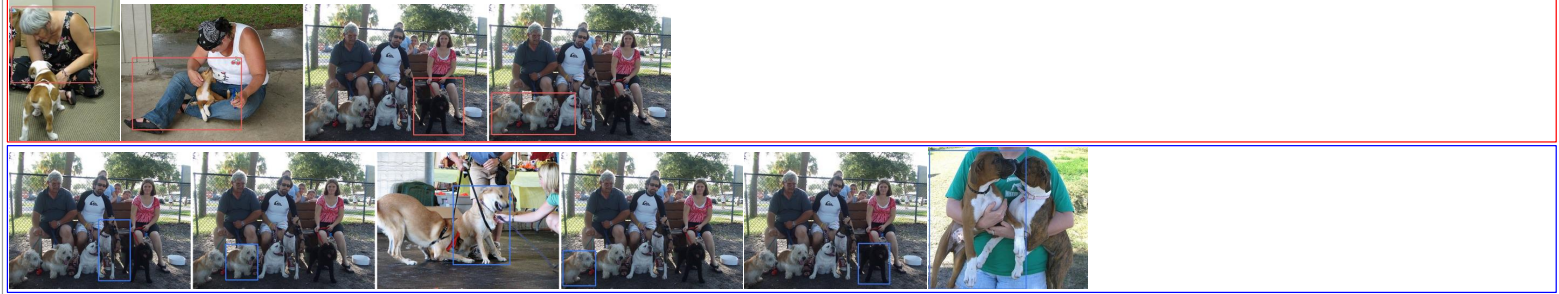

**Explanation:** The common feature of these environments is that they involve people interacting with dogs in various settings, including outdoor parks, benches, and indoor rooms. The most primary reason for False Positives is occlusion caused by people's arms, bodies, and objects, which partially cover the dog's body, leading to misinterpretation as a dog or misclassification based on similar features. The most primary reason for False Negatives is occlusion caused by people, leashes, and other dogs, which obstruct the view of the dog and make it difficult for the model to accurately detect and localize the dog.

**Edge Case 3** | False Positives: Red bounding box ; False Negatives: Blue bounding box

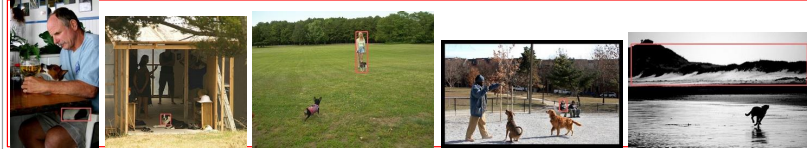

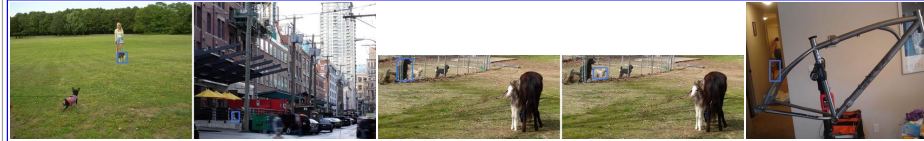

**Explanation:** Common Feature: The common feature of these environments is the presence of objects or elements that visually resemble dogs, such as jackets, clothes, people, and occlusions caused by fences, grass, and other objects. Primary Reason for False Positives: The primary reason for false positives is the model's confusion between visually similar objects or elements, such as mistaking jackets, black Labrador retriever, and mountain range features as dogs. Primary Reason for False Negatives: The primary reason for false negatives is the occlusion or obstruction of the dog's body or features, such as fences, people, grass, and shadows, making it difficult for the model to accurately detect the complete dog.

**Edge Case 4** | False Positives: Red bounding box ; False Negatives: Blue bounding box

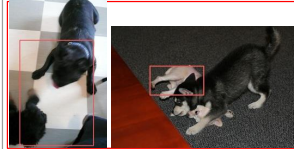

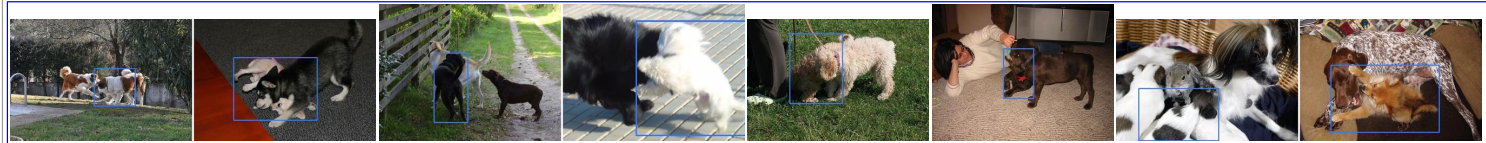

**Explanation:** The common feature in these environments is the presence of multiple dogs and occlusion caused by other objects or dogs. The most primary reason for False Positives is the detection model only detecting parts of the dog, such as the rear end or face, without detecting the full body. The most primary reason for False Negatives is occlusion, where the ground truth dog is partially or fully hidden by other dogs or objects, making it challenging for the model to accurately detect the complete dog.

Edge Case 5 | False Positives: Red bounding box ; False Negatives: Blue bounding box

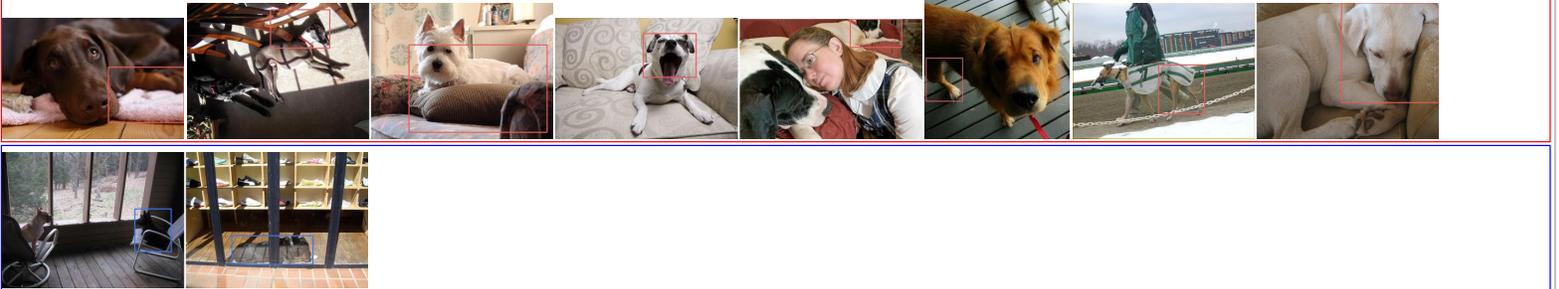

**Explanation:** Common Feature: Several false positives and false negatives occur due to occlusion or obstruction in the scene, where parts of the dog's body are covered or hidden from view. Primary Reason for False Positives: The primary reason for false positives is the model mistaking objects or elements in the scene, such as towels, pillows, patterns, or furniture, as part of the dog due to similarities in color, texture, or shape. Primary Reason for False Negatives: The primary reason for false negatives is the occlusion or obstruction caused by window frames, furniture, or other objects, which prevent the model from accurately detecting the complete dog. Dark lighting conditions can also contribute to false negatives.

Edge Case 6 | False Positives: Red bounding box ; False Negatives: Blue bounding box

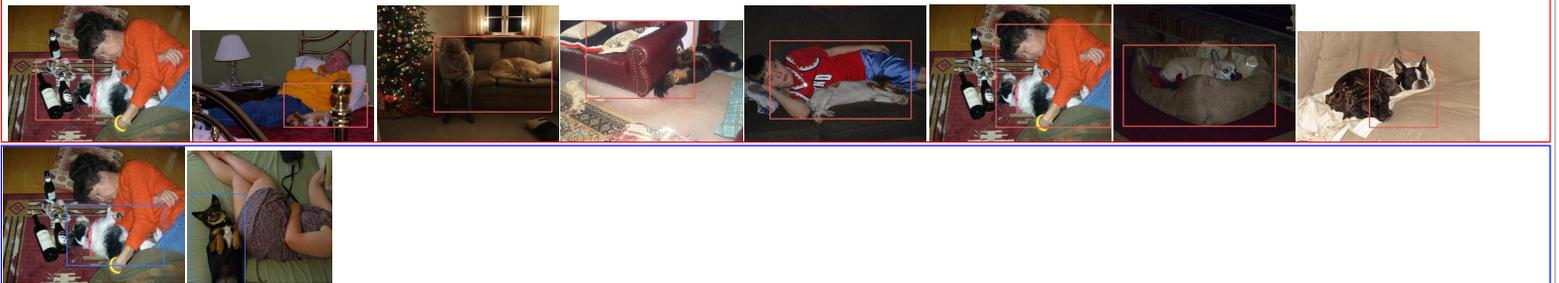

**Explanation:** Common Feature: The scenarios involve dogs in various locations such as beds, couches, and floors, often with occlusion from objects or people. Most Primary Reason for False Positives: Occlusion caused by objects or people obstructing the view of the dog, similarity in color and texture between the dog and the surroundings, and misinterpretation of background objects as separate dogs. Most Primary Reason for False Negatives: Obstruction of the dog's head or body by people or objects, resulting in limited visibility and inability to accurately detect the dog.

**Edge Case 7** | False Positives: Red bounding box ; False Negatives: Blue bounding box

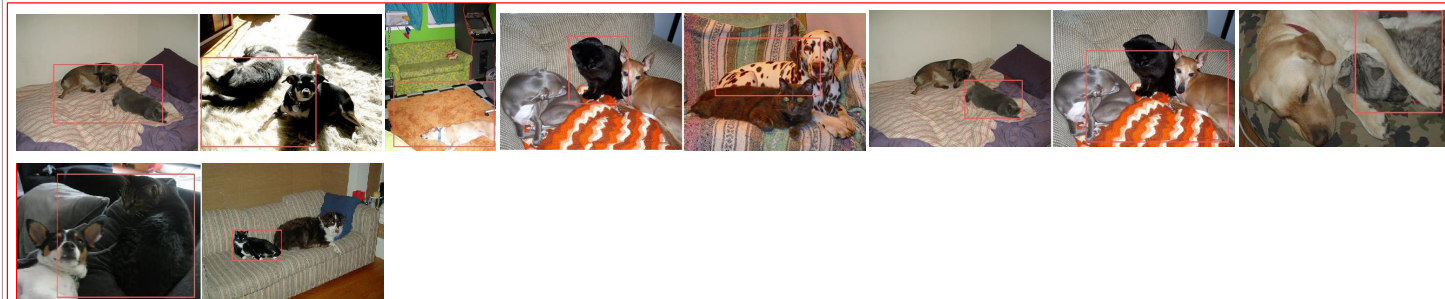

**Explanation:** : Common Feature: The common feature in these scenarios is that they all involve scenes with cats and dogs resting together on a couch or bed. Most Primary Reason for False Positives: The most primary reason for the false positives is the misclassification caused by the partial visibility of the dogs and their similarity to the background or the cats in terms of color, shape, and context. Occlusion by other objects, such as blankets or furniture, also contributes to the false positives.

## Edge Cases From a Chair Detection Model

We present 15 edge cases derived from a **chair** detection model. Each case showcases 10 random samples of False Positives (FPs) and False Negatives (FNs), accompanied by a textual explanation summarizing the errors of this edge case.

FPs (indicated by red bounding boxes) denote instances where the model incorrectly identifies objects.

FNs (indicated by blue bounding boxes) denote instances where the model fails to detect the ground truth.

**Edge Case 1** | False Positives: Red bounding box ; False Negatives: Blue bounding box

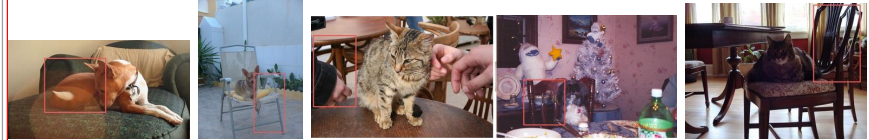

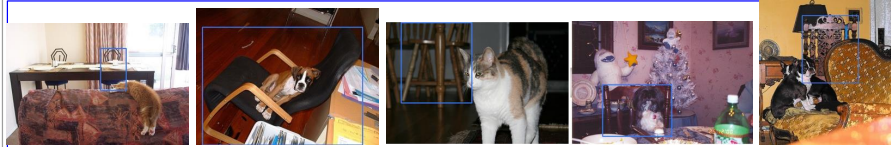

**Explanation:** Common Feature: The common feature in these environments is the presence of animals (cats and dogs) and occlusion caused by them, as well as dim lighting conditions. Primary Reason for False Positives: The primary reason for false positives is the misclassification of animals (cats and dogs) as chairs due to similarity in color, shape, and texture, as well as occlusion caused by the animals obstructing the view of the chair. Primary Reason for False Negatives: The primary reason for false negatives is the occlusion caused by animals (cats and dogs) partially or fully blocking the view of the chair, making it difficult for the model to accurately detect the chair. Dim lighting conditions and the small size of the detected chair parts also contribute to false negatives.

**Edge Case 2** | False Positives: Red bounding box ; False Negatives: Blue bounding box

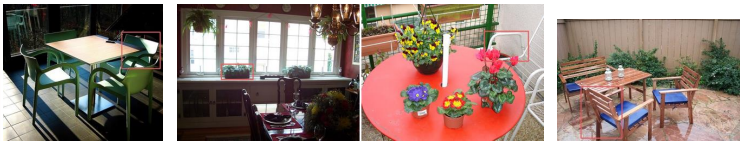

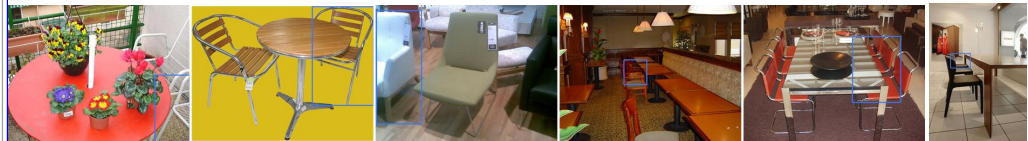

**Explanation:** Common feature of environments: The scenarios involve various indoor settings such as dining rooms, restaurants, and rooms with furniture. Most primary reason for False Positives: Similarity in color, shape, and features between the chair and other objects in the environment, as well as partial visibility and occlusion of the chair, leading to confusion in the detection model. Most primary reason for False Negatives: Occlusion of the chair by other objects in the environment, such as tables or other chairs, combined with partial visibility, dim lighting conditions, and similarity in color between the chair and the obstructing objects, making it challenging for the model to accurately detect the chair.

**Edge Case 3** | False Positives: Red bounding box ; False Negatives: Blue bounding box

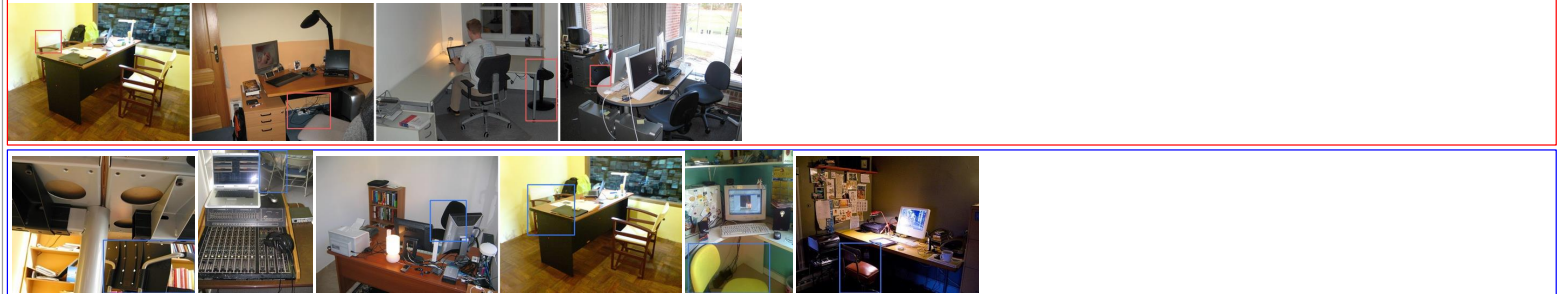

**Explanation:** The scenarios involve various instances where the chair is partially visible and obstructed by other objects such as tables, desks, monitors, keyboards, and other items commonly found in workspaces or rooms. The primary reason for the false positive chair detections is the misinterpretation of partially visible chairs and tables as complete chairs. This could be due to the object detection model mistaking the partial visibility and occlusion as the complete chair, leading to incorrect detections. The primary reason for the false negative chair detections is the occlusion caused by other objects, such as tables, desks, monitors, keyboards, and furniture, obstructing the view of the chair. The occlusion makes it difficult for the model to accurately detect the entire chair, resulting in false negatives.

**Edge Case 4** | False Positives: Red bounding box ; False Negatives: Blue bounding box

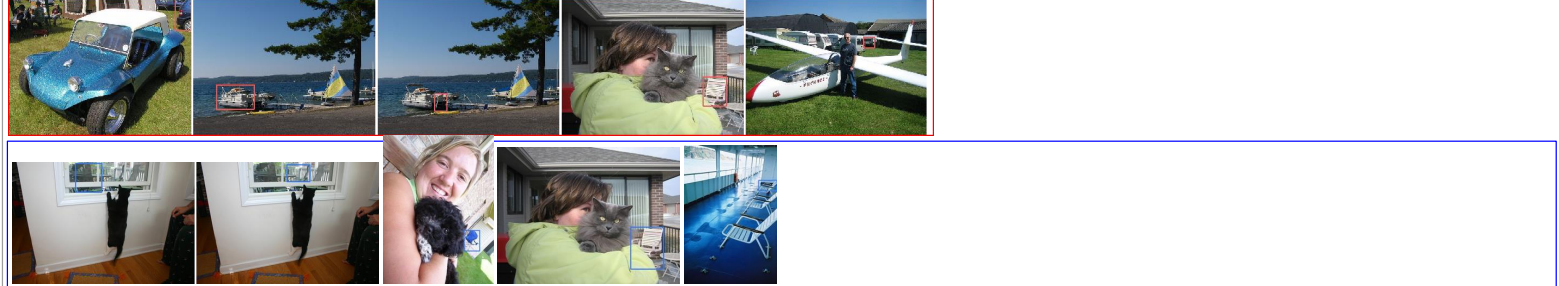

**Explanation:** Common feature of these environments: - Outdoor scenes with various objects and occlusions, such as people, vehicles, boats, and structures like windows and railings. Primary reason for False Positives: - Confusion caused by occlusions, partial visibility, and visual similarities with other objects, such as people, vehicles, boat structures, and ladders. Primary reason for False Negatives: - Occlusion by obstructions like people, glass doors, windows, railings, benches, and boats, as well as partial visibility, complex backgrounds, and smaller size compared to other objects.

**Edge Case 5** | False Positives: Red bounding box ; False Negatives: Blue bounding box

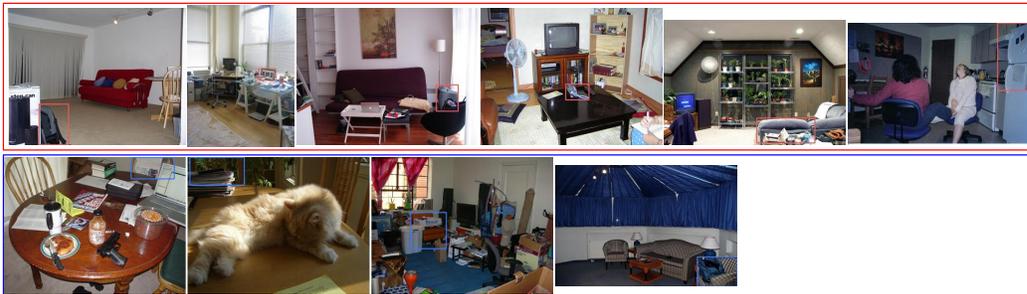

**Explanation:** The scenarios involve cluttered rooms or images with various objects and occlusion, making it challenging for the chair detection model to accurately identify the chair. The primary reason for false positives is the presence of objects with similar shapes, sizes, or colors to a chair, such as boxes, bags, or furniture, leading to confusion for the chair detection model. Additionally, partial visibility and occlusion of the chair by other objects contribute to inaccurate detection. The primary reason for false negatives is occlusion, where the chair is partially hidden or obstructed by other objects such as stacks of papers, bags, or furniture, making it difficult for the model to detect the chair accurately. Dim lighting conditions and the small size of the chair also contribute to the false negative detection.

**Edge Case 6** | False Positives: Red bounding box ; False Negatives: Blue bounding box

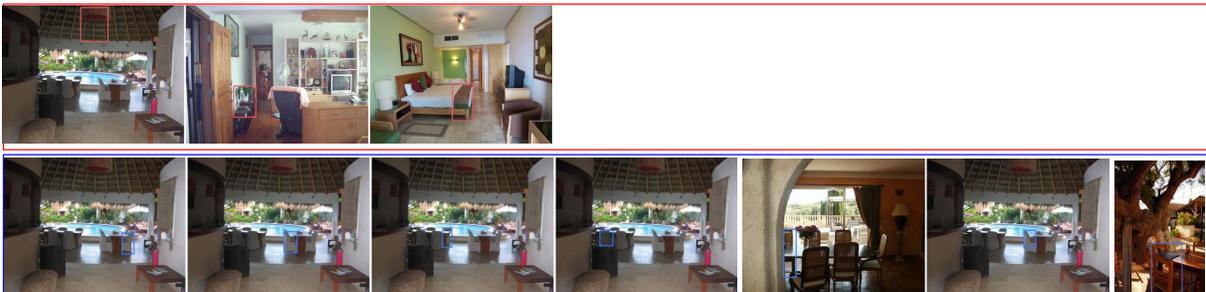

**Explanation:** Common Feature: The common feature in these scenes is the presence of occlusion or obstruction, either by other objects or by the environment (e.g., lighting conditions). The primary reason for false positives is the confusion caused by partial occlusion or obstruction of the chair by other objects, such as beds, potted plants, wooden beams, or similar structures. This confusion leads the detection model to misclassify these objects as chairs. The primary reason for false negatives is the occlusion or obstruction of the chair by other objects, such as tables, towels, potted plants, lamps, couches, or people. The occlusion makes it difficult for the model to accurately detect and localize the chair, resulting in false negatives. Additionally, dim lighting conditions can further affect the model's ability to detect the chair accurately.

**Edge Case 7** | False Positives: Red bounding box ; False Negatives: Blue bounding box

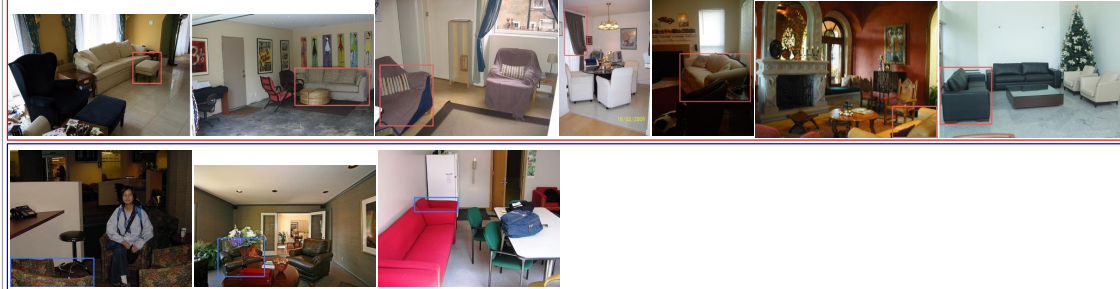

**Explanation:** Common Feature: The common feature in these scenarios is that they all involve living rooms with various furniture and objects, such as couches, tables, pillows, and chairs. The primary reason for the False Positives is the presence of objects that resemble chairs, such as couches, pillows, and tables, which can confuse the object detection model. Additionally, occlusion and partial visibility of the chairs contribute to the incorrect detection. The primary reason for the False Negatives is occlusion caused by other objects, such as pillows, vases, cabinets, tables, and people, which obstruct the view of the chairs and make it difficult for the model to accurately detect them. Additionally, the focus on other objects and the dim lighting conditions may divert attention away from the chairs

**Edge Case 8** | False Positives: Red bounding box ; False Negatives: Blue bounding box

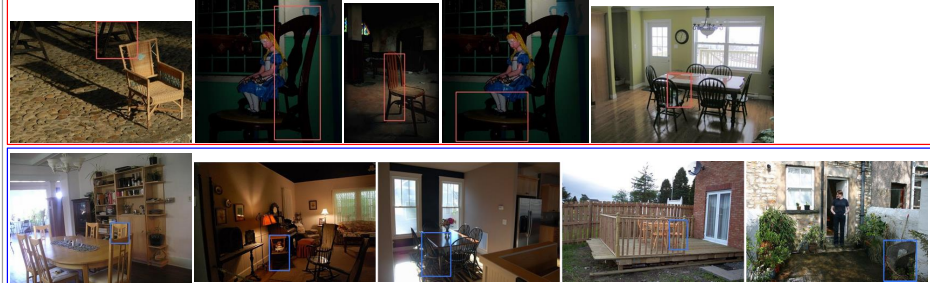

**Explanation:** Common Feature: The common feature in these environments is the presence of occlusion, either due to objects partially covering the chair or the chair being partially visible. Primary Reason for False Positives: The primary reason for false positives is the occlusion caused by objects such as tablecloths, doors, dolls, fabric coverings, and other furniture pieces, leading to a partial detection of the chair's visible parts and confusing the model. Primary Reason for False Negatives: The primary reason for false negatives is the occlusion caused by tables, cabinets, shelves, plants, vases, curtains, and other objects obstructing the view of the chair, making it challenging for the model to accurately detect and localize the chair.

**Edge Case 9** | False Positives: Red bounding box ; False Negatives: Blue bounding box

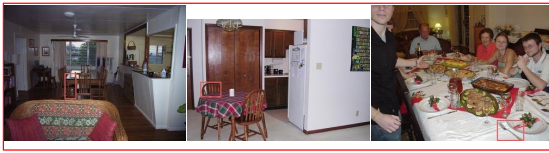

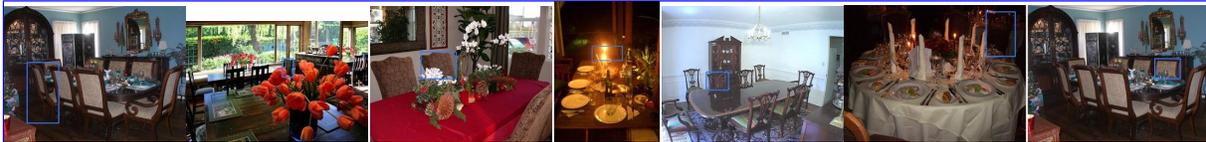

**Explanation:** Common Feature: The common feature in these scenes is that they all involve dining tables or dining areas with chairs, often with occlusions or obstructions present. Primary Reason for False Positives: The primary reason for False Positives is the confusion caused by occlusions or obstructions, such as tables, objects on the table, or other chairs, leading to misclassification by the chair detection model. Primary Reason for False Negatives: The primary reason for False Negatives is the occlusion or partial visibility of the chairs, often caused by obstructions like potted plants, vases, tablecloths, or other objects, making it challenging for the model to accurately detect the chairs.

**Edge Case 10** | False Positives: Red bounding box ; False Negatives: Blue bounding box

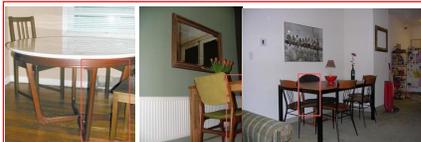

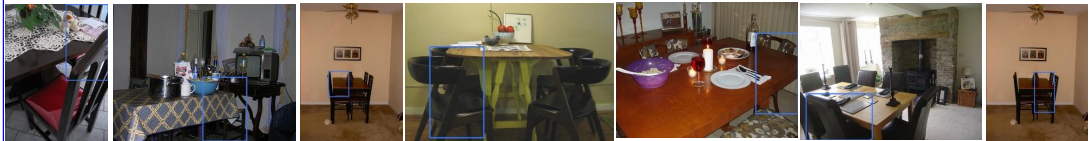

**Explanation:** Common Feature: The common feature in these scenes is the presence of occlusion and obstruction, either by other objects or by the environment (e.g., tables, curtains, bags, plants, etc.). The scenes often have dim lighting conditions and a variety of other objects present. The most primary reason for False Positives is the occlusion caused by tables, obstructing the view of the chair. Additionally, low-angle perspectives, small chair sizes, and confusion with similar objects contribute to false positives. The most primary reason for False Negatives is occlusion caused by other objects such as tables, vases, suitcases, cushions, and rugs, obstructing the view of the chair. Dim lighting conditions, truncation, and partial visibility of the chair further contribute to false negatives.

Edge Case 11 | False Positives: Red bounding box ; False Negatives: Blue bounding box

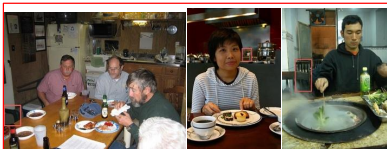

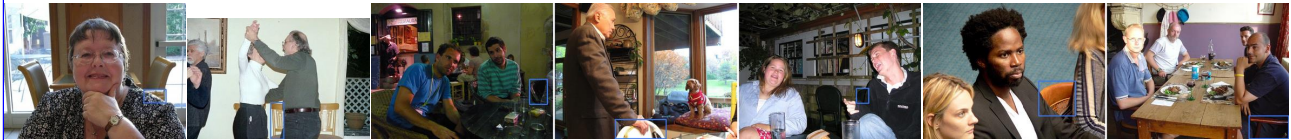

**Explanation:** - The common feature in these environments is that they contain various occlusions and obstructions, such as people, objects, and fabrics, which partially or completely cover the chairs. - The most primary reason for False Positives is the model mistaking parts of the chair, such as armrests, backrests, or fabrics covering the chair, as complete chairs. This is due to similar colors, textures, and shapes between the chair and other objects in the scene. - The most primary reason for False Negatives is the occlusion and obstruction caused by people, objects, or fabrics partially covering the chairs. The limited visibility, combined with dim lighting conditions, make it difficult for the model to accurately detect the complete chairs.

Edge Case 12 | False Positives: Red bounding box ; False Negatives: Blue bounding box

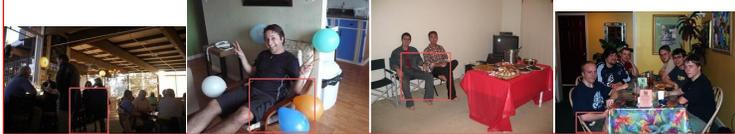

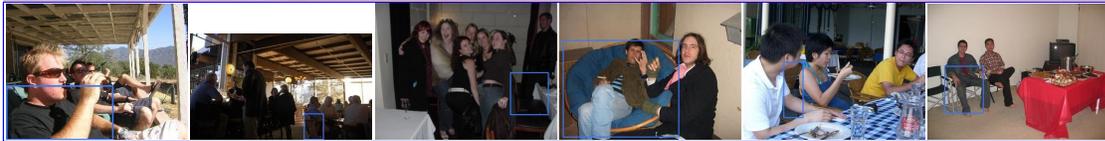

**Explanation:** The common feature in these scenes is the presence of various objects and people, including chairs, in different environments such as restaurants, dining areas, dimly lit rooms, and cozy settings. False Positives: The confusion caused by similar visual features, such as shape, color, and texture, between the chairs and other objects in the scene, including stairs, railing, wooden steps, person's clothing, backrest, and chair-like patterns. Occlusion from objects and people in the scene also contributes to the false positives. False Negatives: The occlusion caused by people sitting on the chairs, obstructing the view of the chair's design and structure. Additionally, partial visibility of the chair, small size, dim lighting conditions, and the focus of the image on people and surroundings rather than the chair itself contribute to the false negatives.

**Edge Case 13** | False Positives: Red bounding box ; False Negatives: Blue bounding box

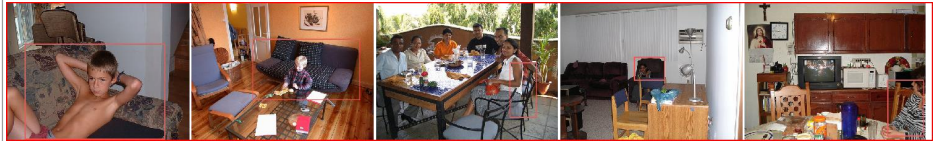

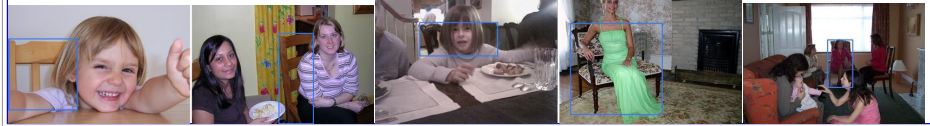

**Explanation:** Common feature of these environments: - The images often contain people sitting at dining tables or on couches in dimly lit rooms. Most primary reason for False Positives: - Chairs being partially hidden or obstructed by people or objects, leading to confusion in the object detection model and resulting in false positive detections.
Most primary reason for False Negatives: - Occlusion caused by people sitting on or partially obstructing the chairs, making it difficult for the model to accurately detect and localize the complete chair.

**Edge Case 14** | False Positives: Red bounding box ; False Negatives: Blue bounding box

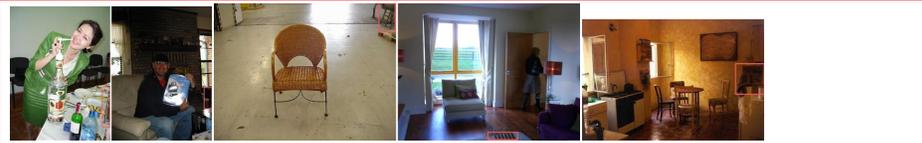

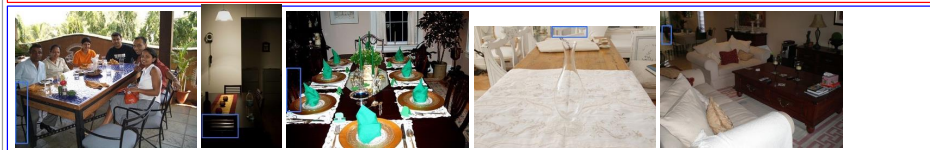

**Explanation:** Common Feature: The common feature among these scenes is the presence of occlusion or partial visibility of the chair. False Positives: The primary reason for false positives is the presence of spurious correlation or confusion due to similar features, shapes, or colors between the chair and other objects in the scene, leading to misclassification. False Negatives: The primary reason for false negatives is the occlusion or obstruction of the chair, either by other objects or by the scene itself, making it difficult for the model to accurately detect the chair.

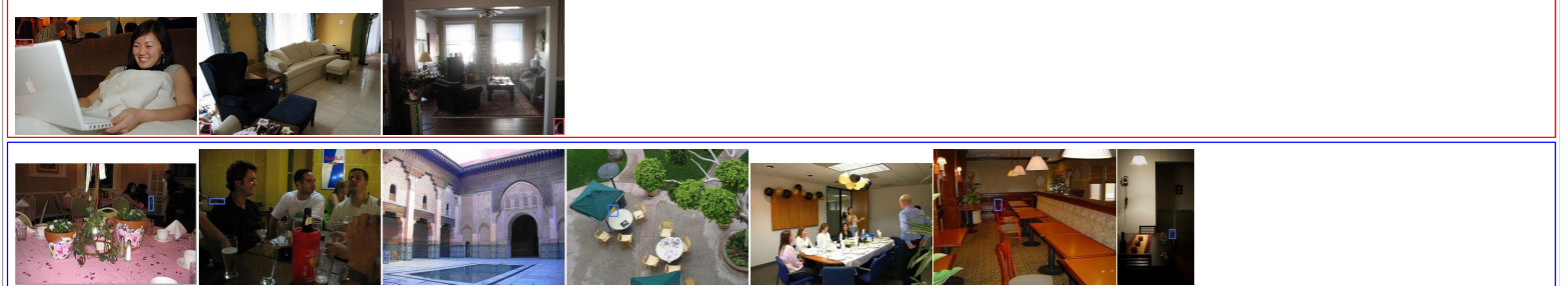

Edge Case 15 | False Positives: Red bounding box ; False Negatives: Blue bounding box

**Explanation:** Common Feature: The common feature in these environments is that they involve various forms of occlusion, such as objects, people, or furniture obstructing the view of the chairs. Primary Reason for False Positives: The primary reason for false positives is the model mistaking objects or textures in the scene, such as fabric, spheres, reflective surfaces, or wigs, as chairs due to similarities in color, shape, or texture. Primary Reason for False Negatives: The primary reason for false negatives is the combination of occlusion, small chair size, limited visibility, dim lighting conditions, and the model's difficulty in accurately detecting and localizing chairs that are partially visible or obstructed by other objects or people.

# Edge Cases From a Person Detection Model

We present 24 edge cases derived from a **person** detection model. Each case showcases 10 random samples of False Positives (FPs) and False Negatives (FNs), accompanied by a textual explanation summarizing the errors of this edge case.

FPs (indicated by red bounding boxes) denote instances where the model incorrectly identifies objects.

FNs (indicated by blue bounding boxes) denote instances where the model fails to detect the ground truth.

**Edge Case 1** | False Positives: Red bounding box ; False Negatives: Blue bounding box

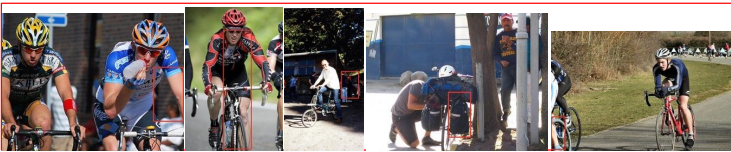

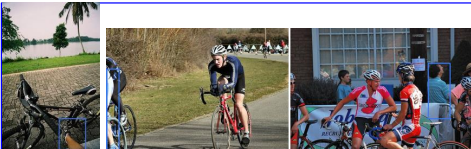

**Explanation:** : Common Feature: The common feature in these environments is the presence of bicycles and people wearing various outfits in outdoor settings. The primary reason for False Positives is the occlusion caused by the presence of bicycles, backpacks, chairs, and other objects, as well as the similarity in color, shape, and texture between these objects and a person. Additionally, the limited visibility and partial detection of body parts contribute to the misclassification. The primary reason for False Negatives is the occlusion caused by other objects, such as bicycles and people, obstructing the view of the person. The model struggles to detect the complete person when their body is partially hidden, leading to the false negative.

**Edge Case 2** | False Positives: Red bounding box ; False Negatives: Blue bounding box

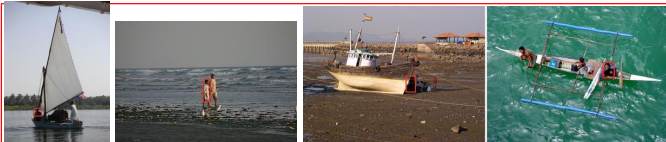

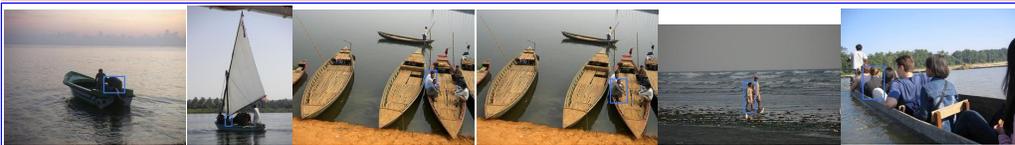

**Explanation:** Common Feature of Environments: - Sunny outdoor scenes with various objects like boats, beaches, wooden piers, and rocky shorelines. Primary Reason for False Positives: - Occlusion and limited visibility due to obstructions, such as ropes, poles, and other people, leading to the model incorrectly identifying parts of a person's body as a complete person. Primary Reason for False Negatives: - Occlusion caused by objects like paddles, wooden planks, trees, and other people, obstructing the view of the person's body and making it difficult for the model to accurately detect the complete person.

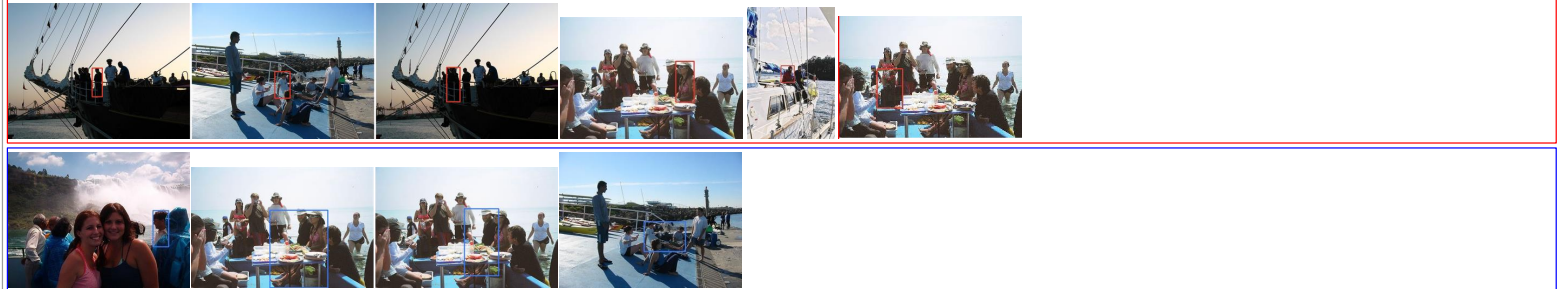

**Edge Case 3** | False Positives: Red bounding box ; False Negatives: Blue bounding box

**Explanation:** Common Feature of Environments: - The scenarios all involve images taken on a boat or near water. Primary Reason for False Positives: - Occlusion and limited visibility of the person's body parts, leading to misinterpretation by the object detection model. Primary Reason for False Negatives: - Occlusion caused by objects, people, or body parts obstructing the view of the person, making it difficult for the model to accurately detect them.

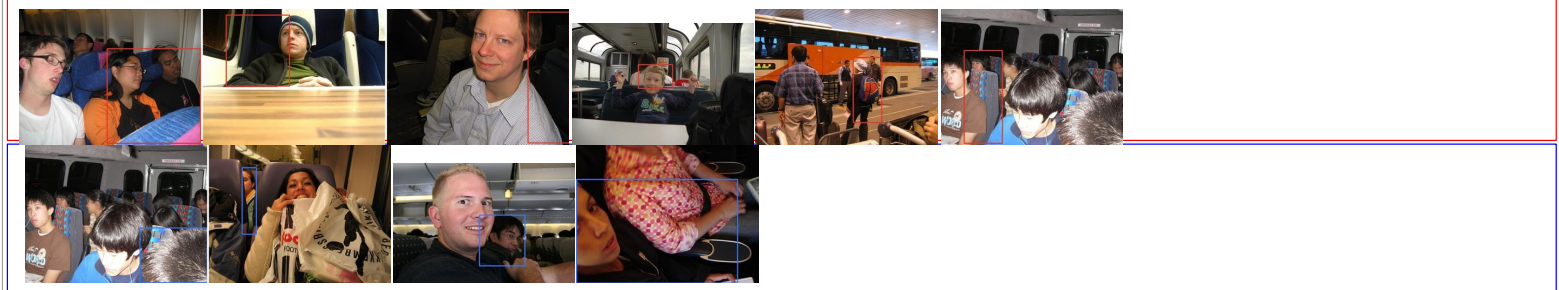

**Edge Case 4** | False Positives: Red bounding box ; False Negatives: Blue bounding box

**Explanation:** Common Feature: The scenarios involve various environments such as buses, trains, rooms, and outdoor settings with obstructions, limited visibility, and dim lighting conditions. False Positives: The primary reason for False Positives is occlusion caused by objects like jackets, bags, seats, and other people, leading to the model mistakenly identifying these objects as body parts of a person. False Negatives: The primary reason for False Negatives is occlusion caused by obstructions like seats, other people, and objects, as well as limited visibility of the person's face and body due to hair, clothing, and lighting conditions.

**Edge Case 5** | False Positives: Red bounding box ; False Negatives: Blue bounding box

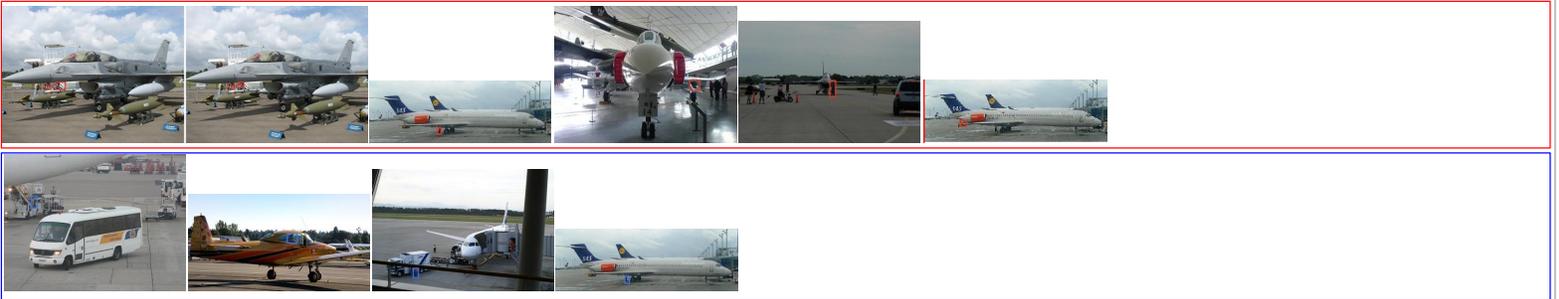

**Explanation:** : Common Feature of Environments: The scenarios involve various environments such as parking lots, airports, grassy fields, dimly lit rooms, and outdoor areas with different objects and occlusions. Primary Reason for False Positives: The primary reason for false positives is the misclassification of objects or structures, such as cars, propellers, buildings, tires, and airplanes, as persons due to their shape, size, color, or similarity to human body parts. Primary Reason for False Negatives: The primary reason for false negatives is the occlusion caused by other people, objects, or structures in the scene that obstruct the view of the person, making it difficult for the model to accurately detect the person's body or features.

**Edge Case 6** | False Positives: Red bounding box ; False Negatives: Blue bounding box

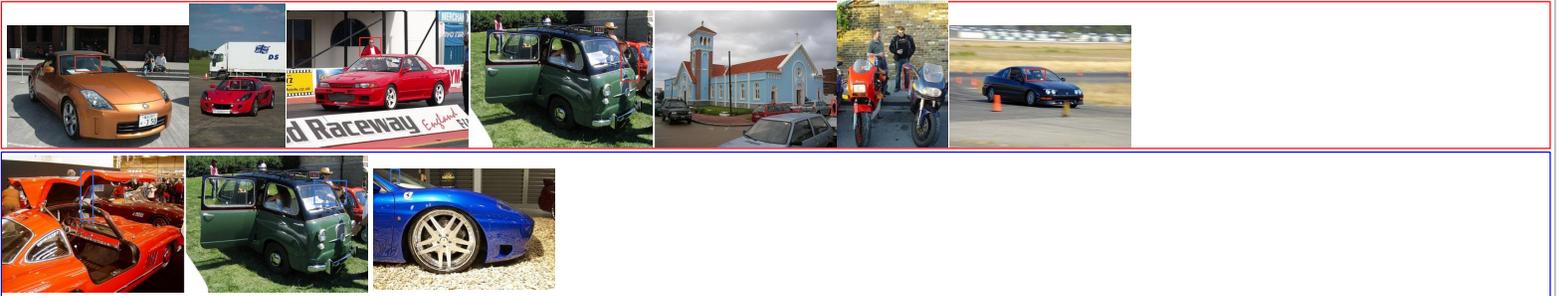

**Explanation:** The scenarios involve various outdoor settings, such as sunny days, parking lots, grassy fields, and cityscapes, with cars and people present. The most likely reason for false positives is the presence of objects or occlusions that resemble human characteristics, such as car seats, red objects, reflections, or shadows. Occlusion and partial visibility also contribute to the false positive detection. The most common reason for false negatives is occlusion caused by objects like car windshields, steering wheels, side mirrors, and other cars, obstructing the view of the person. Additionally, focus on specific body parts, such as the face or head, and dim lighting conditions can contribute to false negatives.

**Edge Case 7** | False Positives: Red bounding box ; False Negatives: Blue bounding box

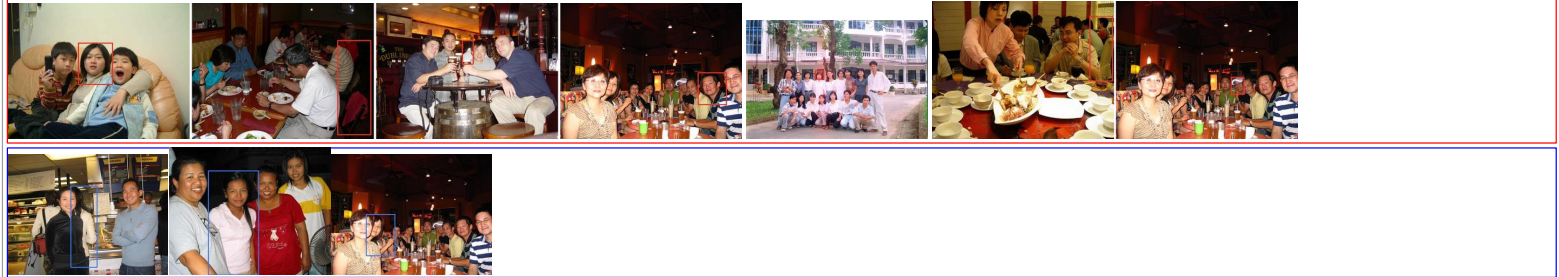

**Explanation:** The common feature in these environments is the presence of dining tables or food- related scenes, where people are sitting and enjoying meals together. The primary reason for false positives is occlusion caused by objects, other people, or obstructions in the image. The detection model mistakenly identifies parts of objects, such as tables, utensils, and bags, or visible body parts, like arms, legs, or heads, as people due to their similarity in appearance and prominence in the image. The primary reason for false negatives is occlusion caused by obstructions in the image, such as hands, glasses, food, or other objects, which block the view of the person's body or face. The focus on specific body parts, like upper bodies or faces, and the presence of blurry or visually dominating elements further contribute to the missed detections.

**Edge Case 8** | False Positives: Red bounding box ; False Negatives: Blue bounding box

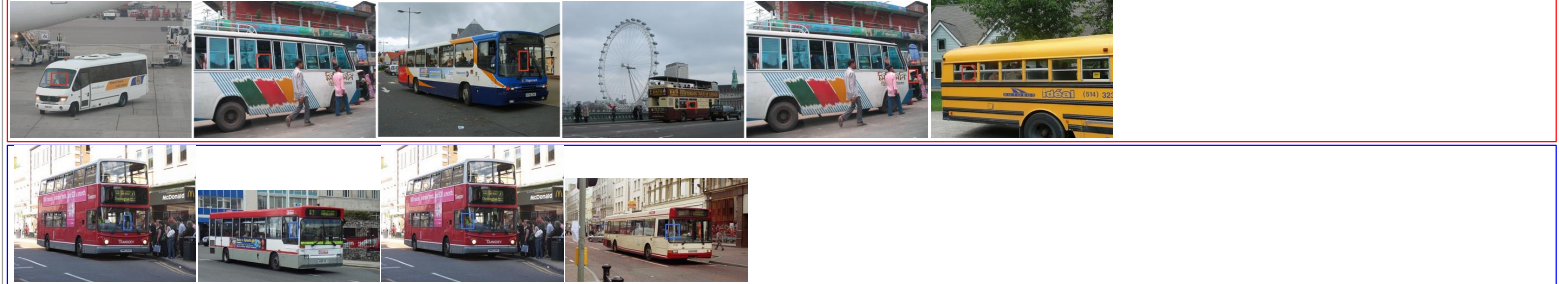

**Explanation:** Common Feature: The common feature in all the scenarios is the presence of a bus, car, or train, with various objects and occlusions in the environment. Primary Reason for False Positives: The primary reason for false positives is the spurious correlation between objects in the scene (e.g., reflective surfaces, poles) and the appearance of a person, combined with occlusion and partial visibility. Primary Reason for False Negatives: The primary reason for false negatives is the occlusion caused by windows, window frames, and other objects, obstructing the view of the person and making it challenging for the model to accurately detect them.

**Edge Case 9** | False Positives: Red bounding box ; False Negatives: Blue bounding box

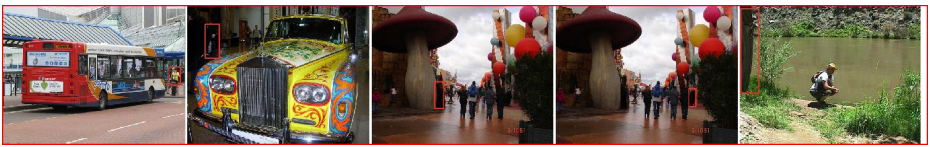

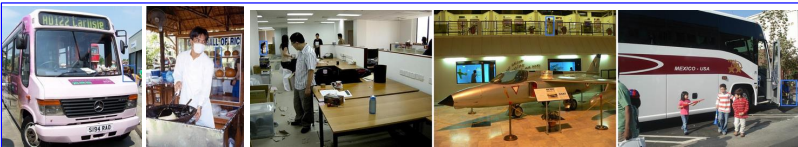

**Explanation:** : Common Feature of Environments: The common feature in these environments is the presence of occlusion, low lighting conditions, and partial visibility of the person's body. Most Primary Reason for False Positives: The most primary reason for the false positives is the misinterpretation of objects or occlusions resembling a person, such as a dog's collar, curtains, bottles, or other objects with similar shape or color to a person. Most Primary Reason for False Negatives: The most primary reason for the false negatives is the occlusion caused by objects, such as fences, handbags, cell phones, or other people, obstructing the view of the person's body and making it difficult for the model to accurately detect the complete person.

**Edge Case 10** | False Positives: Red bounding box ; False Negatives: Blue bounding box

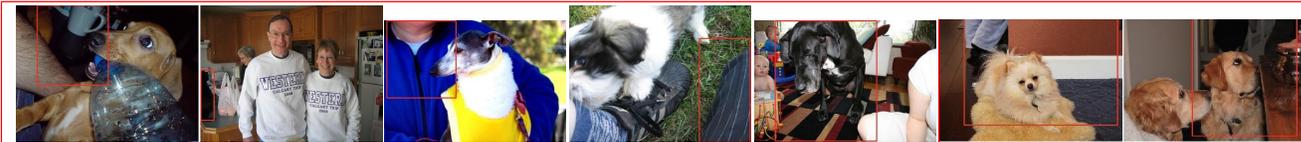

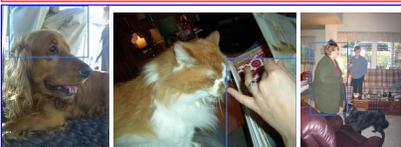

**Explanation:** Common feature: The common feature in these environments is the presence of dogs and their interaction with people. Most primary reason for False Positives: The most primary reason for False Positives is the confusion between the appearance, size, and shape of dogs and humans. The object detection model may mistake dogs for people due to similarities in body shape, size, and occlusion caused by the dog's presence. Most primary reason for False Negatives: The most primary reason for False Negatives is occlusion and partial visibility of the person's body caused by dogs. The presence of dogs obstructs the view of the person, making it difficult for the model to accurately detect them.

Edge Case 11 | False Positives: Red bounding box ; False Negatives: Blue bounding box

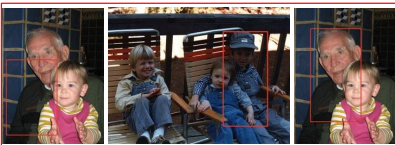

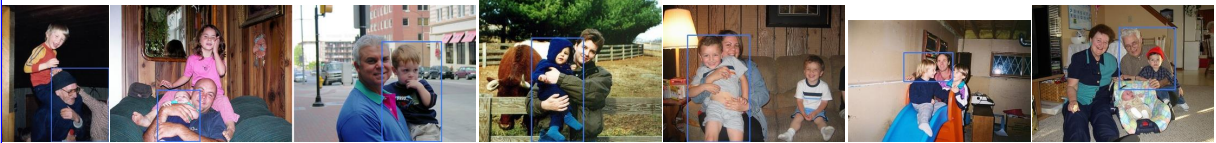

**Explanation:** Common feature of these environments: The environments in these scenarios involve people interacting with each other, often in close proximity, and with various occlusions and lighting conditions present. Primary reason for False Positives: The primary reason for False Positives in person detection is occlusion caused by objects or body parts obstructing the view of the person, leading to only partial detection and mistaken identification of non-person objects or body parts as a person. Primary reason for False Negatives: The primary reason for False Negatives in person detection is occlusion caused by objects, body parts, or other people obstructing the view of the person, making it difficult for the model to accurately detect and localize the person in the image.

Edge Case 12 | False Positives: Red bounding box ; False Negatives: Blue bounding box

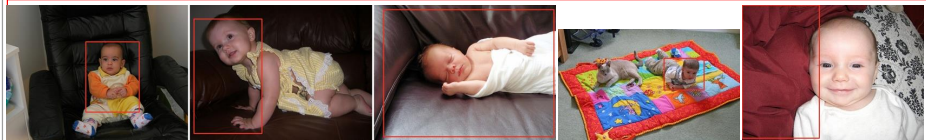

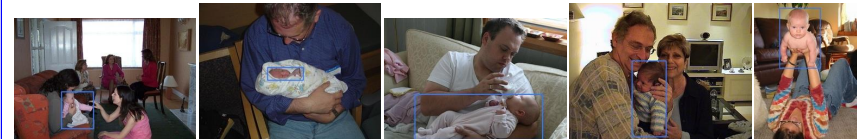

**Explanation:** The common feature in these environments is the presence of babies or young children, often in close proximity to adults, on couches or blankets, with various objects and toys in the surrounding area. The primary reason for the False Positives is the occlusion caused by the presence of blankets, toys, and other objects partially covering the babies or young children. This occlusion, along with the small size of the detected parts and similarity in color or shape between the objects and a person, leads to the misclassification as a person. The primary reason for the False Negatives is the occlusion caused by the presence of blankets, bottles, toys, and hands partially obstructing the view of the person. This occlusion makes it difficult for the model to accurately detect and localize the person's body and face.

Edge Case 13 | False Positives: Red bounding box ; False Negatives: Blue bounding box

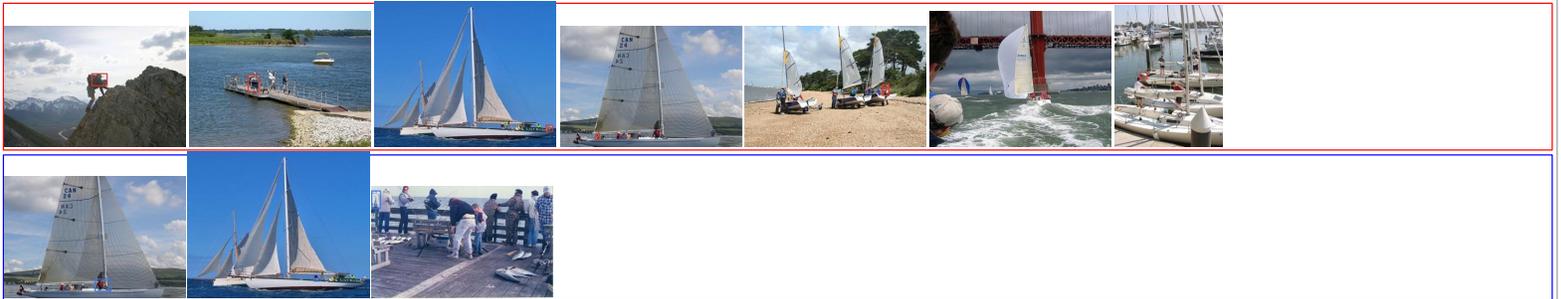

**Explanation:** The scenarios mostly involve outdoor settings with boats, water, and sunny or bright lighting conditions. - The primary reason for false positives in person detection is the presence of objects or patterns that resemble the appearance of a person, such as red shirts, helmets, life jackets, or boat structures. These similarities in color, shape, and size can trigger misclassifications. Primary Reason for False Negatives: - The primary reason for false negatives in person detection is occlusion and limited visibility of the person's body, often caused by objects, other people, or the angle of the photo. The lack of clear facial features or truncated body parts make it challenging for the model to accurately detect the entire person.

Edge Case 14 | False Positives: Red bounding box ; False Negatives: Blue bounding box

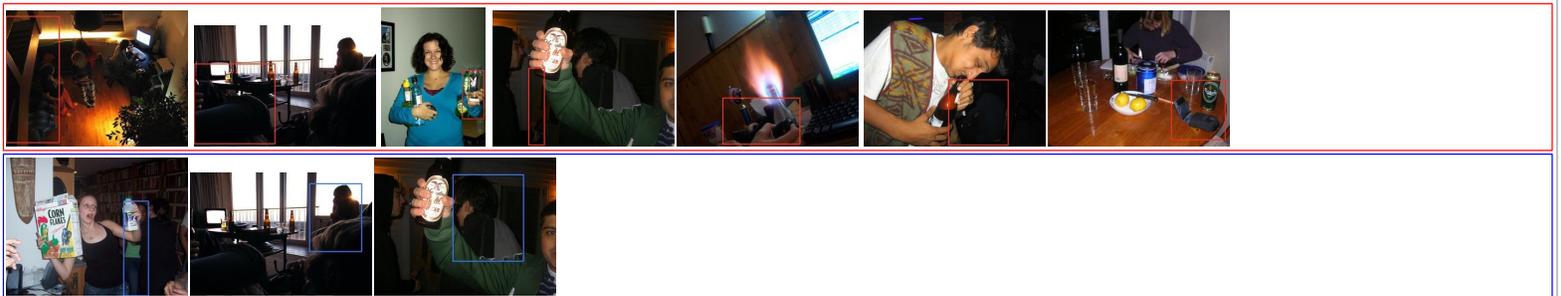

**Explanation:** Common Feature: - Many scenarios involve dark lighting conditions or partially obscured views of the person. Primary Reason for False Positives: - Occlusion by objects such as bottles, tables, or other people, leading to confusion in the detection model. Primary Reason for False Negatives: - Partial visibility of the person's body, with only specific body parts being detected, making it difficult for the model to accurately detect the complete person.

Edge Case 15 | False Positives: Red bounding box ; False Negatives: Blue bounding box

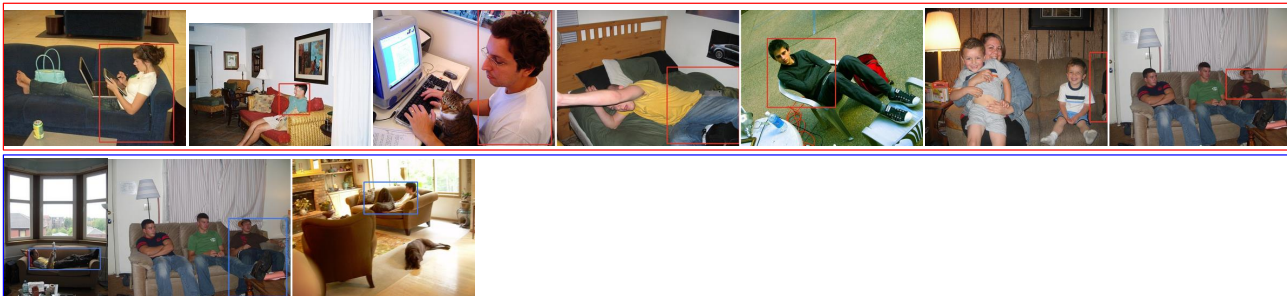

**Explanation:** Common Feature: Dimly lit rooms, presence of occlusions, and objects resembling body parts. Primary Reason for False Positives: Occlusion caused by objects such as pillows, couches, desks, and plants, leading to incomplete or ambiguous features for person detection models. Primary Reason for False Negatives: Occlusion caused by furniture, magazines, curtains, and fans, obstructing the view of the person's body, combined with low angle viewpoints and dim lighting conditions.

Edge Case 16 | False Positives: Red bounding box ; False Negatives: Blue bounding box

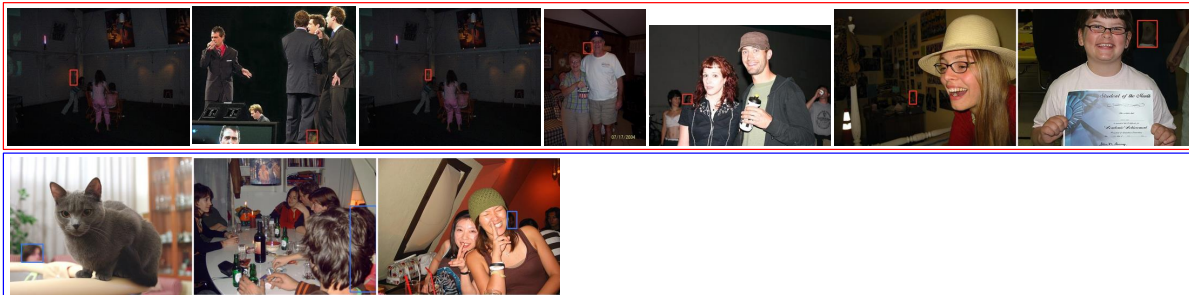

**Explanation:** All scenarios involve dark rooms or images with limited lighting, occlusion, and partial visibility of the person's head or body. The most primary reason for false positives is the model mistaking the partially visible head or body parts, along with occlusions and similarities in color or shape, as complete persons due to the lack of contextual information and accurate ground truth. The most primary reason for false negatives is the occlusion caused by objects or body parts, such as chairs, curtains, or hair, combined with limited visibility, blurry images, and dark lighting conditions, making it challenging for the model to detect the complete person.

**Edge Case 17** | False Positives: Red bounding box ; False Negatives: Blue bounding box

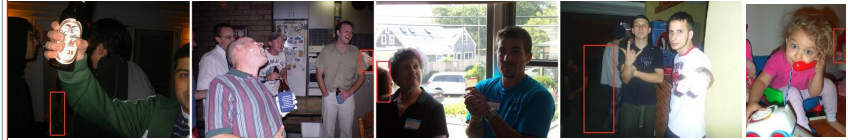

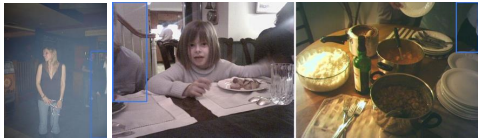

**Explanation:** The common feature in these environments is that they are all dimly lit rooms with various occlusions and obstructions present. The primary reason for false positives is the model mistakenly detecting objects such as clothing, scarves, hands, and body parts as complete persons due to their prominence in the image, lack of visibility of the rest of the body, and spurious correlations. The primary reason for false negatives is the occlusion caused by obstructions such as walls, objects, handbags, and other body parts, which prevent the model from accurately detecting the person's complete body.

**Edge Case 18** | False Positives: Red bounding box ; False Negatives: Blue bounding box

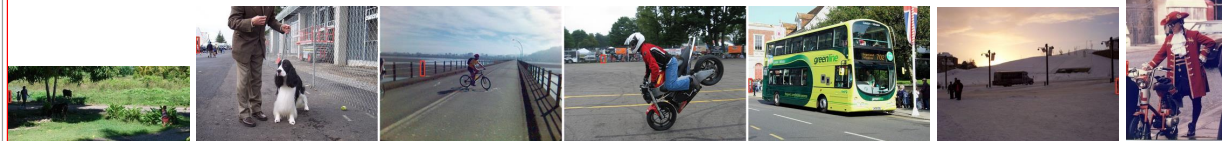

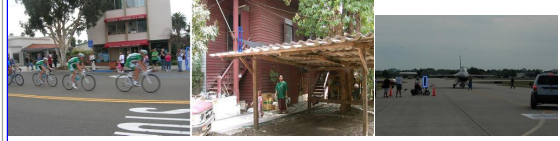

**Explanation:** Common Feature of Environments: - Many scenarios involve low lighting conditions, blurry backgrounds, and occlusion. Most Primary Reason for False Positives: - The main reason for false positives is the presence of occlusion, partial visibility of body parts, and similarity in color or shape between the main object and a person class. Most Primary Reason for False Negatives: - The primary reason for false negatives is occlusion, limited visibility of body parts, and focus on specific body parts rather than the entire body.

Edge Case 19 | False Positives: Red bounding box ; False Negatives: Blue bounding box

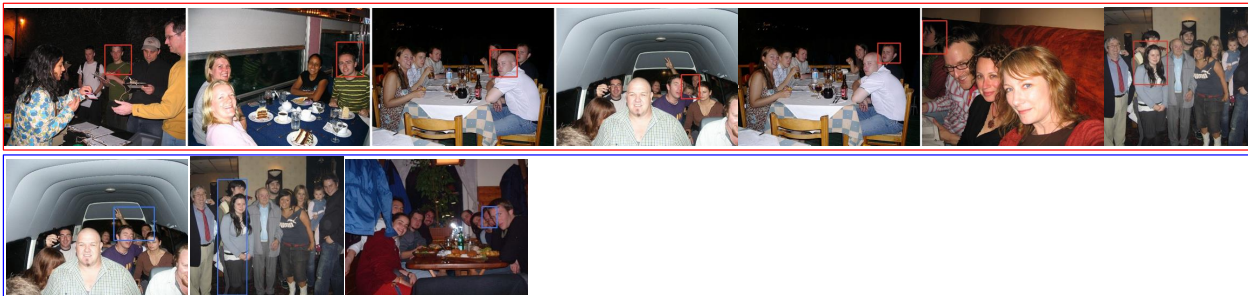

**Explanation:** - The scenarios involve various indoor settings, often with dim lighting conditions and occlusions caused by objects, other people, or facial features like glasses or facial hair.
Primary reason for False Positives: - Occlusion and limited visibility of the person's body or face, leading to the model mistakenly detecting partial features or objects as people. Primary reason for False Negatives: - Occlusion and obstruction caused by objects, other people, or the person themselves, preventing the model from accurately detecting the person's full body or face.

Edge Case 20 | False Positives: Red bounding box ; False Negatives: Blue bounding box

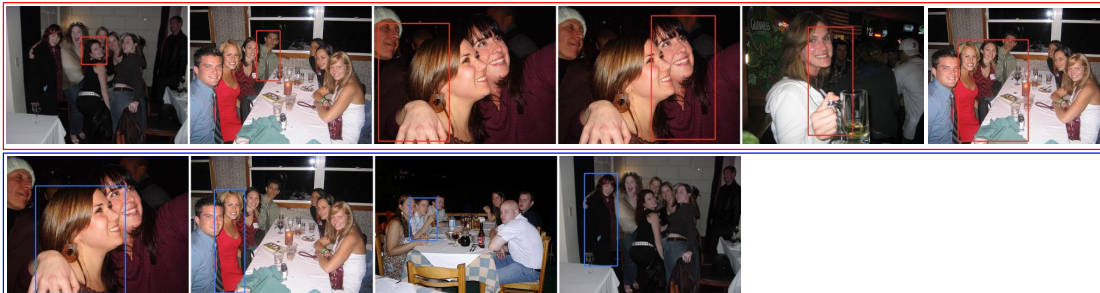

**Explanation:** The common feature among these environments is that they involve group settings with multiple people present, often posing for pictures or sitting together. The most primary reason for false positives is occlusion, where the presence of other people or objects obstructs the view of the ground truth person, leading to the model mistakenly detecting parts of the person as a separate entity. The most primary reason for false negatives is also occlusion, where other people or objects block the view of the ground truth person, making it difficult for the model to accurately detect the complete person. Additionally, factors such as dim lighting, viewpoint, and partial visibility of the person contribute to false negatives.

**Edge Case 21** | False Positives: Red bounding box ; False Negatives: Blue bounding box

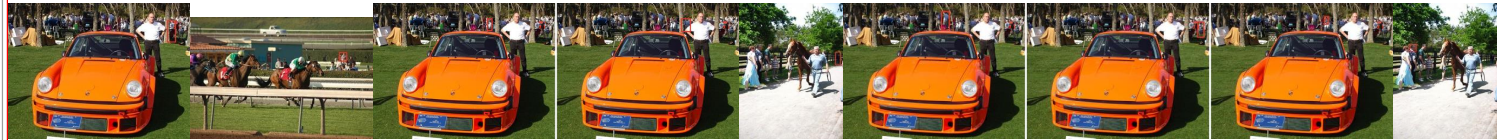

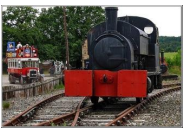

**Explanation:** Common Environment Feature: - The scenarios generally involve outdoor settings, such as parks or grassy areas, with a mix of people and objects in the background. Most Primary Reason for False Positives: - Occlusion and limited visibility of the person's body, leading to misinterpretation by the object detection model. Most Primary Reason for False Negatives: - Partial visibility of the person's body, particularly the upper body or head, leading to a misclassification by the object detection model.

**Edge Case 22** | False Positives: Red bounding box ; False Negatives: Blue bounding box

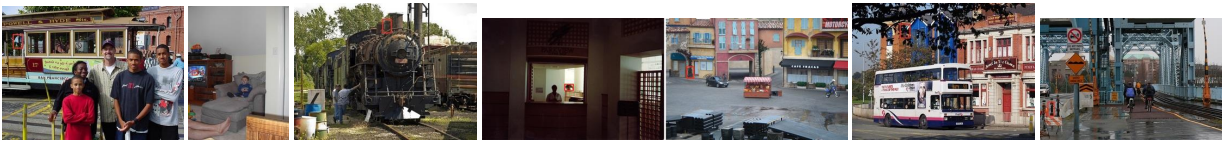

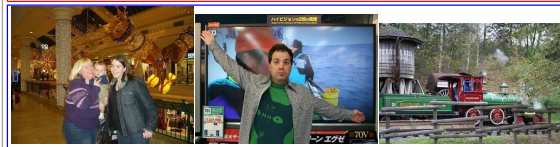

**Explanation:** Common Feature: - The scenarios involve various objects or backgrounds that share visual similarities or correlations with a person, such as shape, color, texture, or occlusion. Primary Reason for False Positives: - The primary reason for the false positives is the spurious correlation between the objects or backgrounds and human features, leading to misclassification by the object detection model. Primary Reason for False Negatives: - The primary reason for the false negatives is the occlusion or partial visibility of the person, which makes it difficult for the object detection model to accurately detect the entire person.

**Edge Case 23** | False Positives: Red bounding box ; False Negatives: Blue bounding box

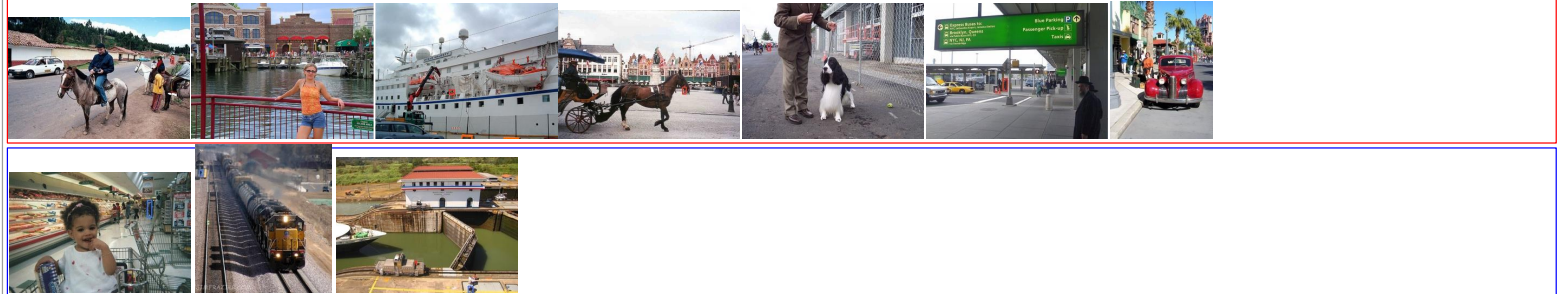

**Explanation:** Many of the false positives and false negatives occur in scenarios with partial visibility of the person, occlusion, blurriness, and similarity in color or texture between the person and the background or other objects. The primary reason for false positives is the combination of partial visibility, occlusion, and similarity in color or texture between the person and the background or other objects, leading to misinterpretation by the object detection model. The primary reason for false negatives is the occlusion caused by objects or body parts obstructing the view of the person, making it difficult for the model to accurately detect and recognize the person as a valid person object.

**Edge Case 24** | False Positives: Red bounding box ; False Negatives: Blue bounding box

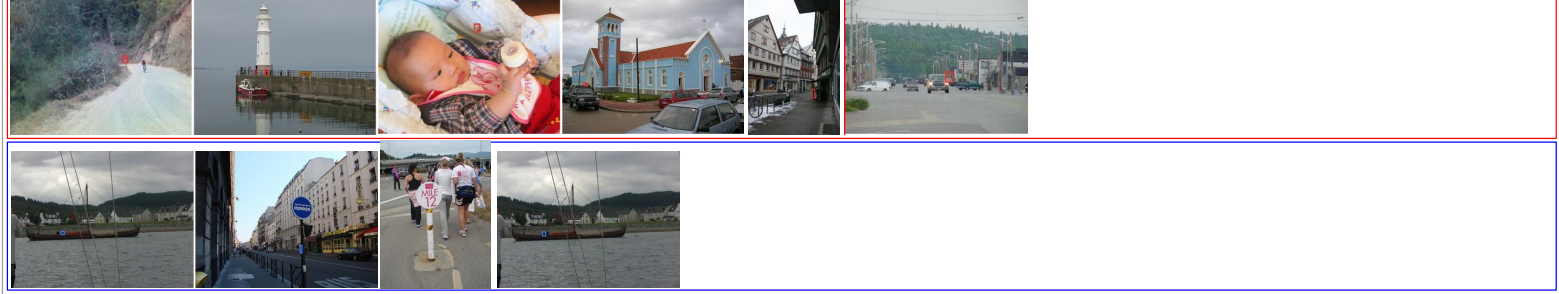

**Explanation:** Many of the scenarios involve partially visible persons, occlusion, blurry images, dark lighting conditions, and focus on specific body parts or objects. The primary reason for false positives is the spurious correlation between the characteristics of objects or backgrounds (such as color, shape, or patterns) and the appearance of a person, leading to misclassification by the object detection model. The primary reason for false negatives is the obstruction of the person's body or face, blurry or dark images, limited visibility of body parts, and the focus on specific objects or body parts, making it difficult for the model to accurately detect and classify the person.

# Individual Explanation Examples

We present individual explanation examples for the four datasets used in the paper: Dog detector, Car detector, Chair detector, and Person Detector.

Each dataset includes False Positives and False Negatives, with obviously incorrect explanations marked as "Incorrect".

## Dog Detector
- Individual Explanations: 8 False Positives and 4 False Negatives.

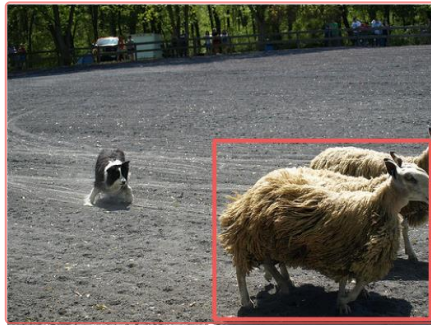

The scenario is an image featuring two sheep, one in the foreground with long curly wool and the other in the background, standing on a dirt road in a sunny and bright environment. The most possible reason for the False Positive dog detection is that the sheep in the foreground with its long curly wool may have been mistakenly identified as a dog due to its appearance and size, leading to an incorrect classification.

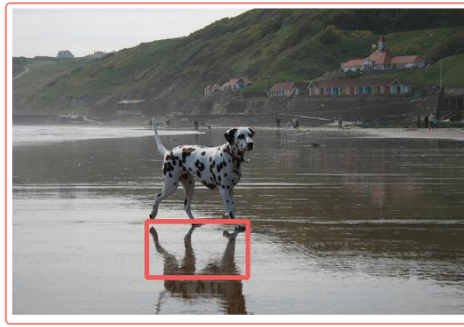

The scenario is a beach scene with a Dalmatian dog walking and its reflection in the water. The most possible reason for the False Positive dog detection is that the detection model mistakenly identified the dog's reflection as a separate dog due to its similarity in appearance, size, and position, leading to a false positive detection.

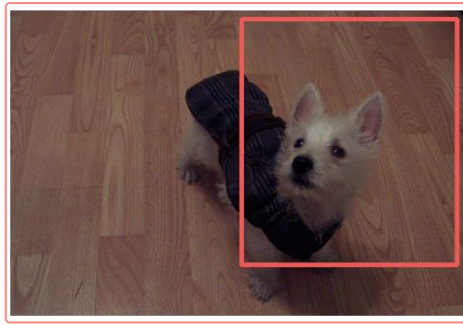

The scenario is a false positive dog detection in an image of a small white dog wearing a sweater on a wooden floor, where only parts of the dog's face are detected. The most possible reason for the false positive detection is the occlusion caused by the dog wearing a sweater, which partially covers its body and obstructs the view of the rest of the dog.

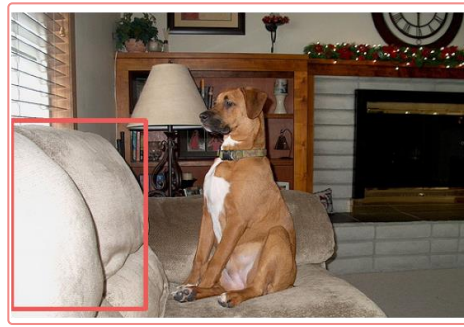

The scenario is a cozy living room with a dog sitting on a couch, surrounded by books and a bookshelf in the background. The most possible reason for the False Positive dog detection is that the dog's presence and position on the couch, combined with the background of books and the cozy atmosphere, might have led the object detection model to erroneously classify other objects, such as the couch or books, as dogs.

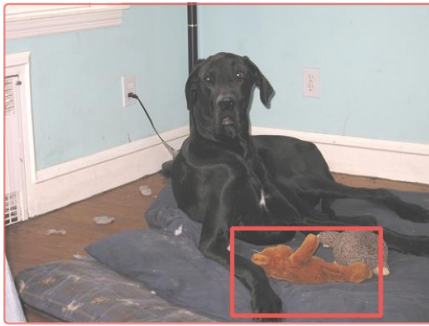
The scenario involves a false positive dog detection in an image of a black dog and a brown teddy bear on a bed. The most possible reason for the false positive detection is that the dog and teddy bear exhibit similar color and texture characteristics, causing the dog detection model to mistakenly identify the teddy bear as a dog. Additionally, the occlusion and partial visibility of the teddy bear may have contributed to the confusion.

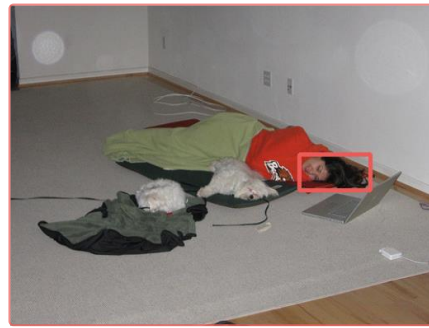
The scenario depicts a woman lying on a red surface, wearing a red hoodie, in a dimly lit room with a laptop nearby, giving off a relaxed atmosphere. The most likely reason for the False Positive dog detection is the presence of the woman's head, which may have features that resemble a dog's face, such as the hair and the top of the head, leading to a misclassification by the object detection model.

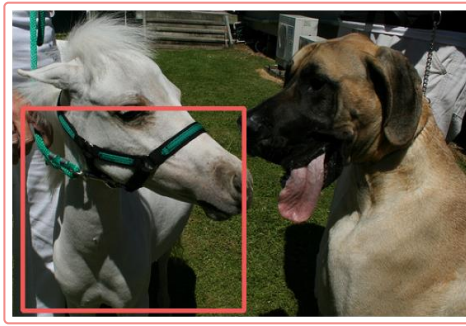
The scenario depicts a false positive detection of a dog in an image of a horse with a harness, two dogs, and a person in a grassy area, possibly due to the presence of a small white horse-like dog and the similarity in appearance between the horse and a dog, leading to misclassification by the dog detection model.

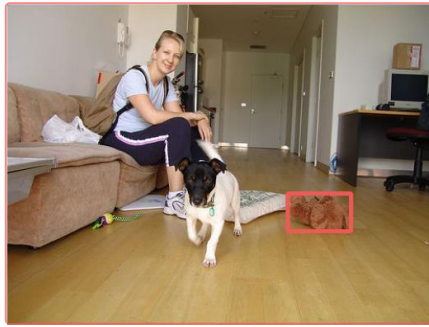
(**Incorrect**) The scenario depicts an image of a brown teddy bear on a wooden table, where a False Positive dog detection occurred due to the presence of a brown piece of food partially covered by a brown piece of paper, leading to a misclassification by the object detection model. The reasons for the False Positive may include the similarity in color and texture between the food and a dog, as well as the partial occlusion caused by the paper.

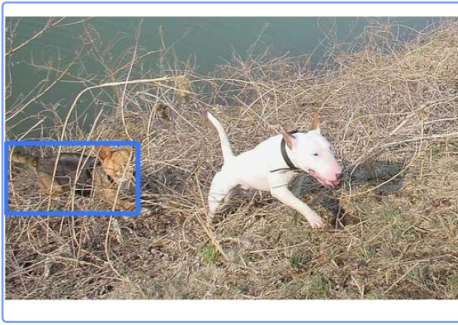
The scenario depicts a dog lying down in a grassy area, partially hidden by tall grass and bushes. The most possible reason for the False Negative in the dog detection is that the dog's body and face are partially obscured by the surrounding vegetation, making it difficult for the detection model to accurately identify and detect the dog. The dim lighting conditions and the similarity of the dog's fur color to the surrounding grass further contribute to the difficulty in detecting the dog.

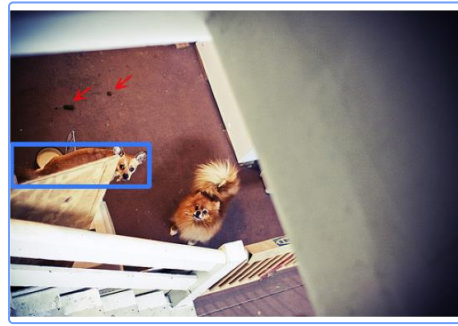
The scenario involves a False Negative in a dog detection model, where a small brown dog's face is obstructed by a wooden object, resulting in the model failing to detect it. The possible reasons for the False Negative are: occlusion caused by the wooden object, partial visibility of the dog's body, dark lighting conditions, and the focus on the close-up of the dog's face instead of the entire dog.

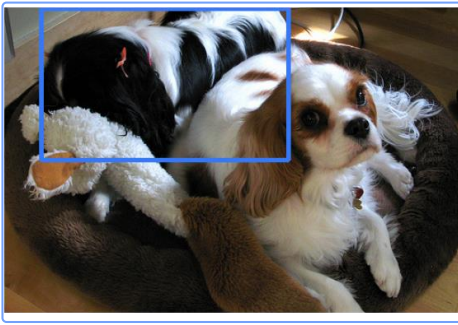
The scenario involves a cropped image of two dogs lying down on a couch, with one dog partially obstructing the view of the ground truth dog, a Cocker Spaniel, which is lying down next to a stuffed animal. The most possible reason for the False Negative dog detection is the obstruction caused by the other dog and the stuffed animal, which partially covers the ground truth dog's face and body, making it difficult for the model to detect it accurately.

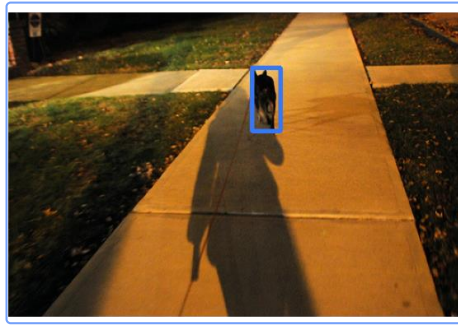
The scenario depicts a cropped image of a black dog walking on a sidewalk, with a red rope obstructing the view of the dog from the front, and the dog's shadow visible on the wall. The most possible reason for the False Negative in dog detection is the focus on the dog's rear end, making it difficult to see the dog's face and other body parts, combined with the occlusion caused by the red rope.

## Car Detector
- Individual Explanations: 12 False Positives and 4 False Negatives.

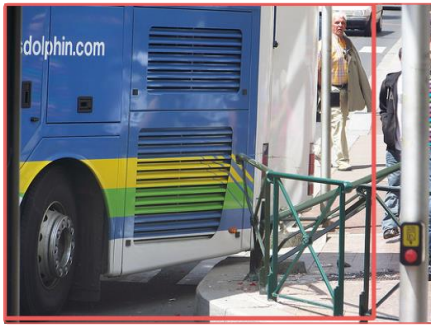

The scenario is a sunny day with a blue and yellow bus parked on the side of a street, a man standing next to it, and cars parked nearby. The most possible reason for the False Positive car detection could be the similarity in color and shape between the bus and a car, leading to the model mistakenly detecting the bus as a car.

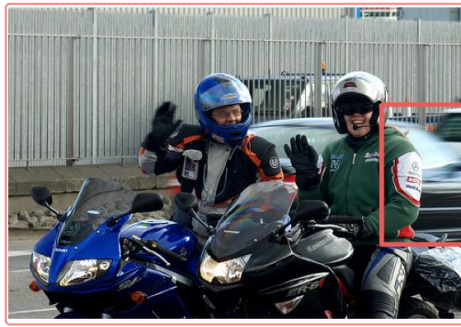

The scenario involves a False Positive detection of a car in an image of a person riding a motorcycle, and the most possible reason for the False Positive is the presence of a partially visible car in the background, coupled with the spurious correlation between the car logo on the person's jacket and the ground truth car. The size, viewpoint, and localization error might have also contributed to the False Positive.

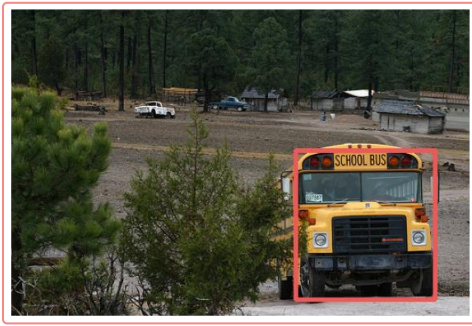

The scenario depicts a yellow school bus parked on a dirt road with people, trucks, and a small building in the background. The most possible reason for the False Positive car detection is that the model may have mistakenly identified the bus as a car due to similarities in shape, size, and color, as well as the presence of headlights on the bus. Additionally, occlusion and the partial visibility of the bus may have contributed to the incorrect detection.

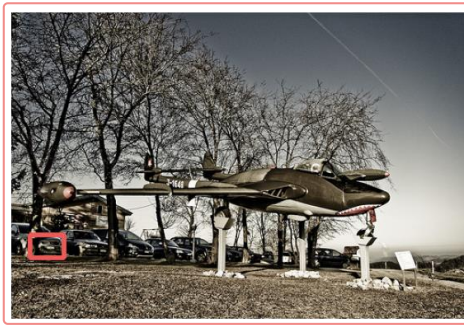

The scenario is a car detection model detecting a false positive of a car in an image of a parking lot, where the car is partially obstructed by a tree and only parts of the car, specifically the car hood, are visible. The most possible reason for the false positive detection is that the model is mistaking the visible car hood for a full car due to its shape and similarity to a car's front end.

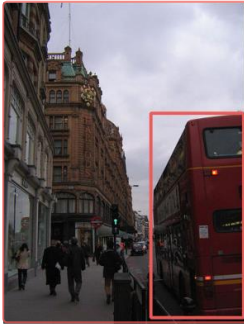
The scenario is that a false positive car detection occurred in an image of a red double-decker bus driving down a city street. The most possible reason for the false positive detection is that the object detection model mistakenly identified the bus as a car due to similarities in shape, size, and appearance, and potentially due to occlusion of certain parts of the bus or errors in localization.

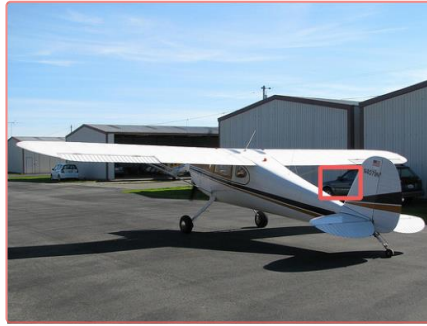
The scenario involves a False Positive detection of a car in a parking lot, where only parts of the car are detected, and the car is obstructed by the wing of a plane. The most possible reason for the False Positive detection could be the occlusion caused by the airplane wing, which partially covers the car and may confuse the object detection model. Additionally, the limited visibility of the car and the presence of other objects in the background, such as the truck, could also contribute to the False Positive.

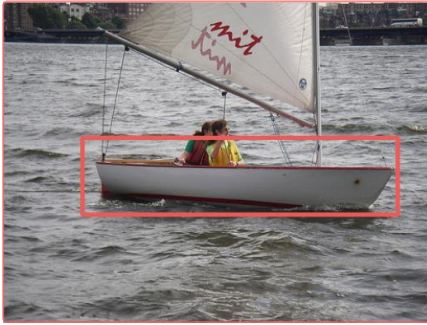
The scenario depicts a sunny day where a man and a woman are enjoying a boat ride on the water. The most possible reason for the False Positive car detection is that the boat's shape and the person's sitting position in the boat might resemble the shape and size of a car, leading to a misclassification by the object detection model. Additionally, the presence of a steering device held by the person might further contribute to the misclassification.

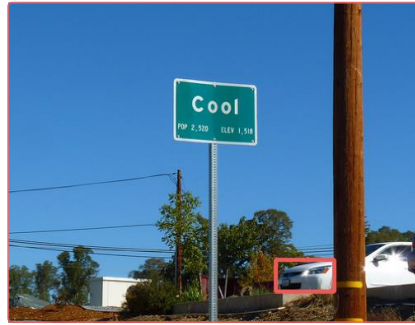
The scenario is a sunny outdoor image featuring a white car parked next to a pole, with the side view mirror partially detected as a false positive due to the bright reflection caused by the sun shining on the mirror and the car's window, creating a visually striking scene that can affect the visibility of the detected parts of the car.

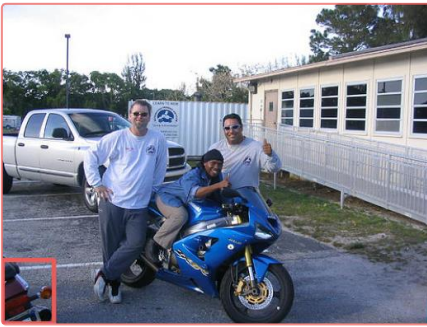
The scenario involves a false positive car detection in an image of a red motorcycle parked in a parking lot. The most possible reason for the false positive car detection could be the presence of spurious correlation, where the detection model mistakenly associates certain characteristics of the motorcycle, such as its red color and visible tail light, with those of a car.

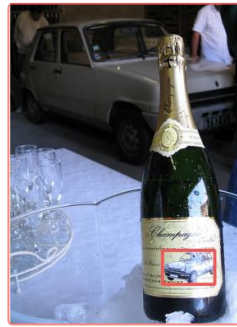
The scenario involves a False Positive detection of a car in an image featuring a bottle of champagne with a car image on the label, where the car appears to be a small, silver convertible, and is partially visible. The most possible reason for the False Positive detection is that the car image on the champagne bottle label may have similarities with the car detection model's training data, leading to a spurious correlation and a false positive detection.

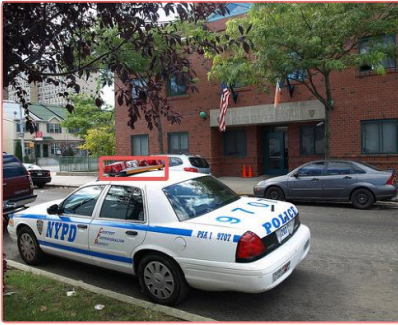
(**Incorrect**) The scenario involves a False Positive detection of a car in an image containing a main object of a train, parked on a snowy surface, with only a portion of it visible. The most possible reason for the False Positive detection is the similarity in color and partial overlap between the train and the main car, as well as the presence of other cars and a person in the scene, potentially causing confusion in the object detection model.

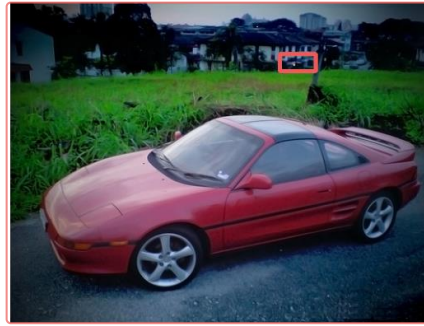
(**Incorrect**) The scenario is a false positive car detection in an image of a house with a clock tower, where the car parked in front of the house is mistakenly detected as a car. This false positive may have occurred due to the small size of the car in the image, the presence of other objects (such as the clock tower), and possibly unclear or inaccurate ground truth data.

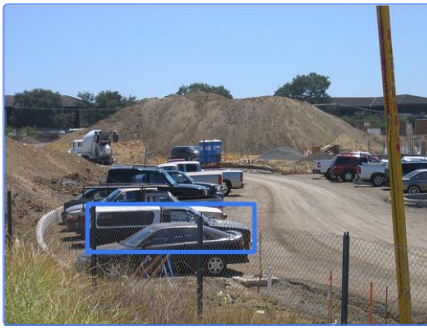
The scenario involves a car detection model that has produced a False Negative, where the ground truth car is not detected correctly. The most possible reason for this False Negative is the occlusion caused by the chain-link fence obstructing the car, which prevents the model from accurately detecting the car's presence and features. The focus on the mesh grill, which is the main subject of the photo, further adds to the difficulty in detecting the entire car.

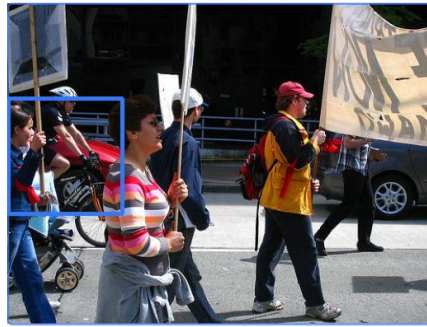
The scenario depicts a partially visible red sports car parked in the background of a crowded protest scene, with the most likely reason for the False Negative car detection being occlusion caused by the presence of a bicycle and a person holding a sign, obscuring the car from the model's view.

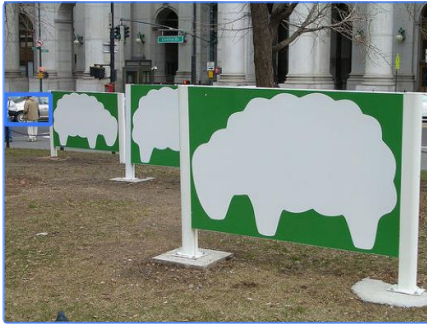
The scenario is that a white van parked on the side of the road is partially obscured by a man and a sign, resulting in a False Negative in the car detection model. The most possible reason for the False Negative is that the car is partially hidden and the detected parts are small and out of focus, making it difficult for the model to accurately detect the car.

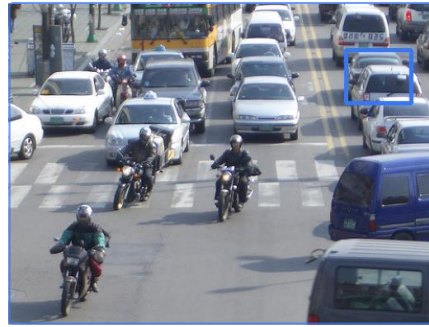
The scenario depicts a sunny day with a congested highway, where a silver compact car is obstructed by a white car parked in front of it, resulting in a False Negative in the car detection model. The most possible reason for the False Negative is the occlusion caused by the white car, which hides most of the ground truth car, making it difficult for the model to detect the complete car accurately.

## Chair Detector
- Individual Explanations: 16 False Positives and 8 False Negatives.

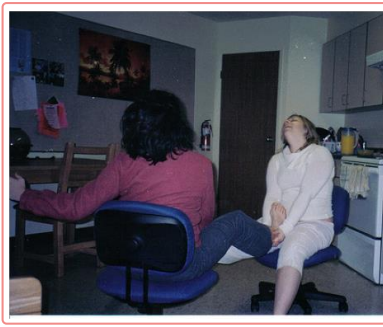

The scenario is a black and white image of a kitchen with a partially visible white refrigerator next to a wall, and various objects on the countertop including a bowl, a bottle, a cup, and a spoon. The most possible reason for the False Positive chair detection could be the presence of objects with similar shape or texture to a chair, such as the bowl or the refrigerator, leading to confusion in the object detection model.

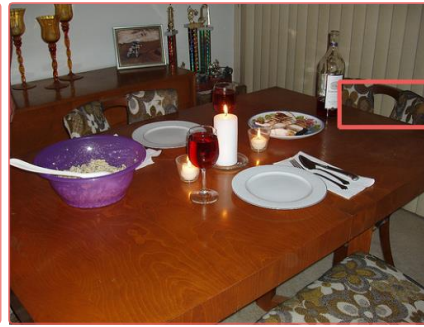

The scenario depicts a cozy dining setting with a partially visible chair obstructed by a table and a bottle of wine, leading to a False Positive detection in the chair detection model. The most likely reason for the False Positive is the occlusion caused by the table and wine bottle, which prevents the model from accurately detecting the complete chair and instead detects only the partially visible backrest.

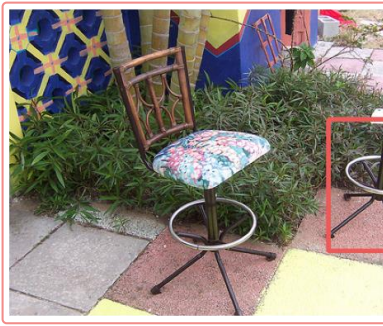

The scenario is an image featuring a black metal stand with a circular base surrounded by plants, where a False Positive detection of a chair has occurred. The most possible reason for the False Positive detection could be the presence of a black metal stand with a circular base, which shares visual similarities with a chair, such as shape and material, leading to confusion in the object detection model.

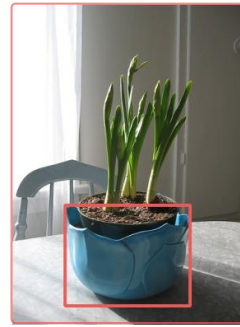

The scenario depicts an image of a blue vase filled with plants on a dining table, where a False Positive detection of a chair has occurred. The most possible reason for the False Positive detection could be the presence of a blue flower pot, which may have a similar color and shape to a chair, leading to confusion in the object detection model.

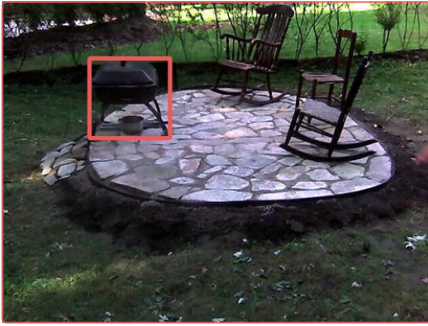

The scenario depicts a false positive detection of a chair in an image of a barbecue grill on a patio surrounded by potted plants. The most possible reason for the false positive detection could be the presence of visual features, such as the shape and color of the grill, that resemble a chair and lead to a misclassification by the chair detection model.

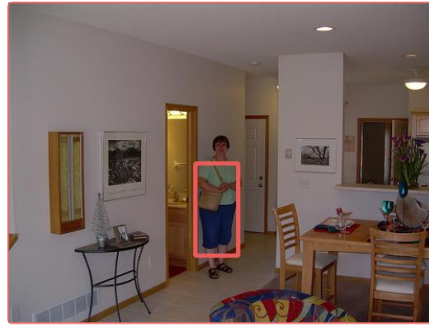

The scenario is a close-up image of a person wearing blue jeans and a green shirt, holding a brown purse, with a kitchen environment in the background featuring a dining table and a chair near it, a bottle on the counter, and a cup close to the sink. The most possible reason for the False Positive CHAIR detection is that the model misinterpreted the shape and color of the person's legs and the surrounding objects as a chair.

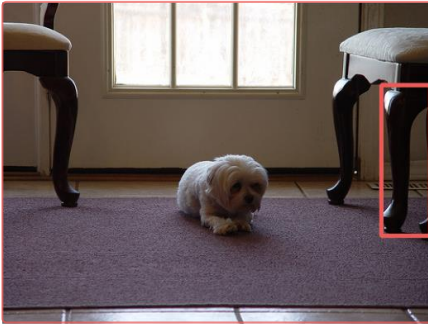

The scenario involves a false positive chair detection in an image of a wooden pole, where the main object is a pole leaning against a wall. The most possible reason for the false positive detection is that the chair detector model mistakenly identified the top part of the pole as a chair due to similar shape and size characteristics, as well as the presence of dark wood texture.

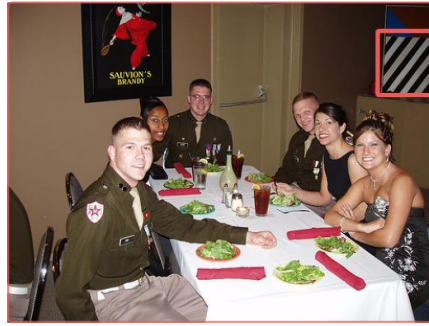

The scenario involves a False Positive detection of a chair in an image featuring a striped pattern and a brown wall with a red candle on a shelf; the most possible reason for the False Positive detection is the presence of the striped pattern in the image, which visually resembles the shape and structure of a chair, leading to a misclassification by the chair detection model.

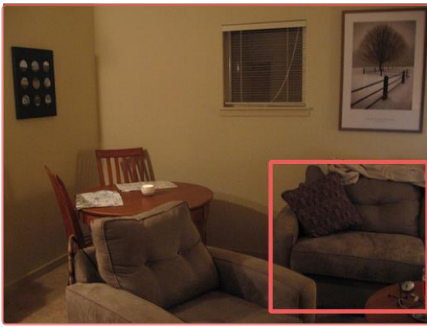 The scenario is a cozy living room with a couch, chair, and other objects. The most possible reason for the False Positive CHAIR detection is that the chair's size and positioning, coupled with the presence of a pillow on the couch and a blanket on the back, may have created visual cues that resemble a chair, leading to a misclassification by the chair detection model. The background and lighting conditions may have also contributed to the false positive.

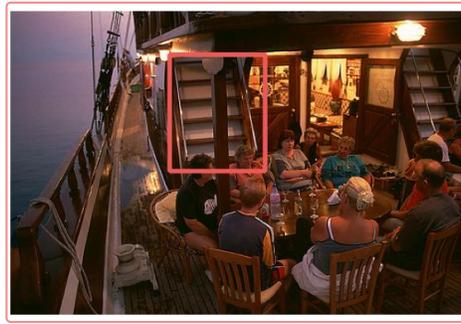 The scenario depicts a group of people gathered around a dining table in a restaurant or boat, with a staircase and pole nearby. The most possible reason for the False Positive chair detection is that the chair model mistakenly identified elements of the staircase, such as the wooden steps or the metal railing, as chairs due to similarities in shape, color, or texture. Additionally, occlusion from objects and people in the scene may have contributed to the confusion.

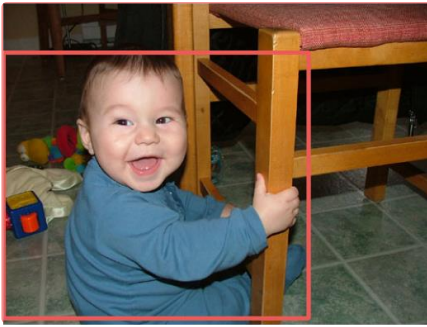 The scenario is a baby sitting on a wooden chair, with the chair partially obstructed by the baby's body and the main focus of the image being the baby and the chair. The most possible reason for the False Positive detection of a chair is that the detection model detected parts of the chair (backrest and armrest) but failed to detect the complete chair due to occlusion and the presence of the baby.

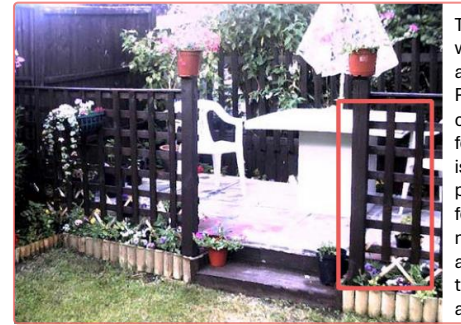 The scenario is an image with a wooden fence, chairs, potted plants, and a white building, where a False Positive detection of a chair has occurred. The most possible reason for the False Positive chair detection is that the chair is partially visible and positioned in front of the wooden fence, causing the detection model to mistakenly classify parts of the fence as the chair due to similarities in color, texture, and shape between the chair and the fence.

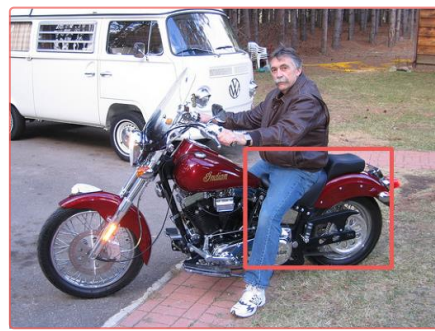
The scenario involves a false positive detection of a chair in an image of a man sitting on a red motorcycle parked on a sidewalk, with a grass background and a car and truck in the distance. The most possible reason for the false positive chair detection could be due to spurious correlation, where certain visual features in the motorcycle, such as its shape or color, might be correlating with the chair class in the model's training data.

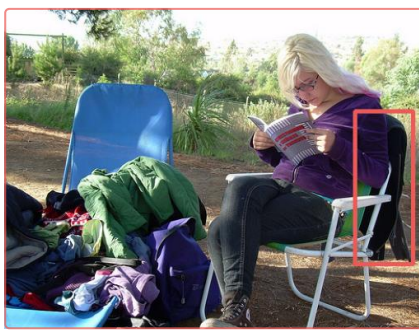
The scenario depicts a woman sitting on a white chair outdoors, with a grassy background and a black handbag nearby. The most possible reason for the False Positive chair detection could be the partial visibility of the chair, as it is positioned next to the person and only the back of the chair is visible, making it challenging for the object detection model to accurately identify the chair.

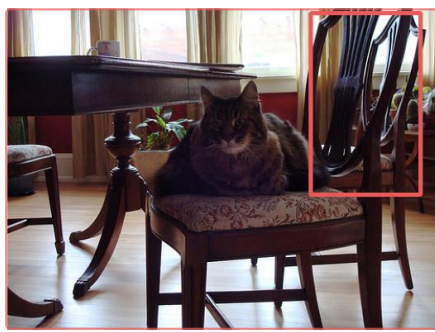
(**Incorrect**) The scenario involves a false positive detection of a chair in an image of a cat sitting on a chair, partially blocking the view of the chair, in a dimly lit room with a potted plant next to it. The most possible reason for the false positive detection is the presence of the cat, which obstructs the view of the chair and causes the model to mistake the cat as a chair. Additionally, the dim lighting and partial visibility of the chair further contribute to the false positive.

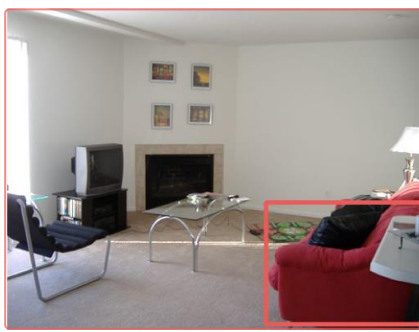
(**Incorrect**) The scenario is a cozy living room with a red leather chair placed next to a red couch, partially occluded, with a black pillow on it, surrounded by smaller black chairs, a dining table, and a TV, with a green carpet, dim lighting, and a sense of relaxation. The most possible reason for the False Positive CHAIR detection is the presence of the red leather chair next to the red couch, which may have led the model to mistakenly identify the couch as a chair. Additionally, the partial occlusion of the red leather chair by the couch might have contributed to the misclassification.

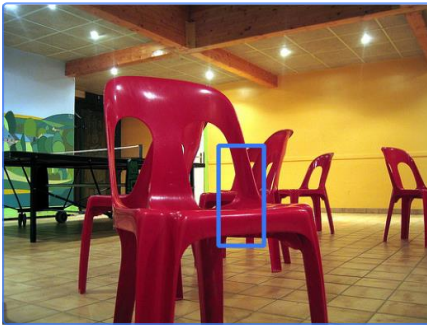 The scenario involves a red chair that is partially visible in a room with other chairs and a dining table, and the most possible reason for the False Negative chair detection is the obstruction caused by another red chair that is blocking the view of the ground truth chair, preventing it from being detected accurately by the model.

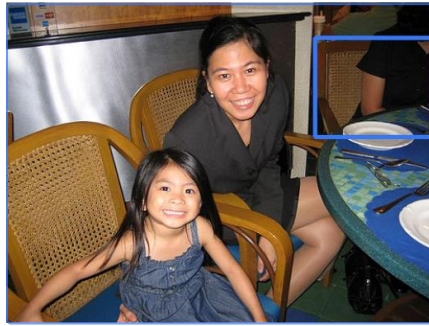 The scenario involves a cropped image of a woman sitting at a dining table, with a partially obstructed chair in the background. The most possible reason for the False Negative chair detection is that the focus of the image is on the person and the table setting, with the chair's back and sides not clearly visible. The detection model may have struggled to identify the chair due to its limited visibility and occlusion by the table.

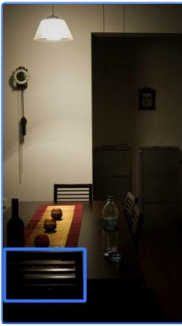 The scenario involves a small, partially obscured black chair in a dark room, obstructed by a red and yellow tablecloth and table runner, leading to a False Negative in chair detection. The most possible reason for the False Negative is that the chair's small size, partial occlusion, dark lighting conditions, and similarity in color to other objects in the image make it difficult for the model to accurately detect the chair.

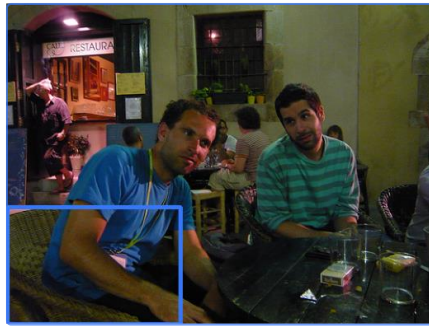 The scenario involves a cropped image of a man sitting on a partially visible chair with a woven, basket-like design, and the most possible reason for the False Negative chair detection is the obstruction caused by the man's arm covering the back of the chair, making it difficult for the model to detect the chair accurately.

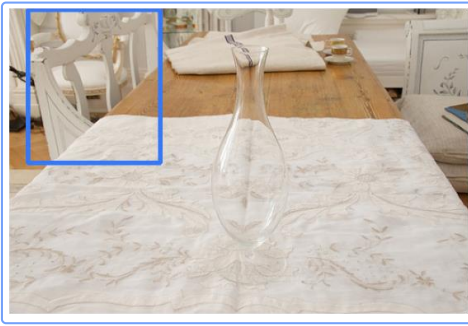
The scenario involves a partially obstructed white chair with a wooden frame and a design on the back, which is partially covered by a white cloth on a table set with various items, including a wine glass and a cup. The most possible reason for the False Negative chair detection is the occlusion caused by the tablecloth and glass, which obstruct the view of the chair's design and structure.

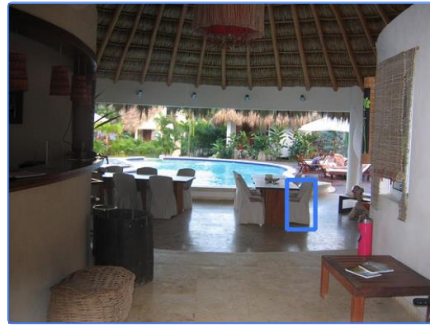
The scenario is a cropped image of a kitchen with a white chair partially obstructed by a towel, a square-shaped wooden table near the chair, and a counter nearby. The most possible reason for the False Negative chair detection is the occlusion caused by the towel, which covers a significant portion of the chair, making it difficult for the model to accurately detect the chair.

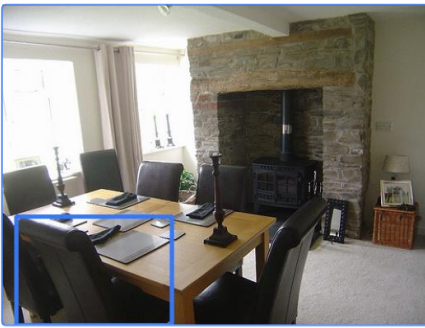
The scenario depicts a cropped image of a dining table with a partially visible black leather chair, obstructed by a potted plant, where the chair's appearance, color, and position can be discerned, leading to a False Negative in chair detection due to occlusion caused by the plant, partial truncation of the chair, and potentially inaccurate ground truth.

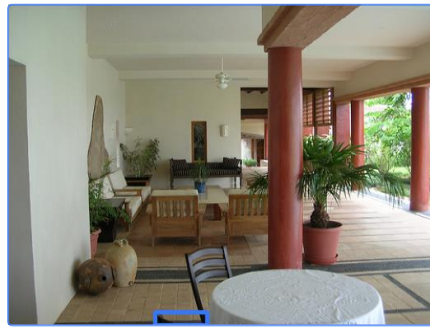
(**Incorrect**) The scenario depicts a False Negative in chair detection, where a partially visible chair in a room with a wall is not correctly detected by the model. The most possible reason for this False Negative could be the occlusion caused by the wall, which obstructs the view of the chair and prevents the model from accurately detecting it. The chair's size, orientation, and partial visibility also contribute to the False Negative.

## Person Detector
- Individual Explanations: 24 False Positives and 8 False Negatives.

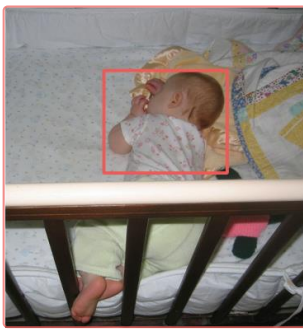

The scenario involves a false positive person detection where a baby girl sleeping on a bed is mistakenly detected as a person, likely due to occlusion by a blanket and pillow, dim lighting conditions, the baby's face being covered, the baby's body not being fully visible, and the size of the baby being small compared to the bed.

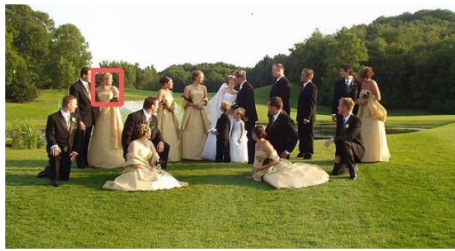

The scenario depicts a couple celebrating a special occasion in a sunny park, where a False Positive person detection occurred due to the occlusion caused by the man's tie and the woman's dress, the close-up view of the woman's face and upper body, and the absence of the rest of her body in the detection region, leading to incomplete detection and misclassification as a person.

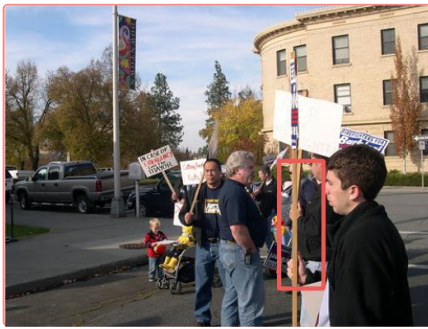

The scenario depicts a group of people, with one person holding a wooden stick, in an image where a False Positive person detection occurred. The most possible reason for the False Positive person detection is that the wooden stick being held by the person, along with the partial obstruction of the person's body by the wooden wall, may have caused confusion for the object detection model, leading to the incorrect classification of the wooden stick as a person.

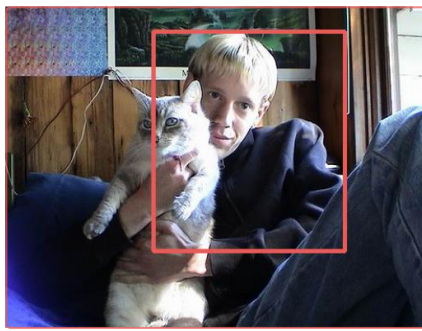

The scenario depicts a young man sitting on a couch, holding a large calico cat, with a partially detected person in the image. The most possible reason for the False Positive PERSON detection is that the person's face and arm were partially visible, leading to a misclassification by the object detection model. This could be due to occlusion by the cat, size of the detected parts, or other factors affecting the visibility of the person.

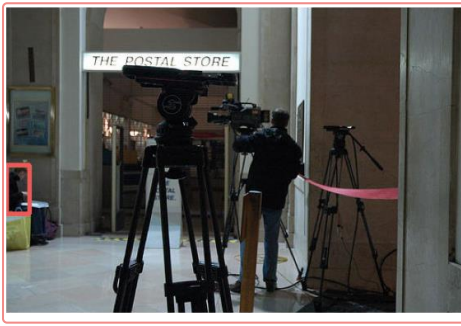
The scenario depicts a person sitting at a table, with their hands resting on it, but only the person's hand and arm are visible in the image. The most possible reason for the False Positive PERSON detection could be due to occlusion, as the rest of the person's body is not entirely visible. and only partial body parts are detected. The model might have mistakenly associated the detected hand and arm with a full person.

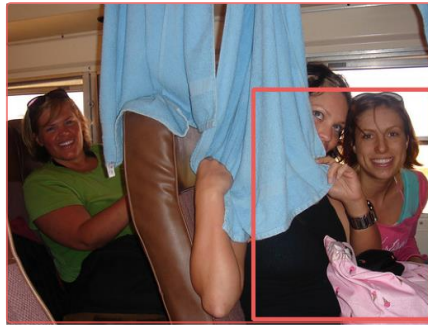
The scenario depicts two women sitting on a bus, with one woman hiding her face behind a blue towel, and various objects such as handbags and a backpack in the image. The most possible reason for the False Positive PERSON detection is that the person's face and arm, along with the hand holding the towel, are only partially visible and may not meet the criteria for a complete person detection. Additionally, the occlusion caused by the towel further complicates the detection.

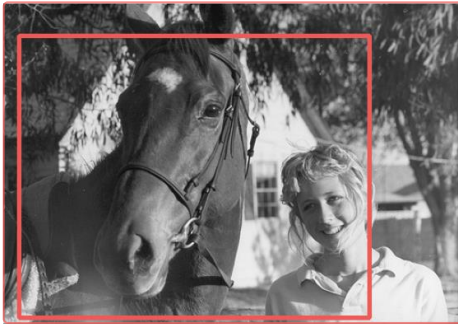
The scenario is a black and white image of a young girl standing next to a brown horse in front of a house, with the girl smiling and the horse wearing a bridle. The most possible reason for the False Positive PERSON detection is that the model may have mistakenly identified the horse's body or parts as a person due to similarities in shape, size, and context, potentially influenced by spurious correlations or missing ground truth.

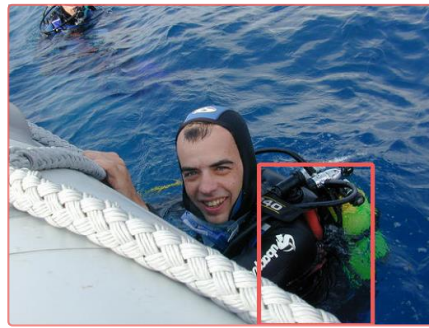
The scenario is an underwater image showing a person wearing a wetsuit and mask, partially submerged in water, surrounded by a blue ocean and a boat in the background. The most possible reason for the False Positive PERSON detection could be the presence of a black and green life preserver, which is partially submerged and partially visible, leading to confusion with a person's body parts.

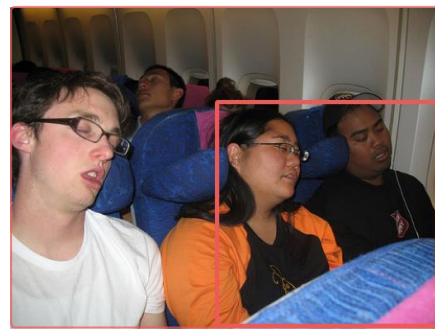
The scenario depicts a group of people sleeping on a plane, with a False Positive detection of a person wearing an orange shirt and glasses. The most possible reason for the False Positive detection could be the occlusion caused by the person's position, the partial visibility of the person's face due to resting on a pillow, and the presence of other objects such as a headrest and another person's head obstructing the view

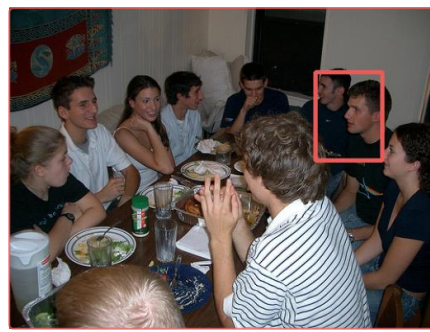
The scenario involves a close-up image of two men's faces, where a person detection model falsely detects a person due to occlusion caused by the heads of other people in the room, the focus on facial features, the close proximity of the detected parts to a bottle in the foreground, and potentially missing or inaccurate ground truth.

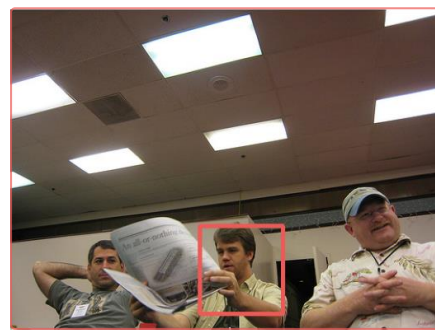
The scenario is a man reading a newspaper in a dimly lit room, with parts of his body and the newspaper obstructed, and another person also reading a newspaper in the room. The most possible reason for the False Positive PERSON detection is that the occlusion caused by the man holding the newspaper in front of his face and the strap covering his body may have led to the detection model incorrectly identifying the visible parts of the man as a separate person. Additionally, the dim lighting condition and the presence of another person reading a newspaper in the room may have added confusion to the detection model.

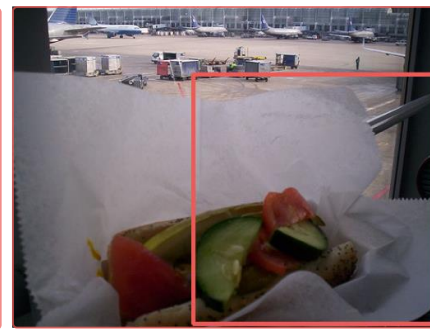
The scenario depicts a sunny day at an airport where a hot dog with toppings is being served, surrounded by parked trucks and cars, airplanes, and people. The most possible reason for the False Positive PERSON detection could be due to the presence of people in the vicinity of the hot dog, which might have led the object detection model to mistakenly identify a person in the image.

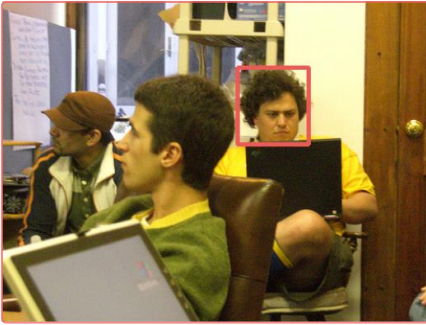
The scenario is a man sitting in a dimly lit room, looking at a laptop, with two other people in the background. The most possible reason for the False Positive PERSON detection is that the model mistakenly classified the man's curly hair as a person due to its prominence in the image, the close-up view of the man's face, and the occlusion of his face by the laptop, leading to a misinterpretation of the detection region.

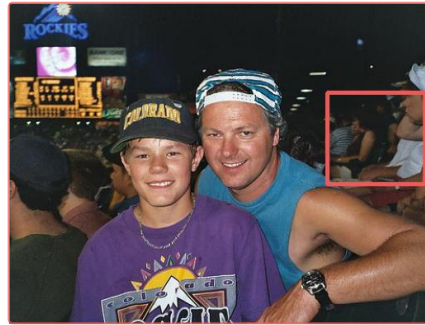
The scenario depicts a crowded stadium with people sitting together, some holding cell phones, and a False Positive person detection occurred due to occlusion and limited visibility of the ground truth person caused by another person sitting in front of them and a person sitting next to them, along with the black and white color scheme adding difficulty in discerning details.

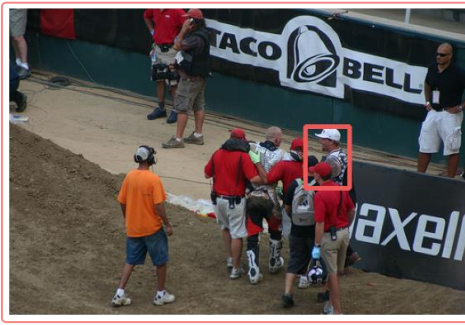
The scenario depicts a group of men standing together at a baseball game, with one man wearing a white hat and another man with a visible tattooed arm. The most possible reason for the False Positive PERSON detection is that the occlusion caused by the other people standing around the person with the tattooed arm, as well as the limited visibility of the person's body, led to a partial detection that matched the model's criteria for a person.

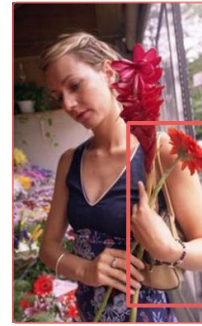
The scenario depicts an image of a woman wearing a blue dress, holding a red flower, with a person detection model producing a False Positive detection of a person. The most possible reason for the False Positive detection could be the presence of the red flower and the handbag, which obstruct the view of the person's body, causing the model to incorrectly classify the flower and handbag as parts of a person. Additionally, the model may be sensitive to the color and shape of the flower and bracelet, leading to a misclassification.

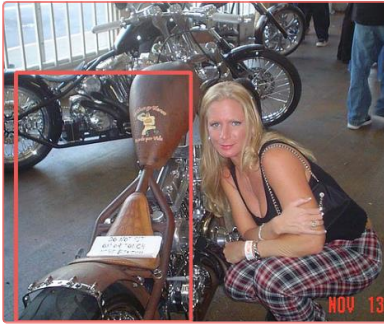
The scenario is a woman posing in front of motorcycles at a gathering or show, with a False Positive person detection caused by the presence of a motorcycle seat decorated with a picture of a man and a sign on the handlebar, along with the occlusion and similarity in appearance between the motorcycle seat and a person.

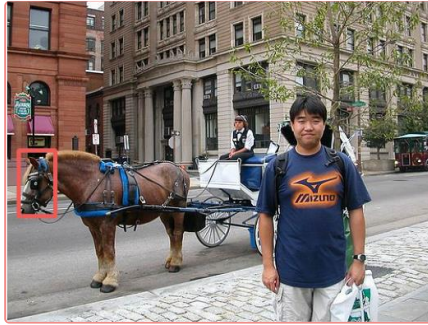
The scenario is a false positive person detection in an image of a brown horse wearing a bridle and harness, standing near a wall in a close-up perspective; the likely reason for the false positive is the presence of the horse's head, which is similar in appearance to a human head, combined with the occlusion caused by the bridle and harness, leading to a misclassification by the person detection model.

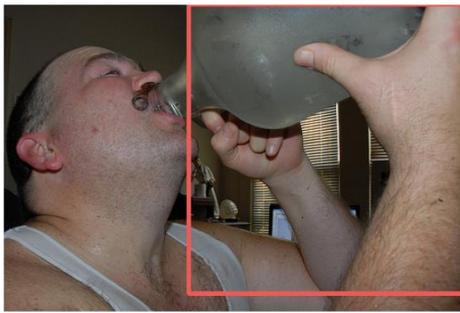
The scenario is a man drinking from a glass in a room, where he is wearing a white tank top and the room has a TV and a keyboard. The most possible reason for the False Positive PERSON detection is that the model mistakenly identified the glass as a person due to its shape and position, and the presence of the man's hand holding the glass further contributed to the confusion.

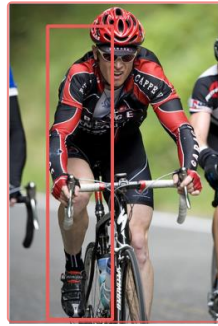
The scenario involves a person riding a bicycle in a sunny environment, with other people present. The most likely reason for the False Positive person detection is that the person's body parts, such as their leg and arm, were mistakenly identified as a person due to their proximity to the bicycle, their red glove, and their outfit, which were the secondary elements in the image

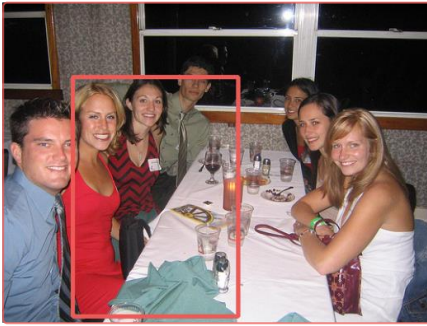
The scenario is a group of people sitting around a dining table, enjoying a meal together, with a False Positive person detection in the image. The most possible reason for the False Positive detection could be the presence of multiple people and objects in the scene, which could cause confusion for the person detection model and lead to incorrect detections. Additionally, factors such as occlusion, viewpoint, and the positioning of the subjects could further contribute to the False Positive detection.

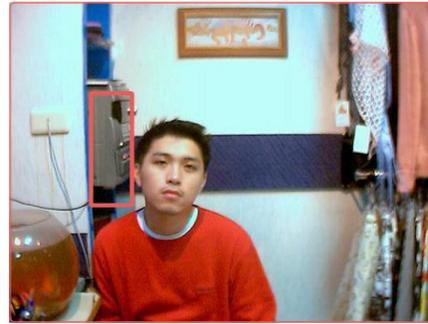
(**Incorrect**) The scenario is that a false positive person detection occurred in an image of a cabinet filled with books, located in a room, with a wall in the background, and partially obscured by a person's head, possibly due to spurious correlation between the shape of the cabinet and a person's body, or due to the presence of books in the cabinet resembling a person's body shape.

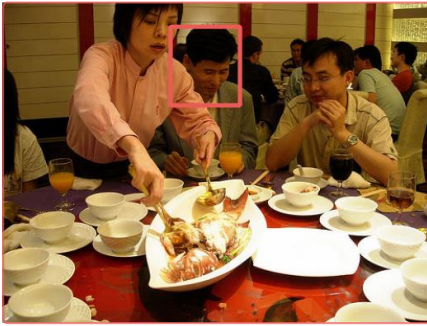
(**Incorrect**) The scenario is a restaurant scene with a man and a woman sitting at a table, surrounded by other people. The most possible reason for the False Positive person detection is the occlusion caused by the reflections of a woman and a man in the mirrors, which obstruct the view of the person. Additionally, the person is sitting down, facing away from the camera, and only parts of their face are visible, leading to a partial detection.

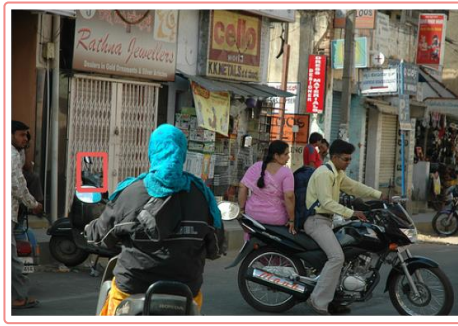
(**Incorrect**) The scenario is set in an urban environment with a motorcycle parked next to a building, and a False Positive person detection occurred due to the reflection of a pair of black boots and a partially visible black mask in the mirror of the motorcycle, where the boots are positioned in front of the mask in the False Positive detection region.

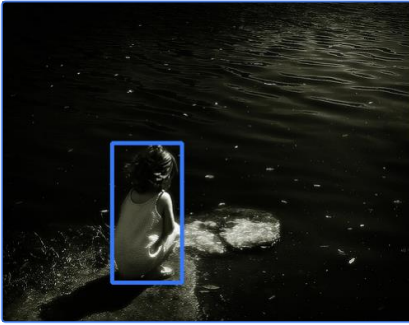
The scenario is a black and white image of a young girl sitting in water at night, wearing a white dress, with a focus on her silhouette and the dark background. The most possible reason for the False Negative person detection is the obstruction caused by the water and darkness, making it difficult for the model to accurately detect and classify the person in the image. Additionally, the lack of discernible facial features and the focus on the upper body further contribute to the False Negative.

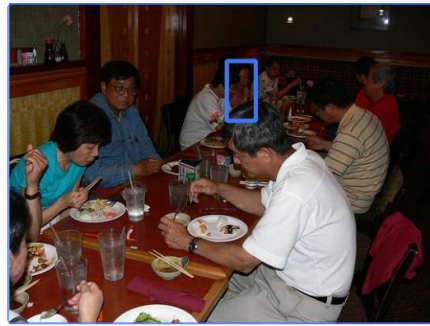
The scenario depicts a False Negative in person detection, where only parts of a person's face are detected, and the most possible reason for this is the obstruction caused by the person's own hair and other objects, such as a potted plant and a piece of paper, partially covering the person's face, as well as the dim lighting conditions and the presence of other people in the scene.

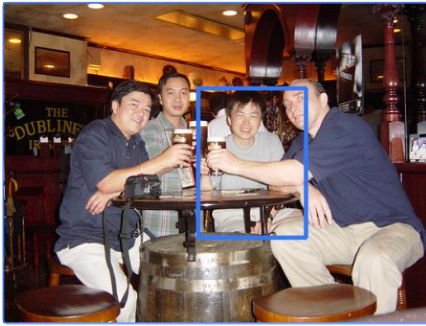
The scenario depicts a group of men sitting at a dining table, enjoying beer and food, with one man holding a glass of beer. The most possible reason for the False Negative in person detection is that the person's head and shoulders are partially obscured by the table and other people, making it difficult for the model to detect the person accurately. Additionally, the dim lighting and crowded environment may further contribute to the False Negative.

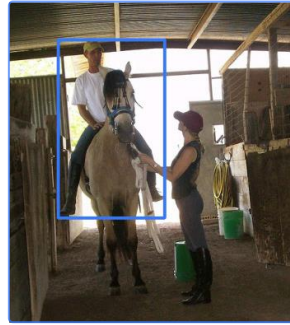
The scenario depicts a man sitting on a horse, with a woman standing next to him, in a dimly lit indoor setting, where the person detection model fails to detect the person due to occlusion by the horse and the woman, low visibility of the person's face and body parts, and the focus of the image being on the horse and its rider rather than the person.

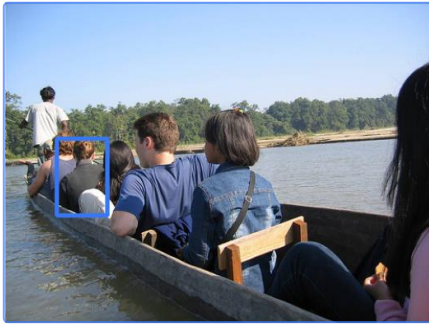
The scenario is a cropped image showing a group of people sitting in a boat on a river, with a person wearing a black jacket and a white shirt partially obstructed from view. The most possible reason for the False Negative in the person detection is the occlusion caused by the backs of the people sitting in front of the person, as the person's upper body and face are not clearly visible in the image.

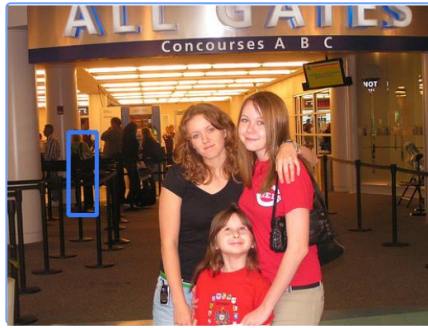
The scenario depicts a person standing behind a wooden fence in a dimly lit environment, wearing a yellow shirt and holding a handbag, with obstructed view, a small size, and only detected parts of the person visible, such as the back, backpack, and hair. The most possible reason for the False Negative detection could be the occlusion caused by the wooden fence, the person's silhouette making it difficult to detect the person's features, and the dark lighting conditions affecting the model's ability to accurately detect the person.

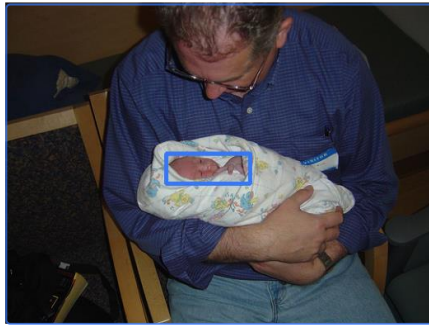
The scenario depicts an image of a baby lying in a crib, partially covered by a blanket, with only the baby's face visible. The most possible reason for the False Negative in person detection is that the model fails to detect the person due to occlusion caused by the blanket and the baby's hand, as well as the limited visibility of the person's body and face in the image.

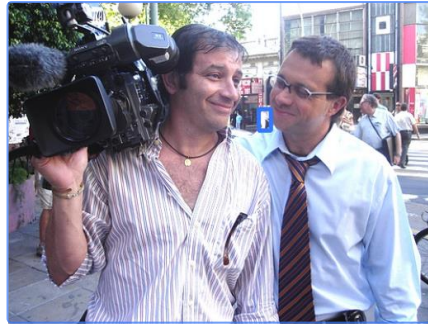
(**Incorrect**)The scenario involves a person wearing a white shirt and hat, standing in a brightly lit area, with their arms obstructing their face, and the lower part of their body not clearly visible. The most possible reason for the False Negative in the person detection is the occlusion caused by the person's own arms, which prevents the model from accurately detecting the full extent of the person's presence.



%---------------------------------

\end{document}